\documentclass[sigconf,nonacm]{acmart}
\AtBeginDocument{%
  }

\setcopyright{acmlicensed}
\copyrightyear{2025}
\acmYear{2025}
\acmDOI{XXXXXXX.XXXXXXX}

\acmJournal{TOG}
\acmVolume{00}
\acmNumber{0}
\acmArticle{000}
\acmMonth{12}

\acmSubmissionID{papers\_2010}

\usepackage{graphicx}
\usepackage{subcaption}
\usepackage{epigraph}
\usepackage{cleveref}
\usepackage{placeins}

\usepackage{algorithm}
\usepackage[noend]{algpseudocode}

\newcommand{\ie}{\textit{i.e.}}
\newcommand{\eg}{\textit{e.g.}}

\begin{document}

\title{Palette Aligned Image Diffusion}

\author{Elad Aharoni}
\author{Noy Porat}
\author{Dani Lischinski}
\affiliation{%
  \institution{Hebrew University}
  \country{Israel}
}
%\email{{elad, dani}@huji.ac.il}

\author{Ariel Shamir}
\affiliation{%
  \institution{Reichman University}
  \country{Israel}
}
%\email{ariel@reichman.ac.il}

% \twocolumn[{%
% \renewcommand\twocolumn[1][]{#1}%
%\maketitle
\begin{teaserfigure}
    \centering
    \includegraphics[width=1.0\textwidth]
    {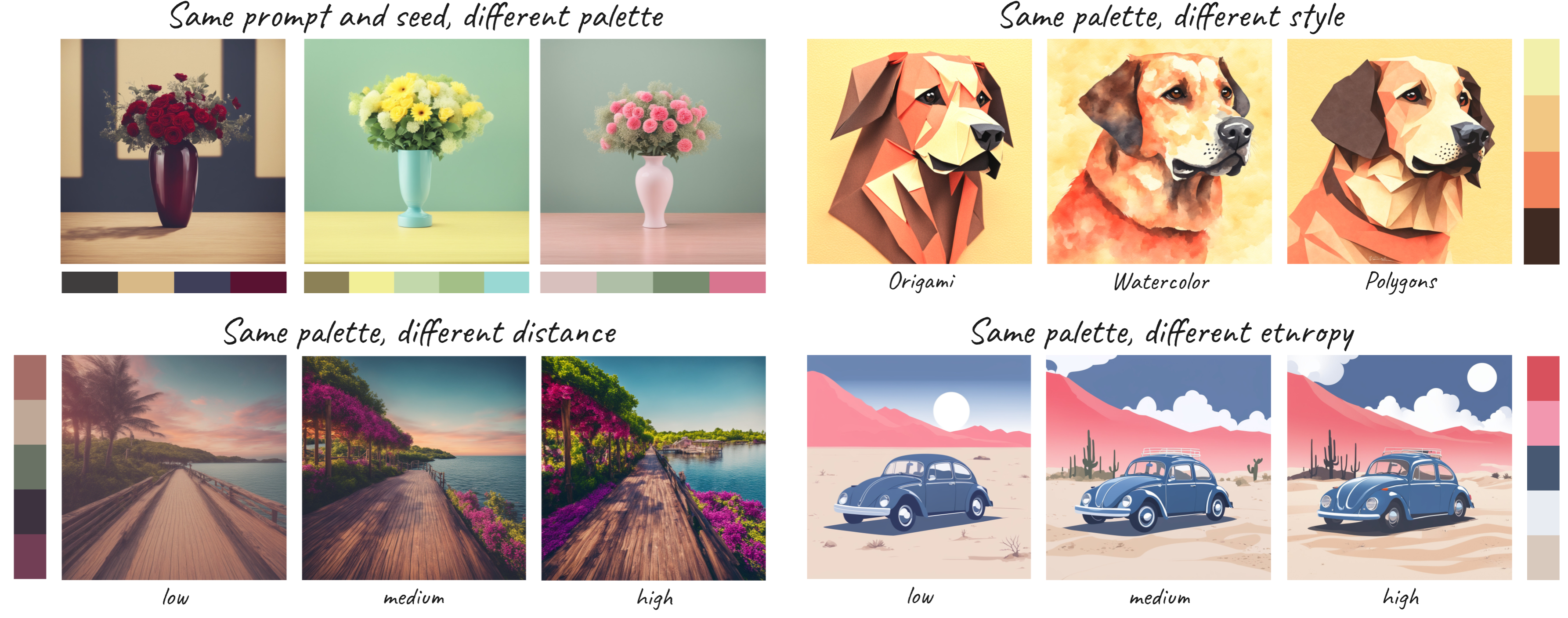}
    \caption{Palette Aligned Image Diffusion: Our method enables precise control over color distribution, generating images, from richly colored to highly quantized, that are perceptually aligned with user-specified color palettes and consistent across different styles.
    }
    \Description[<short description>]{<long description>}
    \label{fig:teaser}
    \Description{A teaser image showing different capabilities of Palette Aligned diffusion.}
\end{teaserfigure}

\begin{abstract}
We introduce the \emph{Palette-Adapter}, a novel method for conditioning text-to-image diffusion models on a user-specified color palette. While palettes are a compact and intuitive tool widely used in creative workflows, they introduce significant ambiguity and instability when used for conditioning image generation. Our approach addresses this challenge by interpreting palettes as sparse histograms and introducing two scalar control parameters: \emph{histogram entropy} and \emph{palette-to-histogram distance}, which allow flexible control over the degree of palette adherence and color variation. We further introduce a \emph{negative histogram} mechanism that allows users to suppress specific undesired hues, improving adherence to the intended palette under the standard classifier-free guidance mechanism. To ensure broad generalization across the color space, we train on a carefully curated dataset with balanced coverage of rare and common colors. Our method enables stable, semantically coherent generation across a wide range of palettes and prompts. We evaluate our method qualitatively, quantitatively, and through a user study, and show that it consistently outperforms existing approaches in achieving both strong palette adherence and high image quality.

%This paper introduces a novel method for controlling the colors in images generated by text-to-image diffusion models, based on a desired color palette. We address the inherent ill-posed nature of color palette conditioning, by representing them as sparse histograms and incorporating a distance measure between the palette and the full image histogram, along with histogram entropy, to allow flexible color control.  Our approach leverages a new IP-Adapter architecture, specifically designed for color palette conditioning. This adapter is trained using a large curated dataset that ensures a balanced representation of diverse colors, mitigating biases present in existing datasets. To enhance color alignment, we introduce a dual conditioning approach using both positive and negative color histograms during classifier-free guidance. This, combined with the entropy condition that controls the level of color quantization, enables precise control over the color distribution, generating images ranging from richly colored to highly quantized, perceptually aligned with user-specified color palettes. Our method is evaluated both qualitatively and quantitatively, demonstrating its effectiveness and superiority over previous approaches in achieving precise color control while maintaining coherence, diversity, and fidelity.
\end{abstract}

\maketitle

\section{Introduction}
\label{sec:intro}

% Paragraph 1: Motivation for color control in image generation and positioning of our contribution
Color plays a central role in visual media, shaping emotion, defining style, and directing attention. In both art and design, the ability to control color is essential:  whether to evoke a mood, adhere to brand identity, or achieve aesthetic coherence. Recent advances in text-to-image diffusion models~\citep{rombach2022highresolutionimagesynthesislatent,ramesh2022hierarchicaltextconditionalimagegeneration,Saharia2022PhotorealisticTD,podell2023sdxlimprovinglatentdiffusion} have enabled the creation of highly diverse and photorealistic images from textual prompts. Meanwhile, mechanisms such as adapters~\citep{ye2023ip-adapter,mou2023t2iadapterlearningadaptersdig} and ControlNet~\citep{zhang2023addingconditionalcontroltexttoimage} have introduced fine-grained control over image structure and style. In this work, we focus specifically on controlling the \emph{colors} that appear in the generated image, by conditioning on a user-specified \emph{color palette}. Unlike 2D maps used in adapters or ControlNet, a palette is a compact and simple 1D discrete representation that is intuitive and is widely used in creative workflows (see \Cref{fig:teaser}).

% Paragraph 2: Define what a palette is and motivate the core challenges of using it as a conditioning signal
A color palette is typically made up of a small fixed set of colors (usually 3–10) that guides artists and designers when creating a specific piece or visual identity. The practice of using color palettes dates back more than 40,000 years, when early human artists adopted a basic set of five pigments~\cite{chaudhary2019influence}. Palettes offer a compact, expressive, and visually intuitive means of specifying the intended color theme of an image, without requiring detailed text prompts, reference images, or full color-distributions. However, conditioning generative models on a palette alone is a highly ambiguous and ill-posed task. One challenge is \emph{non-uniqueness}: different color distributions may share the same palette, and a single palette can correspond to a wide variety of plausible images. A second challenge is \emph{instability}: small changes in the input palette can lead to large, unpredictable differences in the generated output. Moreover, a palette is typically intended as coarse guidance — users may expect images to reflect the general hues and style of the palette, while still allowing for subtle shades, blends, and occasional deviations. This tension between expressive flexibility and visual consistency makes palette conditioning especially challenging.
%\dani{refer to teaser here as well?}

\begin{figure}
    \centering
    \includegraphics[width=1\linewidth]{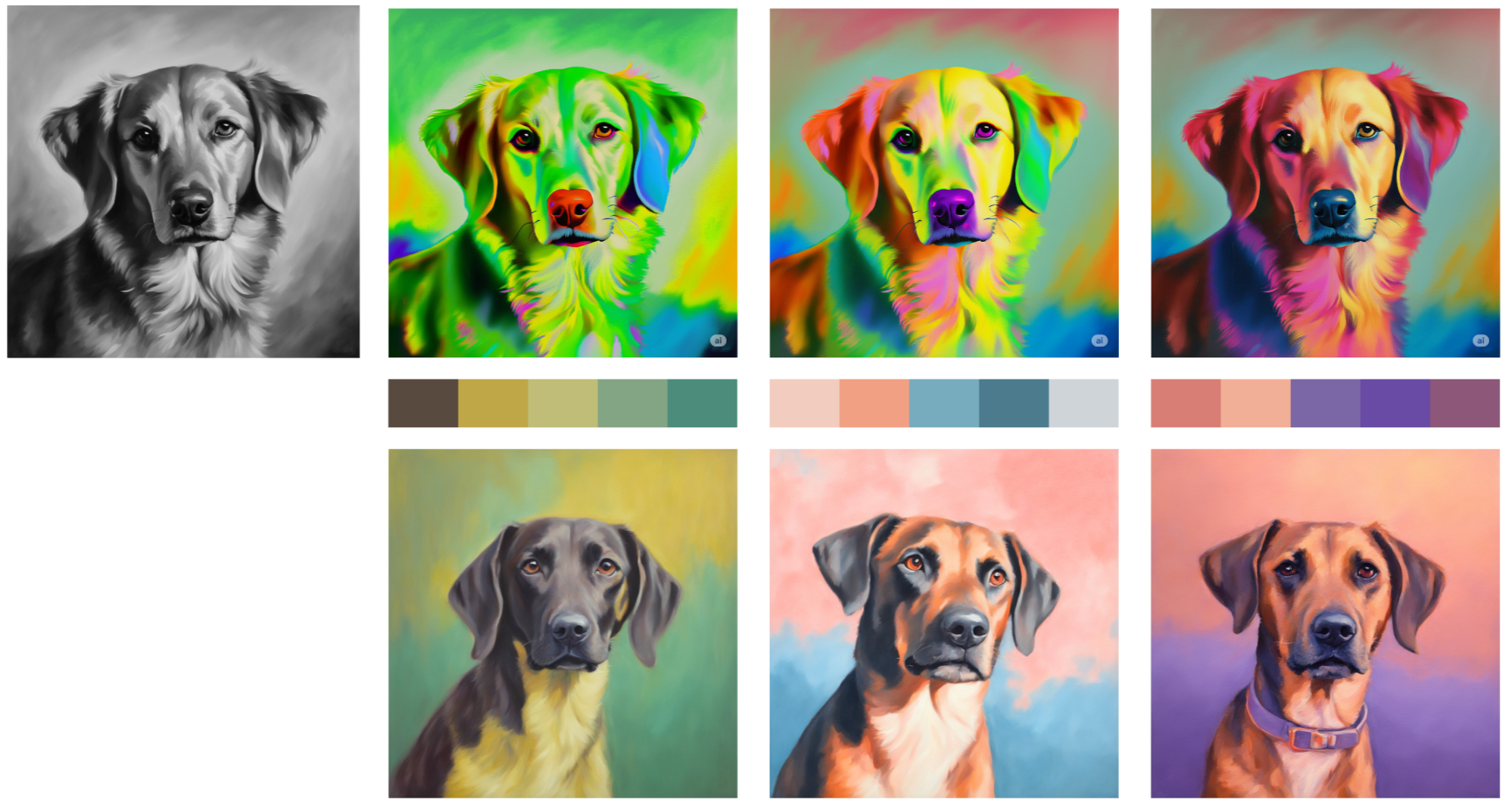}
    \caption{\textbf{Post-hoc recoloring of a grayscale image vs.~palette-guided generation}: we compare Gemini's palette-based grayscale recoloring (top row) and our model's palette-aligned generation (bottom row). Our model yields superior color alignment and adapts content—specifically the dog's breed—to the palette, creating semantically coherent images. Importantly, all images share the same prompt and seed. For Gemini, the grayscale image and palette were provided with the prompt: ``Please recolor the grayscale image using the provided palette.''}
    \Description{will be added}
    \label{fig:recoloring}
\end{figure}

% Paragraph 3: Explain limitations of existing recoloring and joint conditioning methods; motivate the need for a new approach
One might consider a two-stage approach for palette-guided generation: produce a grayscale image, followed by post-hoc recoloring to match the desired hues. However, it is not simple to color a grayscale image using a palette. In addition, such decoupling breaks the alignment between color and content, as the recoloring step is typically unaware of the prompt semantics and may yield visually plausible but semantically inconsistent results (\Cref{fig:recoloring}, top row). Existing methods for joint conditioning attempt to overcome this limitation, but they often rely on strong additional inputs — such as a reference image~\cite{ye2023ip-adapter}, a spatial color layout~\cite{mou2023t2iadapterlearningadaptersdig}, or an implicit color distribution derived from a reference image~\cite{lobashev2025colorconditionalgenerationsliced}. While effective in some scenarios, these approaches do not support lightweight, intuitive control using a simple palette, and they often struggle to maintain prompt-color alignment without sacrificing content fidelity or diversity. Our method addresses these limitations by enabling fine-grained, stable color control using just a prompt and a palette (see \Cref{fig:teaser}).
%\dani{add figure to accompany this paragraph?} \elad{the teaser is not sufficient? what did you have in mind?}

% Paragraph 4: Introduce our method (Palette-Adapter), highlight its architectural and functional novelty, and describe the key control parameters
We introduce the \textit{Palette-Adapter}, a novel adapter for diffusion-based text-to-image models that enables precise and flexible control over color through palette-based conditioning. Although inspired by the IP-Adapter~\citep{ye2023ip-adapter}, our method departs from it in both functionality and design. Unlike IP-Adapter, which encodes spatial image features, Palette-Adapter conditions the model on a non-spatial color palette, applied at a single intermediate block in the denoising U-Net.
%\elad{do we need to add a figure to show the artifacts between single and multi blocks?}
To address the ambiguity of palette-based generation, we interpret the input palette as a degenerate histogram, one with a small number of uniformly weighted bins, and train the adapter using both the full image histogram and a derived palette histogram. In addition, we introduce two scalar conditions: the \emph{entropy} of the target and its \emph{distance} from the full image histogram. During inference, these parameters allow users to modulate the level of adherence to the palette. The entropy condition governs the overall color variability, while the distance condition permits controlled deviations beyond the original palette - enabling results that range from stylized, quantized compositions to rich and naturalistic renderings (see \Cref{fig:teaser,fig:entropy}).

% Paragraph 5: Introduce the concept of a negative histogram for enhanced palette guidance via CFG
To further improve control during inference, particularly when using Classifier-Free Guidance (CFG)~\cite{ho2022classifierfreediffusionguidance}, we introduce the concept of a \emph{negative histogram} — a set of user-specified colors that should be suppressed in the generated image. Unlike standard CFG, which extrapolates away from an unconditioned result, our method allows extrapolation along a more semantically meaningful direction: from the negatively guided output toward the positively guided one. This formulation improves both stability and adherence to the desired palette, allowing users to reduce the presence of unwanted hues while retaining high-fidelity palette-aligned results (see \Cref{sec:CFG}).

% Paragraph 6: Address color bias in training data and explain our curation strategy
Training the Palette-Adapter requires data with diverse and balanced color distributions. We found that the LAION-Art \cite{laion_art} dataset, which we initially used for training, exhibits strong color biases: certain hues appear frequently, while others are relatively rare. As a result, many histogram bins are under-represented, limiting the model’s ability to generalize to diverse palettes.
%\elad{we dont have ablation for this, but perhaps we can generate an image that shows that we do better with saturated rare colors compared to other models. wdyt?}
To address this, we curate an additional set of images that emphasize low-probability colors and augment the training set accordingly. This targeted curation ensures better coverage of the color space and improves the robustness of our model to a wider range of palettes (see Section~\ref{sec:dataset}).

% Paragraph 7: Summarize our contributions and main findings
In summary, our contributions are as follows.
%\elad{do we need to add here the noveltiy of the adapter?}
\begin{enumerate}
\item \textbf{Palette-to-histogram conditioning:} We cast palette conditioning as a special case of histogram conditioning, enabling more structured training and introducing entropy and distance as interpretable control parameters.
\item \textbf{Enhanced guidance mechanisms:} We introduce entropy- and distance-based controls, to enable finer and more stable control over the generated color appearance, as well as positive and negative histogram conditioning. 
\item \textbf{Curated training set:} A color-balanced dataset that improves generalization in rare and diverse palette configurations.
\end{enumerate}

We train our palette adapter on stable diffusion XL~\cite{podell2023sdxlimprovinglatentdiffusion}, and evaluate it on a wide range of prompts and palettes. Our results include qualitative comparisons, quantitative metrics (\Cref{sec:exp_quantitative}), and a user study (\Cref{sec:exp_user_study}), demonstrating the effectiveness and flexibility of our approach. In particular, we show that the entropy and distance parameters provide intuitive handles for controlling stylistic abstraction and palette adherence with minimal user effort (\Cref{sec:entropy}).
Our model and data will be made available upon publication.

\section{Related Work}
\label{sec:related}

\subsection{Color Palettes}

Color palettes are a widespread tool for artists and designers. They are used to ensure visual harmony and consistency. Palettes can be composed of different types of colors based on their relationships on the color wheel, mood, or theme. The evolution of color theory, pigments, and digital media, further developed a set of rules for selecting color palettes (\eg, Contrastive, Analogous, Harmonic, Triadic, and more~\cite{wiki:Color_scheme}). Nowadays many artists use reference books~\cite{ColorComb} or digital tools (\eg, \citet{website:adobe_color} and~\citet{coolorsCoolorsSuper}) to explore, create and choose from millions of aesthetically pleasing, predefined palettes.

\subsection{Color Transfer and Recoloring}

Over the years, there has been much work on the transfer of color from one image to another, or on colorization of grayscale images. Early work included the pioneering approach of ~\citet{reinhardcolortransfer}, which transfers colors by matching global color statistics, an approach that was later improved by several follow-up works, \eg, \cite{PITIE2007123}. Later works explored palette-based recoloring techniques for photographs~\cite{chang:2015:ppr}, and watercolor paintings~\cite{pigmentsbasedrecoloring}. While recoloring techniques can produce pleasing results, they have several limitations:
\begin{enumerate}
\item These methods are not generative, and therefore address the recoloring of an already existing image.
\item The global transfer of color statistics may produce unexpected results.
\item Recoloring does not adapt the composition of the source image to align with the desired palette, potentially violating semantic color coherence.
\end{enumerate}

\subsection{Non-Textual Conditioning in Diffusion Models}
Several methods were proposed to guide diffusion models with non-textual conditions. DALL-E 2~\cite{ramesh2022hierarchicaltextconditionalimagegeneration} makes the first attempt to support image prompts, where the diffusion model is conditioned on image embedding rather than on text embedding. IP-Adapter~\cite{ye2023ip-adapter}, ControlNet~\cite{zhang2023addingconditionalcontroltexttoimage} and T2I-Adapter~\cite{mou2023t2iadapterlearningadaptersdig}, have all demonstrated mechanisms to add conditioning to an existing model. IP-Adapter employs duplicates of the existing cross-attention text layers to allow additional independent conditional input. Maintaining the fidelity and text control during inference is typically done using Classifier Free Guidance (CFG). Our Palette-Adapter is inspired by IP-Adapter, but uses a non-spatial palette condition, and applies it at a single intermediate U-net block.

\subsection{Color-Based Image Generation}

Also related to our work is Composer~\cite{huang2023composercreativecontrollableimage}, which conditions image generation on a smoothed CIELab histogram rather than palettes. However, their model is trained from scratch instead of extending an existing pre-trained generative model. Another closely related work is T2I-Adapter~\citep{mou2023t2iadapterlearningadaptersdig}, which provides a low-resolution 2D image as condition. Thus, this is effectively a super-resolution approach rather than palette guidance.

A recent approach by ~\citet{lobashev2025colorconditionalgenerationsliced} modifies the diffusion sampling process to incorporate the differentiable Sliced 1-Wasserstein distance between the color distribution of the generated image and the reference palette/histogram. However, their method requires a significant number of inference/decode steps (approx. x8), and struggles with out-of-distribution latent perturbation, which we found to compromise image fidelity, especially when working with sparse palettes instead of full histograms. To further clarify, one can consider the degenerate case of an Earth Mover's Distance (EMD) of 0, where each pixel in the image is strictly taken from the provided palette with blanced distribution. This scenario would inevitably produce severely color-collapsed and heavily posterized images. To establish a practical and aesthetically pleasing target, we calculate a ``good'' target EMD by inspecting images for which palettes were meticulously extracted by designers, as further detailed in our quantitative evaluation. We conclude that an optimal palette based generation, approach must strike balance between prompt and palette adherence while maintaining fidelity.

Crucially, neither of these prior works explicitly addresses the inherent ambiguity of palette conditioning or provides controls to mitigate it, such as our use of negative palettes and palette-to-histogram distance. Additionally, neither offers further refinement controls like our entropy-based method, which we found highly beneficial, as demonstrated later in this paper.

\section{Method}
\label{sec:method}

Our goal is to enable palette-based control over the colors in text-to-image diffusion models, while maintaining their ability to generate high-quality images. The resulting image must be semantically aligned with a prompt, as well as perceptually aligned with a color palette. This is challenging since color palettes are inherently ambiguous which makes this an ill-posed task, as explained in \Cref{sec:intro}.
To achieve our goal, we propose a novel Palette-Adapter, inspired by the IP-Adapter \cite{ye2023ip-adapter} architecture, but featuring several key novel components that enable non-spatial palette conditioning:
%Our adapter conditions only a single block within the denoising U-Net, inspired by \citep{frenkel2024implicitstylecontentseparationusing, cohen2024conditionalbalanceimprovingmulticonditioning}. 
%Our approach combines the following key novel components: 

\begin{enumerate}
\item Color palettes are treated as a sparse color histograms, allowing flexible conditioning and quantitative color specification both during training and inference (\Cref{sec:conditioning}). 

\item The distribution and variety of colors in the output is managed using two controls. The \emph{distance} condition controls the amount of deviations of colors outside the palette. The \emph{entropy} controls the variation distribution of colors within the palette (\Cref{sec:entropy}).

\item Color alignment is improved by employing dual conditioning: using both positive and negative color conditions. This is implemented via Classifier-Free Guidance (CFG). Guiding away from unwanted hues improves adherence and reduces unrelated colors (\Cref{sec:CFG}).

\item Training uses a balanced conditioning inspired by \citep{frenkel2024implicitstylecontentseparationusing, cohen2024conditionalbalanceimprovingmulticonditioning}. We use a multistage training strategy. Initially, all U-Net blocks are conditioned to identify the critical block for color control. Subsequently, the adapter is finetuned  by conditioning only on these blocks. This approach helps prevent overfitting and structural artifacts (See supplementary materials, \Cref{sec:colorBlocks}).

\item Training uses a curated dataset that addresses existing color biases and ensures balanced representation of diverse colors. The dataset augments LAION-Art with images containing colors from under-represented bins (\Cref{sec:dataset}).

\end{enumerate}

\subsection{Encoding Color Conditions}
\label{sec:conditioning}

How to compare the multiple colors of an image to a discrete input palette is ambiguous. In addition, there is a vast combinatorial space of possible palettes, and their variable length makes sequence-based encoding non-trivial during training.
To address these challenges, we encode the palette color conditions as sparse histogram embeddings. This approach offers several advantages:
\begin{enumerate}
\item \emph{Unified Representation:} both sparse palettes and full histograms can be supported as positive or negative conditions.
\item \emph{Quantitative Control:} desired weights can be specified for the palette's colors, if desired.
\item \emph{Training Stability:} both full and sparse histograms can be mixed to reduce sample complexity and dependency on a specific palette extraction method.
\item \emph{Additional Controls:} additional conditioning controls can be defined, such as \emph{entropy} and palette-to-histogram \emph{distance}.
\end{enumerate}

We chose to discretize the color space using a 3D histogram in HSV color space with (34, 12, 10) bins for the H, S, and V channels respectively. Such representation is inspired by the Munsell Color System, whose color atlases commonly feature 40 distinct hues to represent significant perceptual differences,  5–10 chroma levels, and 11 lightness levels. This choice of granularity allows us to capture subtle color variations reflecting the human capacity to discern numerous hues. Our scheme provides a balance between perceptual color resolution and computational tractability, remaining efficient for conditioning and data sampling.

The color histogram of an image is defined by  projecting all pixel colors into the histogram bins, while a palette is represented by projecting its colors into the corresponding histogram bins with uniform probabilities. During training, we alternate between the palette and the histogram conditions with equal probability, employing a dedicated scalar to inform the model of \emph{condition type} (\Cref{fig:training}). Explicitly providing this condition type allows the model to learn and generalize the inherent differences between palettes and histograms, while also effectively capturing the color information within each bin. 

\subsection{Adapter Architecture and Training}
\label{adapter_architecture}
We propose a multistage training of an IP-Adapter \cite{ye2023ip-adapter} as further explained in the supplementary materials (\Cref{sec:colorBlocks}). We use conditioning of pre-trained frozen diffusion model by adding decoupled cross-attention layers to their denoising U-Net. However, while IP-Adapters are typically used with spatial image data, and used to target \emph{all} U-Net layers, we modify them to accept non-spatial data, and target \emph{specific} layers, making them suitable for our histogram-based color conditioning.

\begin{figure}
    \centering
    \includegraphics[width=1\linewidth]{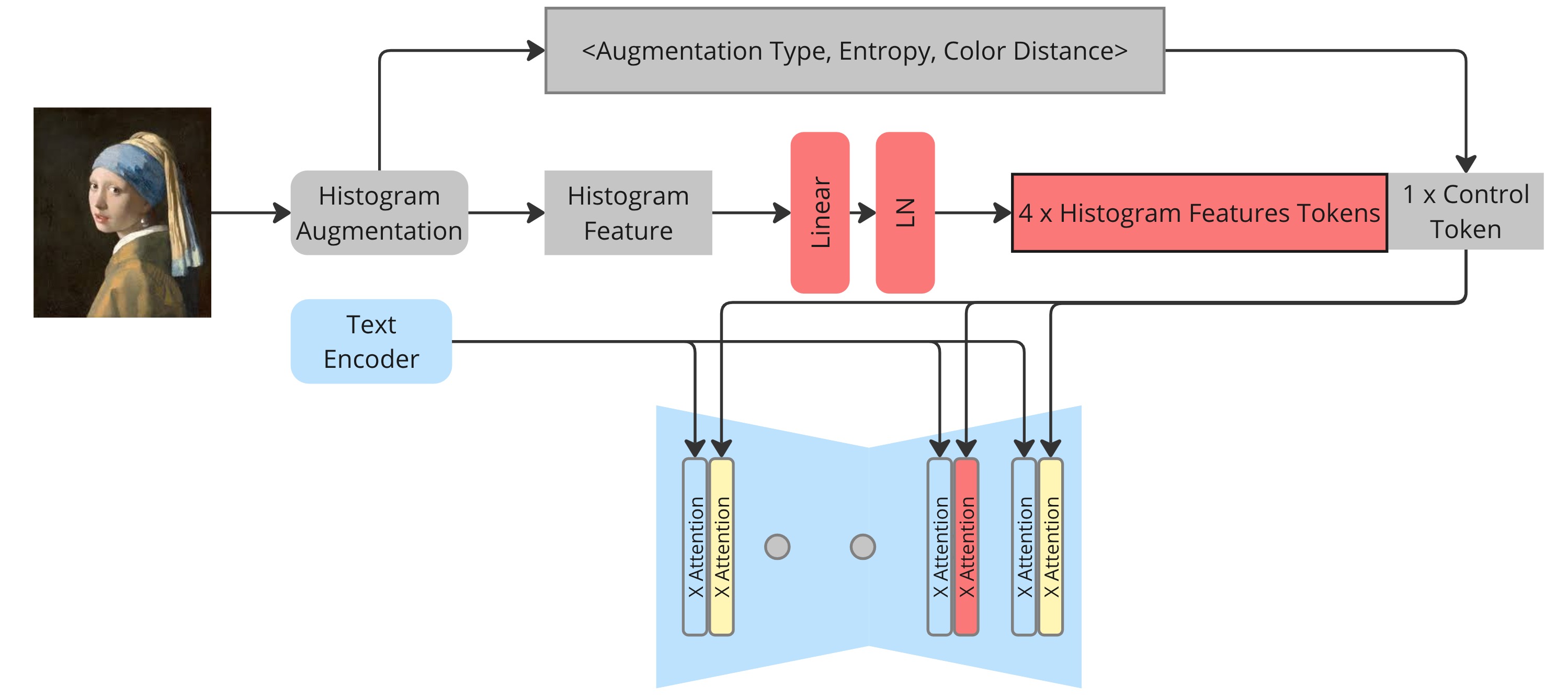}
    \caption{\textbf{Training the Palette-Adapter:} During training, we randomly choose the augmentation type (histogram, palette, or unconditioned), and compute the distance between the extracted palette/histogram and the full image histogram, as well as the full histogram's entropy. The histogram is projected and normalized into 4 tokens. The remaining features are added as a separate 5th token. All tokens are fed into the adapter's cross-attention layers. We employ a multistage training strategy as described in ~\Cref{sec:colorBlocks}: trainable layers are colored \emph{\color{red}red}, while frozen layers are \emph{\color{blue}blue}. Layer participating only during the initial training phase are colored \emph{\color{yellow}yellow}. All gray components are fixed.}
    \label{fig:training}
    \Description{will be added}
\end{figure}

Our architecture, shown in \Cref{fig:training}, randomly chooses whether to use a sparse color palette or a full histogram as the condition for each training image. The model projects the chosen condition into four tokens through a trainable projection layer, feeding them into the adapter's cross-attention layers. This enables color conditioning throughout the entire diffusion process, similarly to text conditioning. Crucially, this allows the model to both enforce the color conditions and generate compositions that are semantically aligned with them. For instance, a palette can affect the type of the flowers in \Cref{fig:teaser}, or the breed of the dog in \Cref{fig:recoloring}.

To enhance control and alignment, we supplement the color condition with three additional scalar features:
\begin{enumerate}
\item \emph{Augmentation type}: palette, histogram, or unconditioned;
\item Distance between the condition and the full histogram of the input image;
\item Entropy of the full image histogram.
\end{enumerate}

\emph{Entropy} is a global measure of the colors distribution information in the image. Posterized or monochromatic images have low entropy, while vivid, colorful and gradient-rich images have high entropy. We show later how entropy can be used to guide the generation process independent of the histogram condition.

During training, only the Palette-Adapter layers are updated while the pre-trained diffusion model and color condition extraction modules remain frozen. The model processes five attention tokens: four for the projected histogram and one for the additional scalar features.
The procedure for identifying color-critical layers is presented in~\Cref{sec:colorBlocks}.

In our implementation, we adopt the following design choices, though alternatives are viable if consistently applied:
\begin{enumerate}
    \item \emph{Palette extraction}: For training, we use the median-cut implementation by \citet{qtiptipPylette} to extract color palettes from images in our dataset. We extract up to 8 colors to increase the diversity of the sampled colors (compared to just 3--5).
    \item \emph{Color distance}:  To measure distances between color, we use a clipped and sharpened variant of CIEDE2000~\cite{sharma2005ciede2000}, similar to~\cite{fastandrobustearthmoversdistance}.
    \item \emph{Palette to histogram distance}: To measure the distance between the extracted palette and the full histogram of the image, we use the Quadratic-Chi Histogram Distance~\cite{inproceedings_quadratic_chi}. This method takes cross-bin relationships into account, while reducing the effect of differences caused by bins with large values. For bin-to-bin similarity we use the above mentioned color distance metric.
\end{enumerate}

% Negative color condition please do not remove.
%\algrenewcomment[1]{\(\triangleright\) #1}
%\algnewcommand{\LineComment}[1]{\State \(\triangleright\) #1}
%\begin{algorithm}
%\caption{Generating the ``negative'' palette}\label{euclid}
%\footnotesize
%\begin{algorithmic}[1]
%\Procedure{NegativePalette}{}
%\For {i in $num\_bins$}
%\LineComment{
%Distance between histograms using \hfil \break $skimage\_color.deltaE\_ciede2000$ 
%where kL=1, kC=0.3, kH=0.1 \hfil
%}
%\State $palette^-(i) \gets Distance(palette^+, onehot(i))$
%\EndFor
%\State $palette^-(i) \gets Max(palette^-(i) - palette^+,0)$
%\State $palette^-(i) \gets Pow(palette^-(i), 4)$
%\LineComment{Normalize to sum 1.}
%\State $palette^- \gets Normalize(palette^-)$ 
%\State return H
%\EndProcedure
%\end{algorithmic}
%\label{negative_hist_algorithm}
%\end{algorithm}

\begin{figure}
    \centering
    \includegraphics[width=1\linewidth]{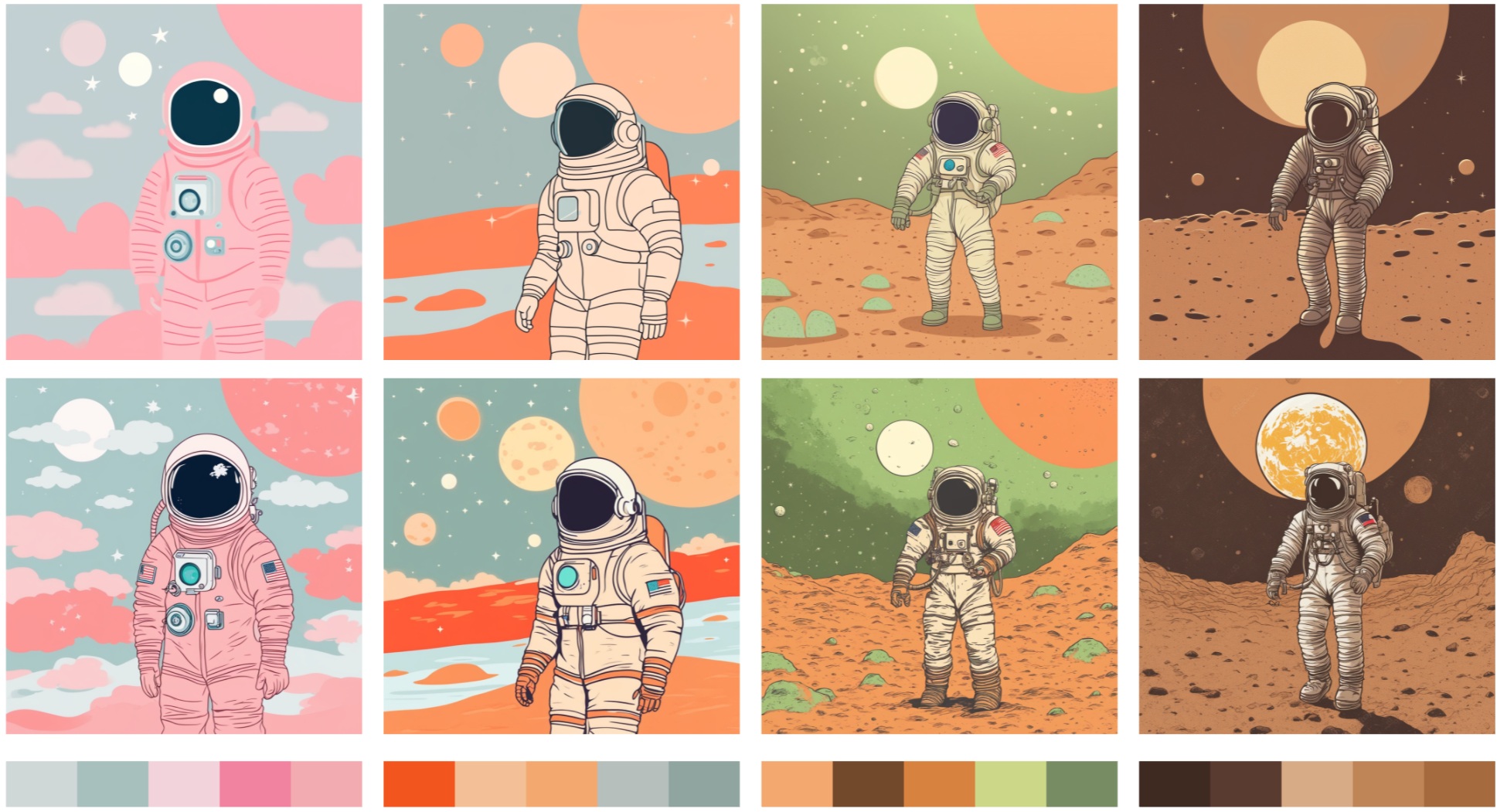}
    \caption{Entropy conditioned images, generated with low and high levels of relative entropy, and different palettes for the prompt ``A poster of an astronaut on the moon. Flat colors. Flat vector sticker style'' and the same seed. Note how as the relative-entropy grows in the second row, more color shades, texture and details are added.}
    \label{fig:entropy}
    \Description{will be added}
\end{figure}

\begin{figure}
    \centering
    \includegraphics[width=1\linewidth]{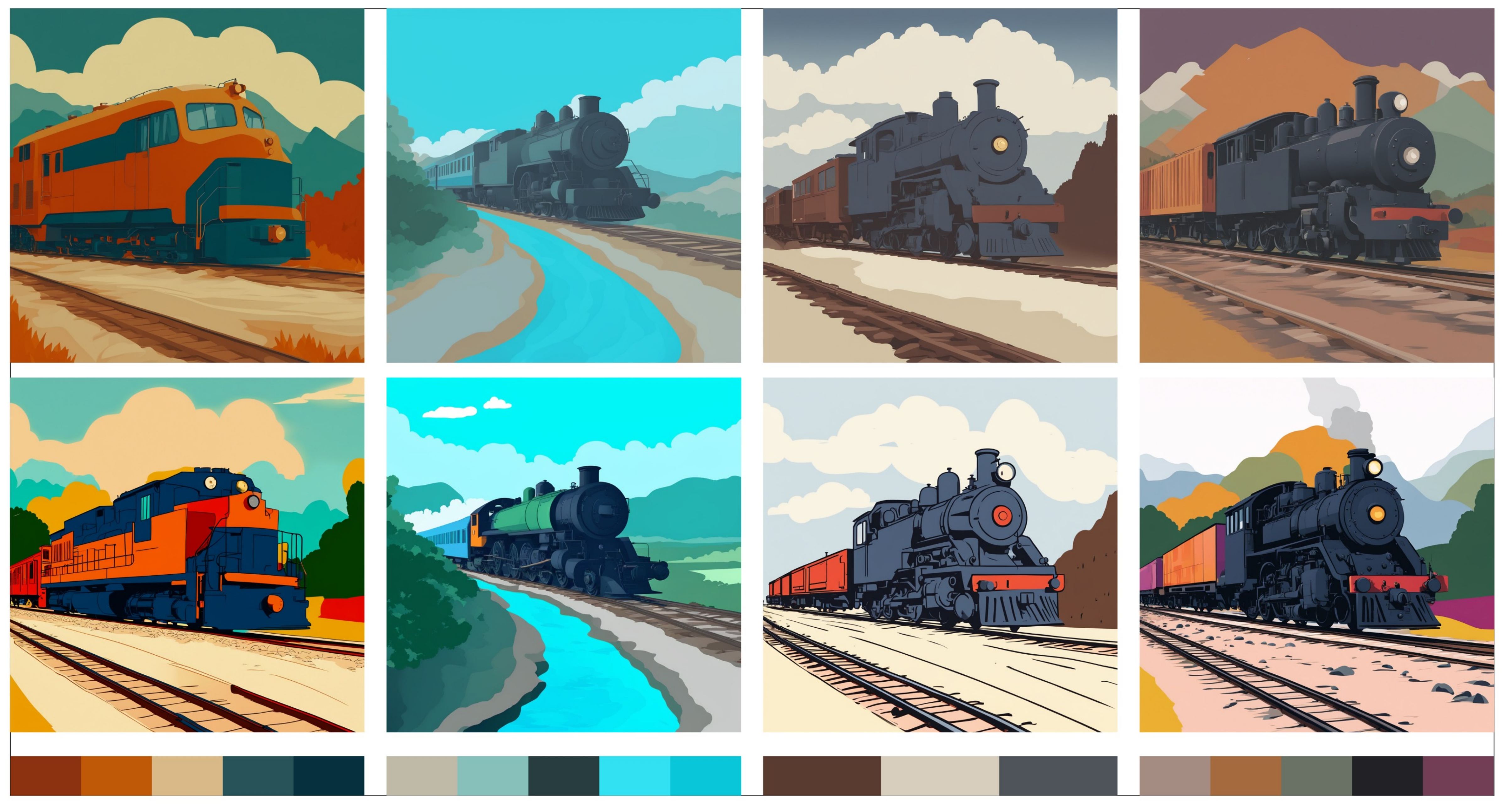}
    \caption{Distance conditioned images, generated with low and high levels of relative distance, and different palettes for the prompt ``A train engine driving through a landscape, flat colors, vector graphic, poster art, 8k.'' and the same seed. Note how as the relative-distance grows in the second row, more hue shades, contrast and details are added.}
    \label{fig:distance}
    \Description{will be added}
\end{figure}

\subsection{Negative Palettes}
\label{sec:CFG}

To control the influence of color conditioning during generation, we employ Classifier-Free-Guidance (CFG) \cite{ho2022classifierfreediffusionguidance}. While traditional CFG uses unconditioned (zero) values for the unconditional denoising step, we found this approach insufficient for strict color alignment. Generated images might acceptably contained colors absent from the input palette, particularly when dealing with prototypical associations (\eg, yellow bananas).

We introduce a novel \emph{negative color condition} approach. Instead of using zero values for unconditional denoising, we allow a manually specified negative palette containing unwanted colors. The predicted noise $\hat{\epsilon}_{\theta}(x_{t},c^\pm,t)$ is calculated using both positive ($c^+$) and negative ($c^-$) palette conditions: 
\begin{equation}
\hat{\epsilon}_{\theta}(x_{t},c^\pm ,t) = 
w\epsilon_{\theta}(x_{t},c^+,t)+(1-w)\epsilon_{\theta}(x_{t},c^-,t)
\end{equation}

\begin{figure}
    \centering
    \includegraphics[width=1\linewidth]{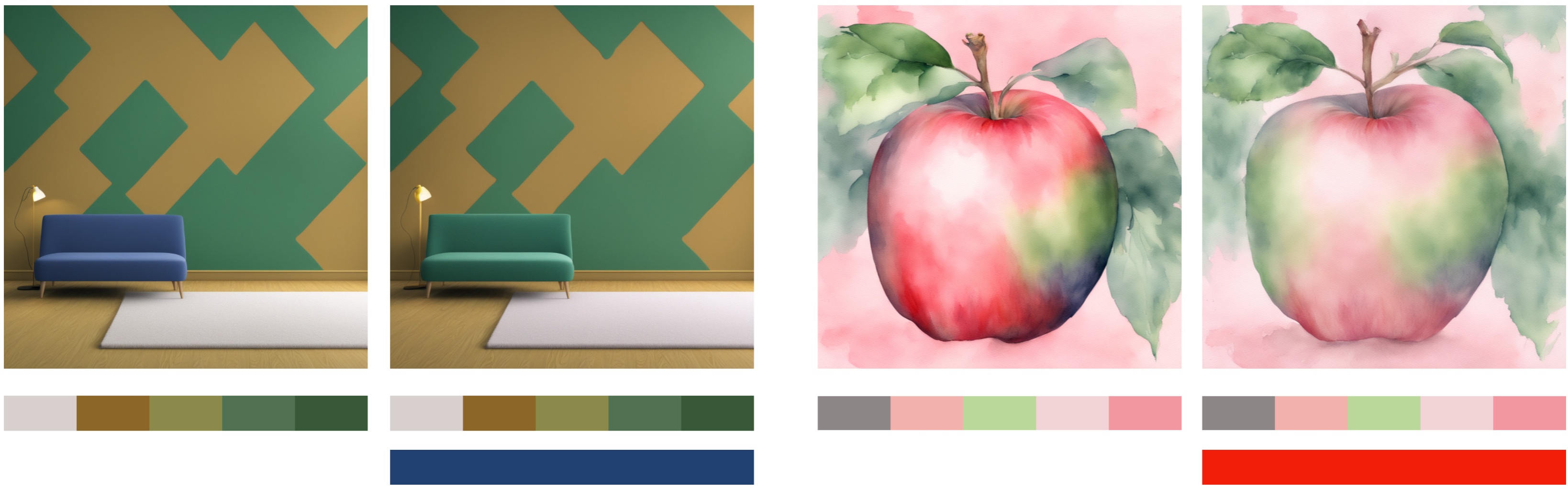}
    \caption{Images generated based on a color palette condition (top color bar) and a manually selected negative condition (bottom color bar). The pair on the left was generated with the prompt ``An interior design showing a sofa, lamp and a wall with a colored wooden pattern'', while the pair on the right was generated with the prompt ``A watercolor painting of one large single centered apple''. We use weighted negative palettes with weights of 0.07 for the sofa, and 0.3 for the apple. A negative palette attenuates or removes the specified color(s). }
    \label{fig:cfg_color_cond_manual}
    \Description{will be added}
\end{figure}

\Cref{fig:cfg_color_cond_manual} demonstrates how manually specifying a negative color palette can improve alignment with the positive condition. We also found it possible to apply weights to the negative palettes to control the level of attenuation. 

\subsection{Distance and Entropy Measures}
\label{sec:entropy}

During training, we use the \emph{distance} between the training image and the input palette or histogram as an additional input. When using a histogram, this distance will be small, while the distance from a palette would be large. At inference time, this allows for additional control over the distribution of colors outside the palette. Larger distances allow more colors outside the palette to appear in the resulting images.

The \emph{entropy} of the color distribution (histogram) can be used as a continuous and effective way to control the level of detail and the color distribution of the generated image. The entropy of the histogram, computed using Shannon's formula~\shortcite{Shannon1948}, effectively controls how much the colors are ``binned''.
If $H$ is the histogram and $p_{i}$ is the probability of bin i:
\begin{equation}Entropy(H)=-\sum_{i=1}^{\#bins} h_i\,log_2\,h_i\end{equation}

Thus, conditioning on low entropy will lead to ``binned'' histogram (flat, monochrome, etc.), and conditioning on high entropy will lead to more distributed colors (gradients, details, etc.). Note that, different from the palette \emph{distance} measure, entropy does not depend on the specific palette and only captures the nature of the distribution.
\Cref{fig:entropy} illustrates images generated for different entropy values for different palettes. Since entropy is a positive measure, to lower the entropy from the unconditioned state, we use CFG with a positive entropy value for the null condition, and a lower value for the positive condition, effectively resulting in a negative ``relative entropy'' (RE) condition.
%Entropy-based conditioning can be done jointly with palette conditioning to provide further control over how the colors are distributed around the conditioned palette bins in a user friendly manner. 

%\begin{figure}
%    \centering
%    \begin{subfigure}[b]{0.115\textwidth}
%        \centering
%        \includegraphics[width=\textwidth]
%        {figures//entropy/golden_gate_entropy_-8.png}
%        \caption{$RE=-8$}
%    \end{subfigure}
%    \begin{subfigure}[b]{0.115\textwidth}
%        \centering
%        \includegraphics[width=\textwidth]
%        {figures//entropy/golden_gate_entropy_0.png}
%        \caption{$RE=0$}
%    \end{subfigure}
%    \begin{subfigure}[b]{0.115\textwidth}
%        \centering
%        \includegraphics[width=\textwidth]
%        {figures//entropy/golden_gate_entropy_3.png}
%        \caption{$RE=3$}
%    \end{subfigure}
%    \begin{subfigure}[b]{0.115\textwidth}
%        \centering
%        \includegraphics[width=\textwidth]
%        {figures//entropy/golden_gate_entropy_5.png}
%        \caption{$RE=5$}
%    \end{subfigure}
%    \caption{Entropy conditioned images, generated with different level of relative entropy ($RE$), and no color condition, for the prompt ``A poster of the Golden Gate at the sunrise. Flat colors. Flat vector sticker style. Minimal palette selection.'' Note how as $RE$ grows, more color shades appear (gradient colors).}
%    \label{fig:entropy}
%\end{figure}

\subsection{Curating The Dataset} 
\label{sec:dataset}

Diffusion models are inherently data-driven, as their inference relies on sampling from a learned distribution. Prior work, \eg,~\cite{perera2023analyzingbiasdiffusionbasedface}, has demonstrated that dataset biases concerning gender, race, and age are reflected in the generated imagery. For colors, most datasets are skewed towards natural images, leading to similar color biases in models trained on them. In general this is a desirable trait, but can become problematic for color-conditioned models, which require sufficiently diverse and balanced data for adequate representation of each color condition embedding (bin). To ensure adequate representation of rare colors, we curate a dataset that contains a more balanced representation of different colors as demonstrated as needed in \Cref{fig:lainon_sdxl_rolor_bias}, and further explained in the supplementary materials in \Cref{sec:dataset}.

% For instance, inspecting 3D color histograms ($8^3$ bins) of LAION-Art~\cite{laion_art} images reveals that the top 100 bins account for 87\% of histogram values, while the 100 rarest bins represent a mere 0.0756\%. This highlights a wide variety of significantly under-represented, or ``rare'', colors (see \Cref{fig:lainon_sdxl_rolor_bias_all_colors,fig:lainon_sdxl_rolor_bias_rare_colors}). \Cref{fig:sdxl_color_bias_dress,fig:sdxl_color_bias_sofa} demonstrate correlation between the color distribution of images generated with minimal and color-neutral prompts (\eg, ``A dress''), using SDXL-1.0 \cite{podell2023sdxlimprovinglatentdiffusion}, compared with the color distribution of the LAION-Art dataset at Figure \ref{fig:lainon_sdxl_rolor_bias_all_colors}.

% To ensure adequate representation of rare colors, we curate a dataset that contains a more balanced representation of different colors. Specifically, we use a 2M images-subset of the LAION-Art dataset, and augment this set with 400K additional images from the LAION-2B-en dataset~\cite{laion2b_en}. These additional image are sampled among those that contain colors from the low valued histogram bins of the LAION-Art dataset. This results in a total of 2.4M images.

\section{Experiments}
\label{sec:experiments}

%\subsection{Experiment Setup}
Our model is based on SDXL 1.0~\cite{podell2023sdxlimprovinglatentdiffusion} and is trained on a curated dataset of 2.4 million images, as detailed in \Cref{sec:dataset} of the supplemental  materials. All images were resized to a resolution of 512x512 pixels for training. We adopted an approach similar to that of \citet{ye2023ip-adapter}, incorporating drop probabilities as specified in \Cref{table:cfg_dropout} to enable Classifier-Free Guidance (CFG), utilizing an empty string for text prompts and zero tensors for other conditioning. 

In the following sections, we present both qualitative and quantitative results, including a user study.

\subsection{Qualitative Results}
\label{sec:qualitative}

\begin{figure}
    \centering
    \begin{subfigure}[b]{0.47\textwidth}
        \centering
        \includegraphics[width=\textwidth]
        {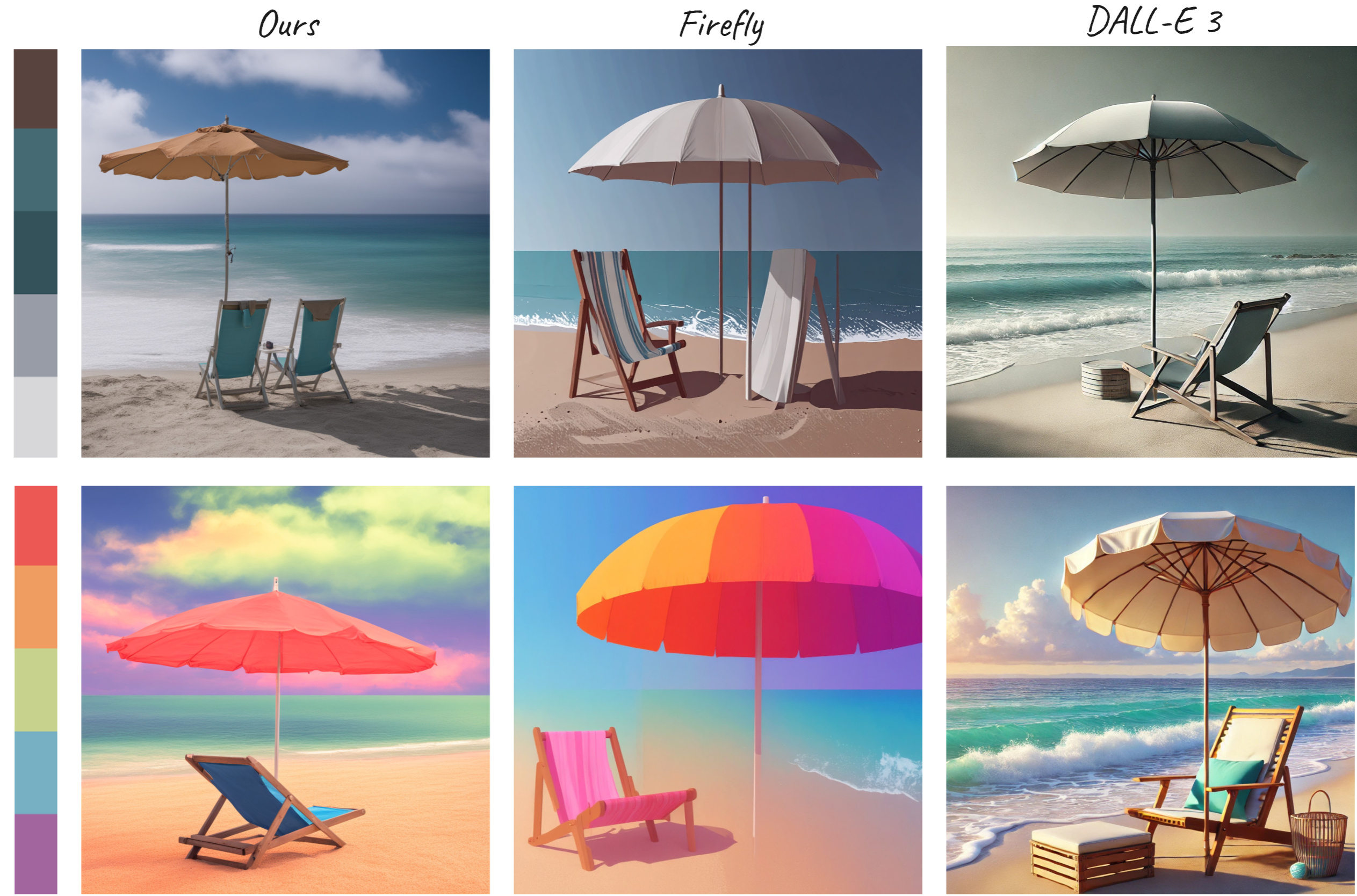}
         \caption{prompt=``A beach chair under an umbrella on the beach.''.}
    \end{subfigure}
    \begin{subfigure}[b]{0.47\textwidth}
        \centering
        \includegraphics[width=\textwidth]
        {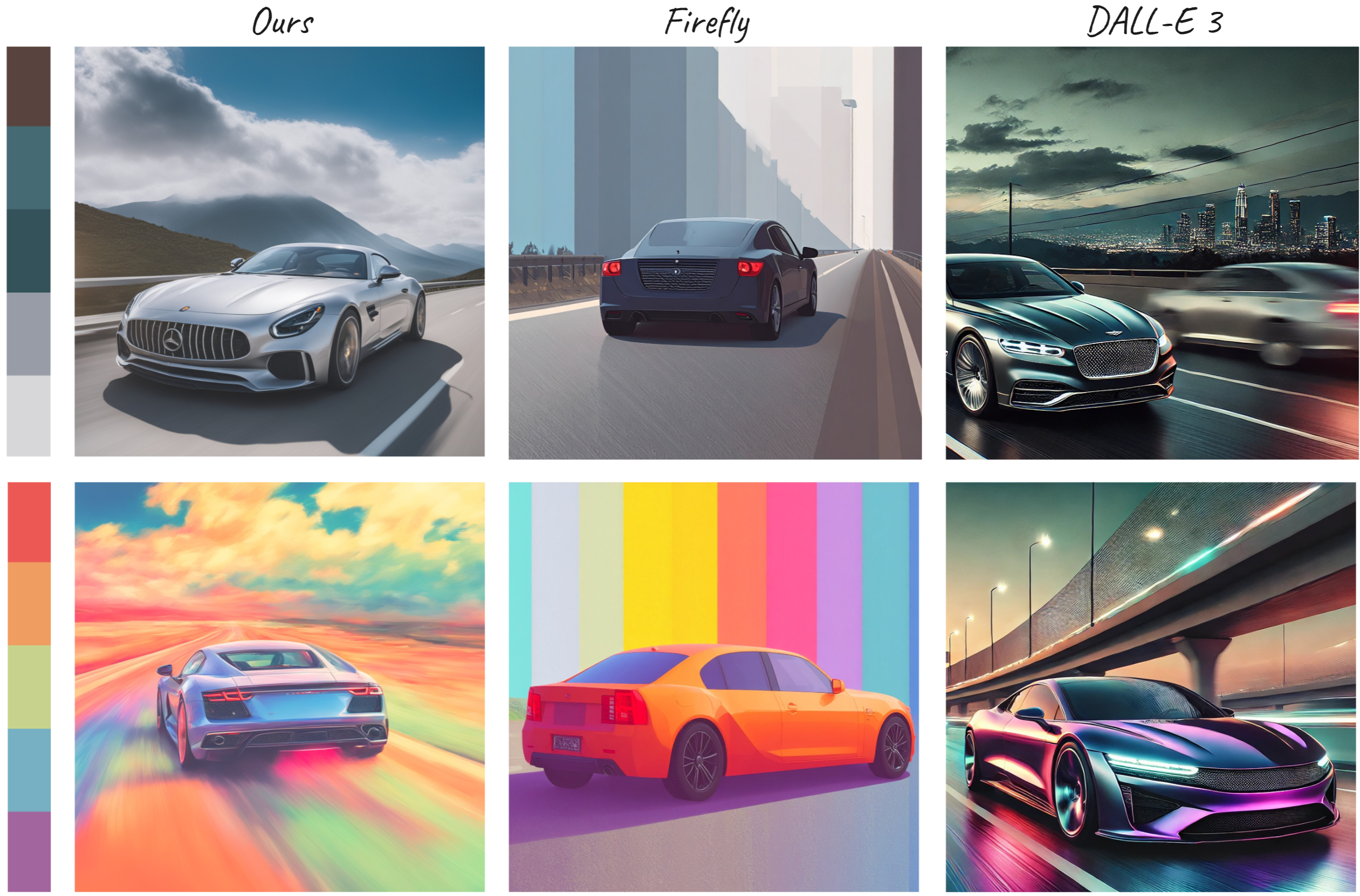}
         \caption{prompt=``An expensive car driving the highway.''.}
    \end{subfigure}
    \caption{\textbf{Palette based recoloring comparison between our method to Firefly and Dall-E 3}:    
    prompt: ``A vintage toy jeep with a soft top roof and open sides, placed on a neutral surface. Our model generates better palette alignment without artifacts.''}
    \label{fig:firefly}
    \Description{will be added}
\end{figure}

The results generated by our method are demonstrated in figures
throughout the paper. Unless explicitly stated, images in this paper generated \emph{without} using the negative palette, entropy and distance controls.
In particular, \Cref{fig:teaser}, \Cref{fig:entropy} and \Cref{fig:distance} demonstrate the core capabilities of our method: fine-grained control (entropy, distance, style) and in particular how different palettes for the same prompt yield dramatically different results, closely matching the conditioning palettes. Additional results may be found in the supplemental material. We show comparison with other method in \Cref{fig:comparison}. Additional comparison results are found in \Cref{sec:appendix_comparison} of the supplemental material including \Cref{fig:comparison_cont}. For completeness, we also include comparison with other methods that we found less suitable or without an available model, such as Composer in ~\Cref{fig:recoloring_composer}, Adobe Firefly and DALL·E 3 in \Cref{fig:firefly}. 

\begin{figure}
    \centering
    \begin{subfigure}[b]{0.47\textwidth}
        \centering
        \includegraphics[width=\textwidth]
        {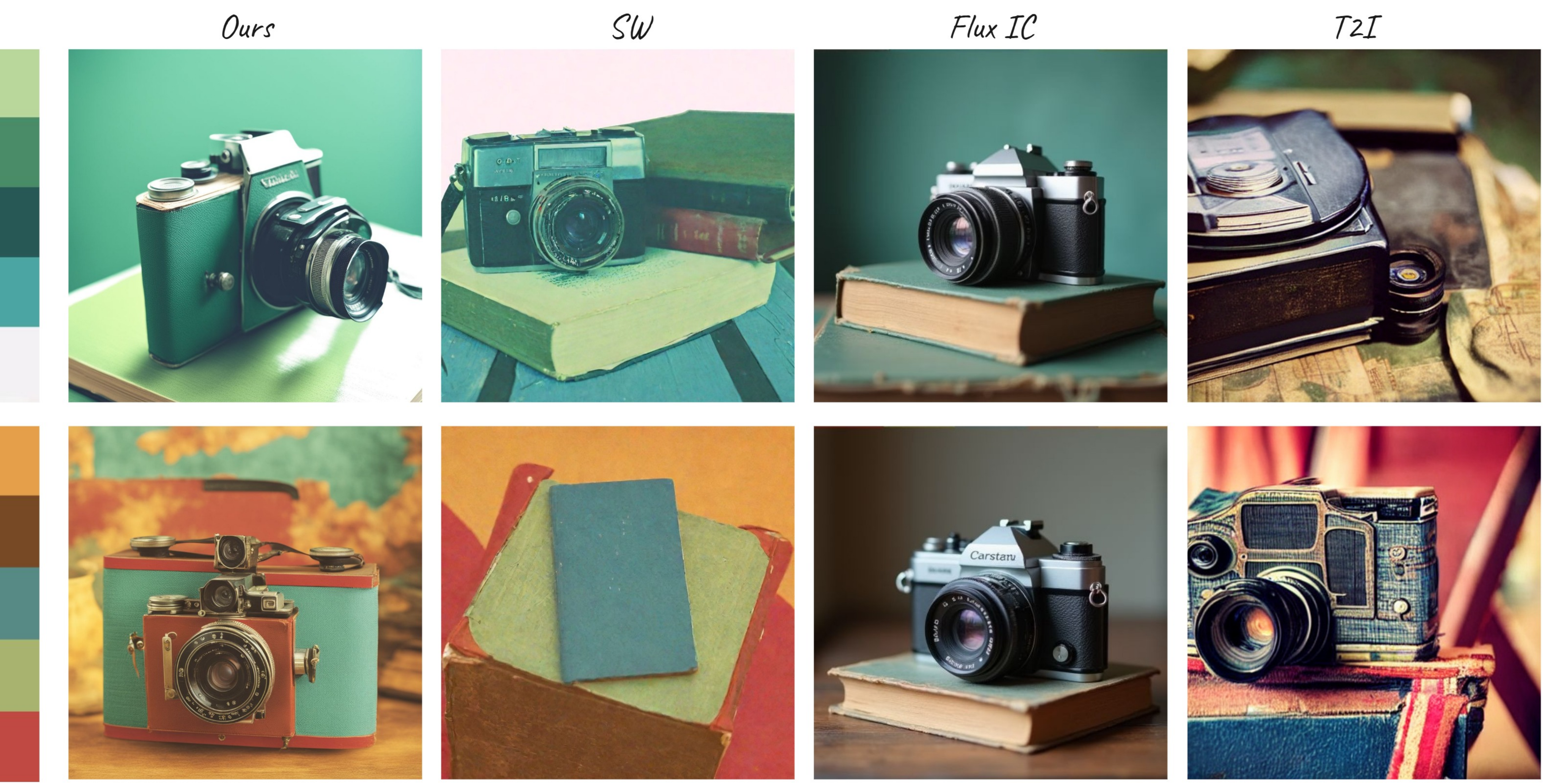 }
        \caption{prompt=``A vintage camera resting on an old book''.}
    \end{subfigure}
    \begin{subfigure}[b]{0.47\textwidth}
        \centering
        \includegraphics[width=\textwidth]
        {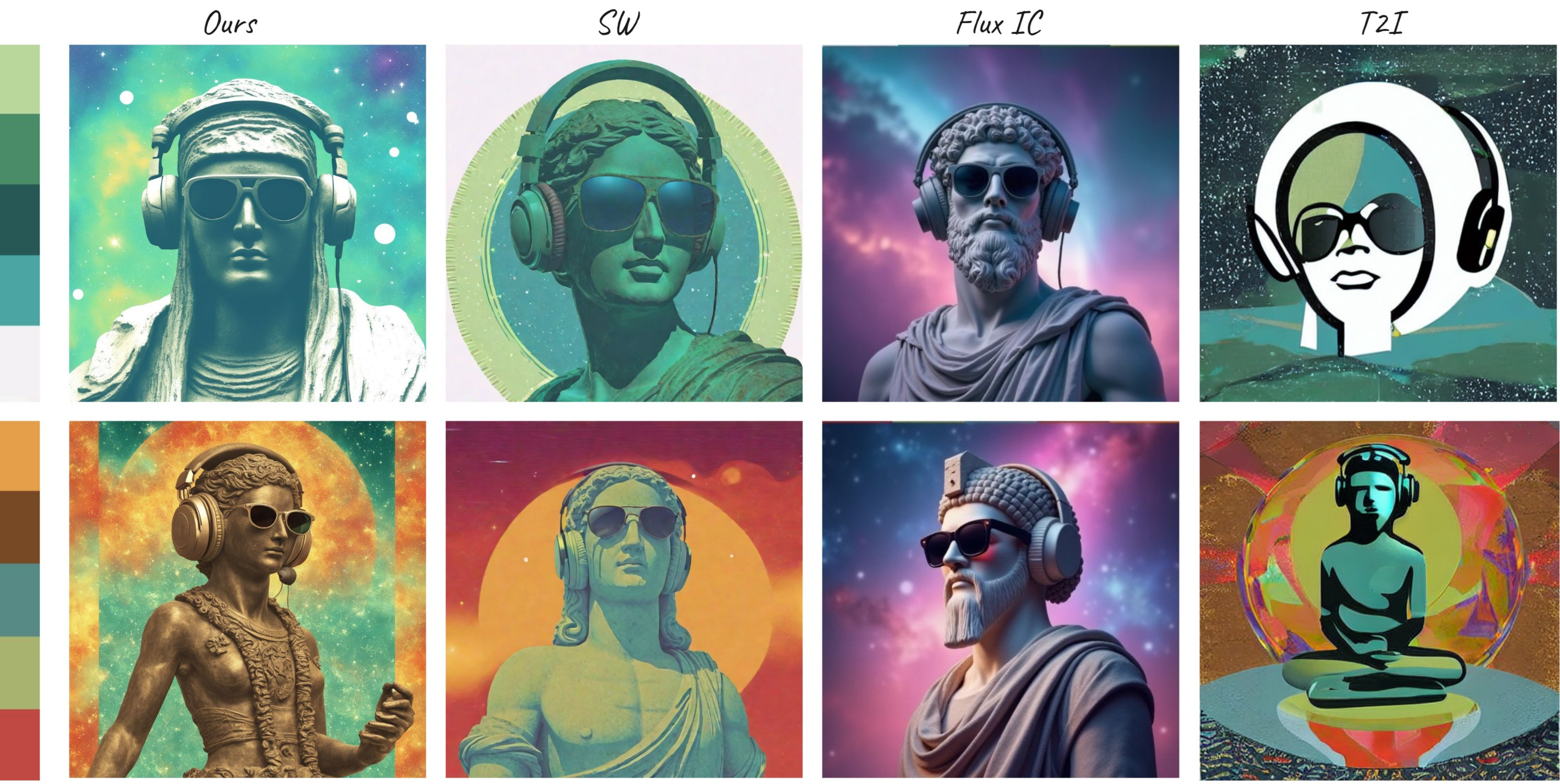 }
        \caption{prompt=``A poster design of an ancient statue in headphones and sunglasses on a cosmic background''.}
    \end{subfigure}
    \caption{\textbf{Comparison of palette based generation between our method, 2TI, SW and FLUX IC. Our methods produces high fidelity results, adhering to both prompt and palette. }}
    \label{fig:comparison}
\end{figure}

\subsection{Quantitative Evaluation}
\label{sec:exp_quantitative}

We compared our model against three palette-conditioned generative models: T2I-Adapter and FLUX IC Palette \cite{GeorgeQi_Color_Palette_Flux_dev_09dd190} which are training-dependent methods, and SW-Guidance \cite{lobashev2025colorconditionalgenerationsliced}, which is a training-free method. These models directly address the problems of palette-conditioned generation without being confounded by unrelated style transfer techniques, ensuring a fair and focused comparison. Some examples can be seen in \Cref{fig:comparison}. We also included the original SDXL 1.0 as a baseline.

Following \cite{lobashev2025colorconditionalgenerationsliced}, we evaluated these methods on the COCO 2017 validation set, initially containing 5000 images. To ensure that color conditioning is strictly driven by the palette provided, we filtered this set to images whose captions do not mention colors and used the first remaining 1000 images. Since FLUX IC, and SW-Guidance require 1D color palettes, we used Pylette with K-means to obtain 5 primary colors of each ground-truth image. For T2I Adapter, we initially generated 2D palettes by down-sampling to 8x8 images, and subsequently up-sampling (to 512x512) the ground-truth image using nearest-neighbor interpolation, to match the authors' instructions. Recognizing the potential for inconsistencies to the extracted 1D palettes (which could negatively effect the T2I scores), we have implemented an optimal transport-based alignment between the downsampled image and the target palette using EMD. Eventually, each model received the same captions as text prompts, same seed, and the extracted palettes (1D or 2D) for generations, resulting in 1000 generated images per model.

To comprehensively evaluate the performance of our model and the baselines, we employed three distinct metrics, each targeting a specific aspect of the generation process: color alignment to the reference palette, image quality, and semantic adherence to the text prompt. For color alignment, we used the EMD, which effectively quantifies the perceptual dissimilarity between the color distributions of the generated images and the provided palettes. To assess the visual quality of the generated images without relying on references, we utilized MusiQ~\cite{50723}, a state-of-the-art No-Reference Image Quality metric developed by Google. Finally, to measure the semantic alignment between the generated images and their corresponding text prompts, we calculated the CLIP similarity score, commonly used to assess prompt alignment. 

The results of the comparison are reported in \Cref{table:image_quantitative}.
Although SW-Guidance provides the lowest EMD score, it achieves it by strongly optimizing against a sliced Wasserstein objective during inference. This strong emphasis on color distance appears to come at the cost of image quality, resulting in a lower EMD than observed in manually extracted palettes (see \Cref{table:image_quantitative}). SW scores the lowest on both MusiQ metrics, Ava and Koniq, designed to measure aesthetic appeal and perceptual quality, respectively. This indicates that the generated images suffer from visual deficiencies, distortions, and a lack of aesthetic appeal. In contrast, this work aimed to balance these competing factors, a goal our model effectively achieves. While not reaching top color accuracy, it delivers the highest perceptual quality, demonstrating successful color palette integration without sacrificing visual coherence.

\begin{table}[h!]
\scriptsize
\begin{tabular}{c|c|c|c|c}
 Category & EMD $\mu\pm\sigma$ & Koniq $\mu\pm\sigma$ & AVA $\mu\pm\sigma$ & CLIP $\mu\pm\sigma$\\ 
 \hline
 Our model & 0.1414 $\pm$ 0.031 & 70.642 $\pm$ 5.789 & 5.619 $\pm$ 0.392 & 0.311 $\pm$ 0.028 \\
 SW & 0.063 $\pm$ 0.019 & 63.628 $\pm$ 8.251 & 4.997 $\pm$ 0.523 & 0.311 $\pm$ 0.032 \\
 FLUX IC & 0.228 $\pm$ 0.071 & 66.429 $\pm$ 7.401 & 5.385 $\pm$ 0.615 & 0.309 $\pm$ 0.027\\
 T2I (EMD) & 0.1903 $\pm$ 0.06 & 68.008 $\pm$ 9.077 & 5.134 $\pm$ 0.601 & 0.308 $\pm$ 0.029 \\
 \hline
 Baseline SDXL 1.0  & 0.257 $\pm$ 0.089 & 72.118 $\pm$ 5.08 & 5.647 $\pm$ 0.409 & 0.316 $\pm$ 0.027 \\
 \hline
 Reference*  & 0.170 $\pm$ 0.089 & - & - & - \\
 \end{tabular}
\caption{ Average distance of generated images from the conditioning palettes. *The reference distance is the average distance of a set of human-extracted palettes taken from \citet{picmonkeyColorCombinations} and is used as a baseline, and is not computed based on COCO prompts and palettes as all others.}
\label{table:image_quantitative}
%\elad{I didn't generate multiple images just used the ones from the user study since I didn't have time yet, plust the numbers already make sense so maybe it's better to leave as is.}
\vspace{-3em}
\end{table}

\subsection{User Study}
\label{sec:exp_user_study}
We conducted a user study to evaluate our model's generated images for perceptual color alignment with given palettes and for overall quality. Participants ranked the match between a palette and an image on a 1-5 Likert scale. A higher score indicated that all palette colors were present in the image and all image colors were close to a palette color. Participants also rated the overall quality of each image, considering the provided palette.

The study used 40 high-quality images and their \emph{human-extracted} palettes, randomly selected from \cite{picmonkeyColorCombinations}, a set curated by designers. For each, we created a color-neutral descriptive prompt (e.g., ``A photo of a bird of paradise flower.''). Crucially, all compared methods generated images using the same seed for each prompt. Our method was used without our fine-grained controls (entropy, distance, or negative color prompting), which can further improve palette adherence.

For each palette and prompt, participants evaluated images corresponding to five conditions: the original reference image, and the image generated using our method, Flux IC Palette, T2I, and SW. They rated color alignment and overall quality for each of the resulting 200 images (40 palettes, for 4 models and a reference). These tasks were divided into four forms, each containing 10 palette sections. Forty participants (aged 20-40, from diverse backgrounds, including designers and researchers) each completed one form.

The results are presented in Table \ref{table:user_study}. Our model's color alignment score (3.87) is very close to SW's (3.97), while our mean quality score was notably higher (3.44 vs. 2.88), despite both models being based on SDXL 1.0 architecture. Our model's palette alignment also ranks higher than FLUX IC, and T2I. The harmonic mean of both scores, along with the Pareto front (see \Cref{fig:user_study_graph} in the supplemental material), shows that our model strikes the best balance between adherence to the palette and quality. As expected, Flux IC Palette, utilizing a newer architecture and larger model, achieved higher overall quality, and all models trailed behind the high-quality real reference images.

\begin{table}[h!]
\footnotesize
\begin{tabular}{c|c|c|c}
 Model & Palette Alignment $\mu$ $\pm$ $\sigma$ & Quality $\mu$ $\pm$ $\sigma$ & $\mu$ Harmonic Mean\\ 
 \hline
 Our Model & 3.873 $\pm$ 1.06 & 3.447 $\pm$ 1.10 & \textbf{3.647}\\
 FLUX IC  & 2.969 $\pm$ 1.24 & 3.867 $\pm$ 1.10 & 3.359\\
 2TI & 3.072 $\pm$ 1.20 & 3.127 $\pm$ 1.22 & 3.099\\
 SW & 3.978 $\pm$ 1.08 & 2.884 $\pm$ 1.23 & 3.344\\
 \hline
 Reference & 3.934 $\pm$ 1.06 & 4.230 $\pm$ 0.938 & 4.076\\
 \end{tabular}
\caption{User study results: users rank palette-image matching, and overall quality based on reference taken from \cite{picmonkeyColorCombinations}.}
\label{table:user_study}
\vspace{-3em}
\end{table}

% \paragraph{Entropy Conditioning.} \Cref{fig:palette_and_entropy_styles} and \Cref{fig:palette_and_entropy_photo} demonstrate the simplicity of mixing color and entropy conditions for controlling the level of details, and color variance with minimal effort, and across different styles. Notably, when increasing the relative entropy for the same palette condition, the model uses more shades of the existing hues, but nevertheless preserves the primary palette colors. To allow even more freedom, distance conditioning may be used.  

% \paragraph{Distance Conditioning.}
% During training, larger distances were used for palettes than for histograms, as the distance from each palette condition, represented as a sparse histogram, to the full histogram is typically considerable. This directs the model to learn that more colors can be used outside the palette to match the input image. 
% Hence, the model learns that more color-choice freedom is possible for large distances. This provides another conditioning control during inference -- the use of large or small distances for more or less color freedom, respectively. \Cref{fig:palette_and_distance} demonstrates the use of varying distances with the same prompt, entropy and palette. It may be seen that increasing the distance introduces new colors and increases the contrasts in the generated images.

%\elad{FIx the teaser to use different palettes, specifically the first two. The table with the styles, not flowers.., and different palettes,}

\section{Summary and Discussion}
\label{sec:discussion}

Our method enables the generation of images conditioned on color palettes by incorporating a novel Palette-Adapter into an existing frozen text-to-image diffusion model. While color palettes offer a compact and standard means of communicating color schemes, they are inherently ambiguous and loosely defined. To address this challenge, we trained our adapter using a combination of color palettes and histograms, while introducing novel controls such as distance, entropy, and negative color conditioning.  We demonstrated the effectiveness of our approach through a variety of experiments and a user study.

%Our main contribution are:
%\begin{enumerate}
%  \item We demonstrate the ability to add color palette conditioning to an existing model using an IP Adapter.
%  \item We augment palettes as sparse color histogram, and mix them with full histograms during training to mitigate the combinatorial complexity of the palette space during data collection, palette extraction and training.
%  \item We introduce the use of a \emph{negative} color condition to improve the alignment with the \emph{positive} color condition.
%  \item We introduce the use of color \emph{entropy} for guidance, and demonstrate the effectiveness of using entropy both implicitly and explicitly.
%  \item We demonstrate correlation in color bias between datasets and generated images, and hence the importance of adequate data sampling of rare colors.
%  extraction.
%\end{enumerate}

\subsection{Limitations and Future Work.}
Our manual experiments reveal two limitations of the current approach.
Firstly, we observe a trade-off between text and color alignment as indicated in \Cref{sec:exp_quantitative} when comparing our method against the baseline SDXL 1.0. This means that not all prompts are as effective with every color palette, hindering the model's ability to consistently fulfill user requests in some cases. This could be compensated using a negative prompt, or trying different seeds. 

%Our experiments encoded color palettes as histograms by projecting them into $8^3$ uniformly spaced bins in RGB, which is lacking both in terms of numerical accuracy and perceptual uniformity of hues and saturation levels. Future work could explore finer-grained histograms, as well as non-linear binning or use of other color spaces such as LAB to address this.

Another limitation is the fact that the scales of the relative entropy and histogram distance measures are somewhat overlapping. Future work can try to produce a more user-friendly version for controlling the palette adherence.

We observe that some highly saturated colors still remain under-represented in our training data, potentially leading to biases towards non-photorealistic outputs when included in the guidance palette.  We plan to investigate the incorporation of synthetic data and additional histogram augmentation techniques, such as sharpening, to mitigate this issue.

While our method enables the specification of a negative color condition, this condition is currently specified manually and could potentially be provided automatically, if instructed so by the user. For example, by applying early decoding of the latent image at a few points during the diffusion process and inspecting significant color deviations.

We believe that our palette-aligned generation method provides a valuable tool for artists and designers.  Therefore, we plan to release our code and model weights to facilitate both practical utilization and future research in this area. 

%\newpage

\bibliographystyle{ACM-Reference-Format}
\bibliography{main-siggraph-hsv}

\clearpage

\newpage
\begin{figure}
    \centering
    \begin{subfigure}{1.0\columnwidth}
        \centering
        \begin{subfigure}[b]{0.155\textwidth}\centering\includegraphics[width=\textwidth]{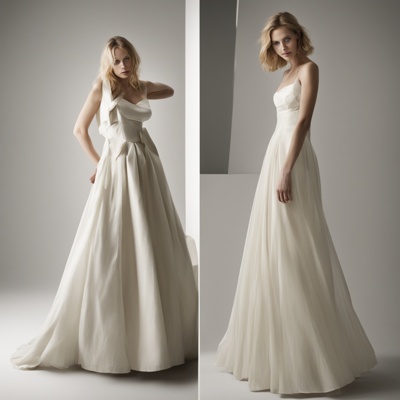}\end{subfigure}
        \begin{subfigure}[b]{0.155\textwidth}\centering\includegraphics[width=\textwidth]{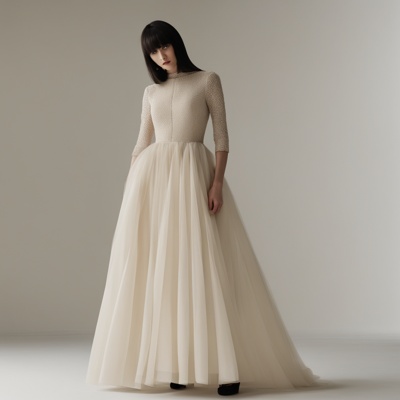}\end{subfigure}
        \begin{subfigure}[b]{0.155\textwidth}\centering\includegraphics[width=\textwidth]{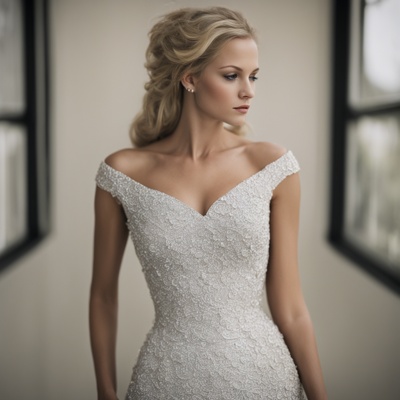}\end{subfigure}
        \begin{subfigure}[b]{0.155\textwidth}\centering\includegraphics[width=\textwidth]{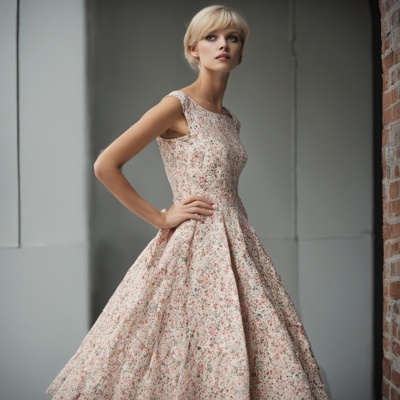}\end{subfigure}
        \begin{subfigure}[b]{0.155\textwidth}\centering\includegraphics[width=\textwidth]{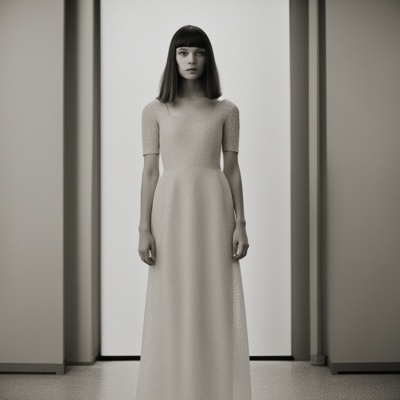}\end{subfigure}
        \begin{subfigure}[b]{0.155\textwidth}\centering\includegraphics[width=\textwidth]{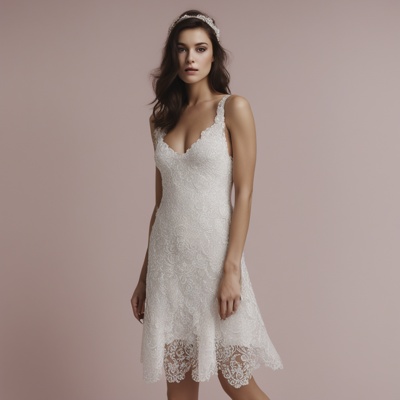}\end{subfigure}
        \begin{subfigure}[b]{0.155\textwidth}\centering\includegraphics[width=\textwidth]{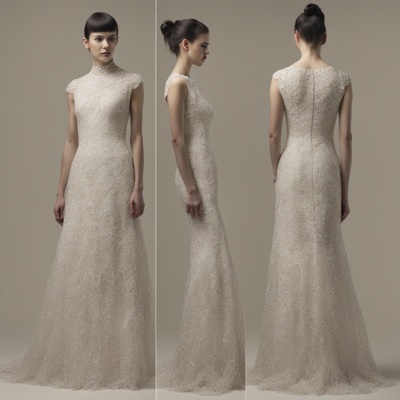}\end{subfigure}
        \begin{subfigure}[b]{0.155\textwidth}\centering\includegraphics[width=\textwidth]{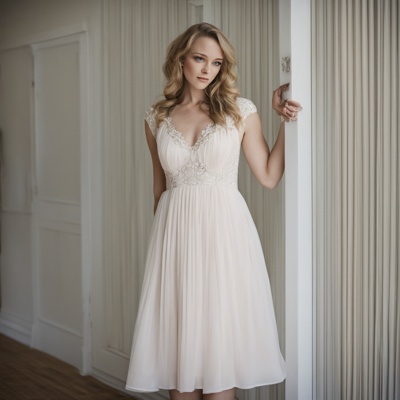}\end{subfigure}
        \begin{subfigure}[b]{0.155\textwidth}\centering\includegraphics[width=\textwidth]{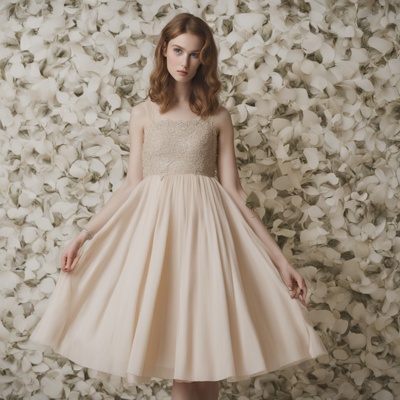}\end{subfigure}
        \begin{subfigure}[b]{0.155\textwidth}\centering\includegraphics[width=\textwidth]{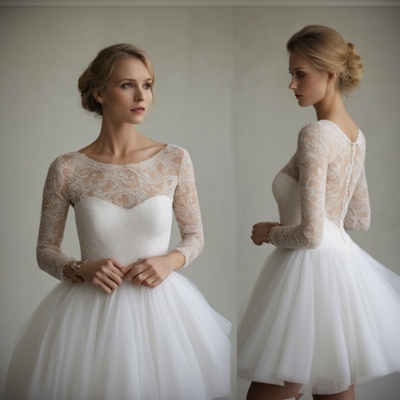}\end{subfigure}
        \begin{subfigure}[b]{0.155\textwidth}\centering\includegraphics[width=\textwidth]{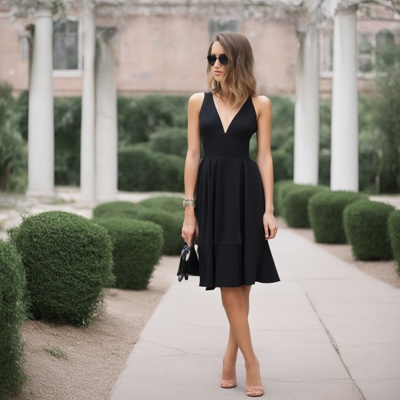}\end{subfigure}
        \begin{subfigure}[b]{0.155\textwidth}\centering\includegraphics[width=\textwidth]{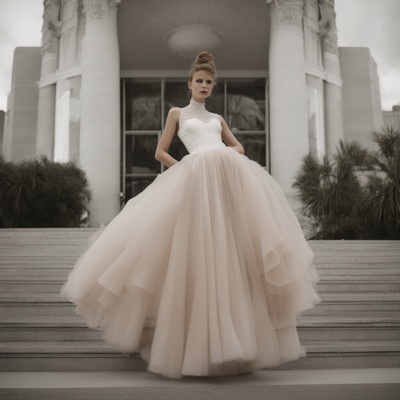}\end{subfigure}
        \caption{Images generated with different seeds for the prompt ``A dress''.}
        \label{fig:sdxl_color_bias_dress}
    \end{subfigure}
    \begin{subfigure}{1.0\columnwidth}
        \centering
        \begin{subfigure}[b]{0.155\textwidth}\centering\includegraphics[width=\textwidth]{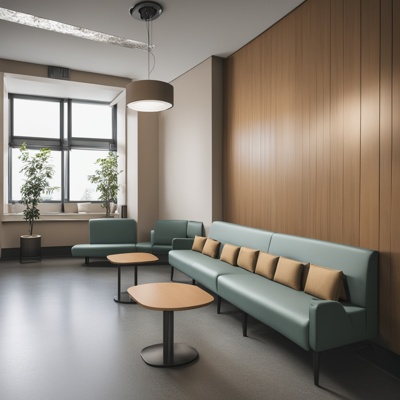}\end{subfigure}
        \begin{subfigure}[b]{0.155\textwidth}\centering\includegraphics[width=\textwidth]{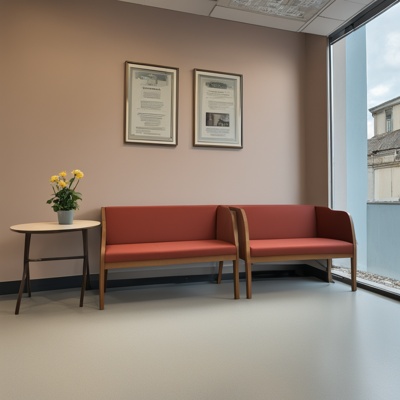}\end{subfigure}
        \begin{subfigure}[b]{0.155\textwidth}\centering\includegraphics[width=\textwidth]{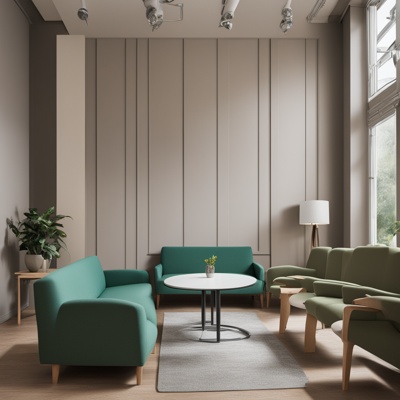}\end{subfigure}
        \begin{subfigure}[b]{0.155\textwidth}\centering\includegraphics[width=\textwidth]{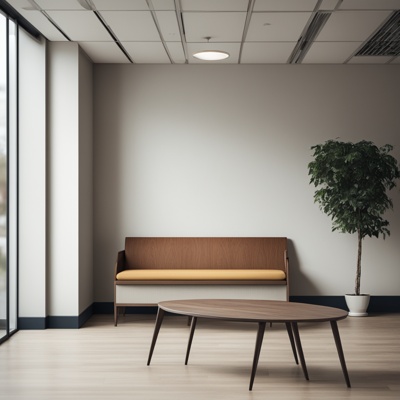}\end{subfigure}
        \begin{subfigure}[b]{0.155\textwidth}\centering\includegraphics[width=\textwidth]{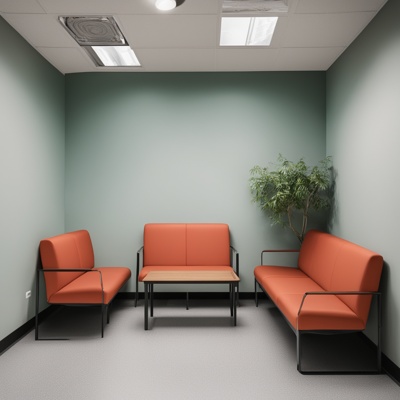}\end{subfigure}
        \begin{subfigure}[b]{0.155\textwidth}\centering\includegraphics[width=\textwidth]{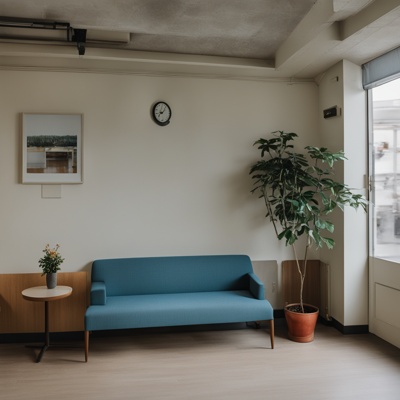}\end{subfigure}
        \begin{subfigure}[b]{0.155\textwidth}\centering\includegraphics[width=\textwidth]{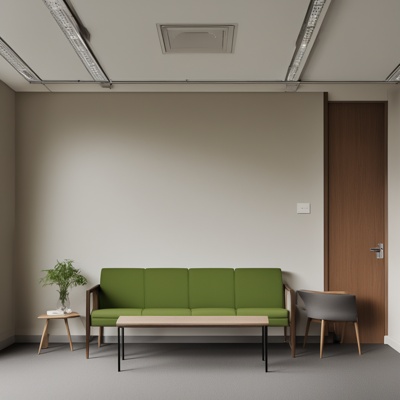}\end{subfigure}
        \begin{subfigure}[b]{0.155\textwidth}\centering\includegraphics[width=\textwidth]{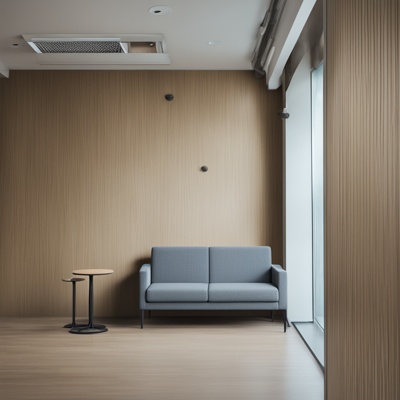}\end{subfigure}
        \begin{subfigure}[b]{0.155\textwidth}\centering\includegraphics[width=\textwidth]{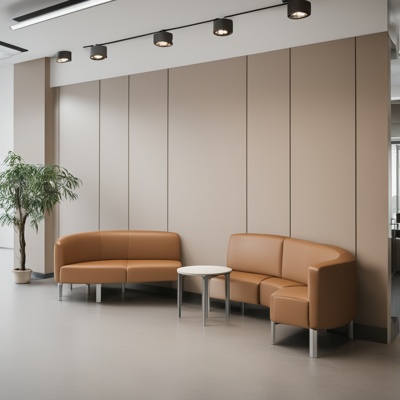}\end{subfigure}
        \begin{subfigure}[b]{0.155\textwidth}\centering\includegraphics[width=\textwidth]{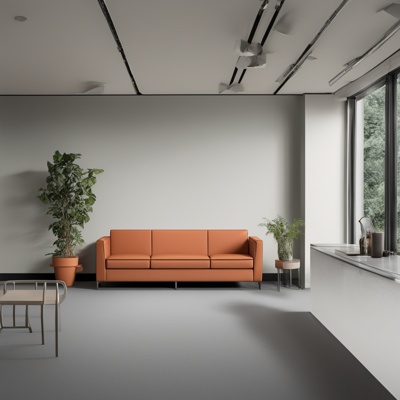}\end{subfigure}
        \begin{subfigure}[b]{0.155\textwidth}\centering\includegraphics[width=\textwidth]{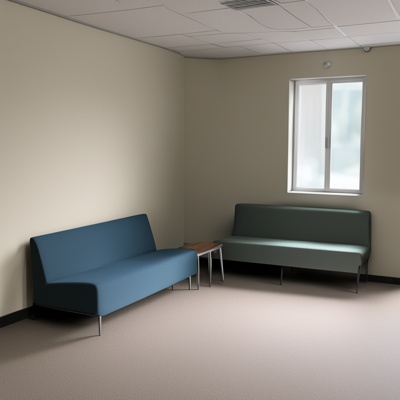}\end{subfigure}
        \begin{subfigure}[b]{0.155\textwidth}\centering\includegraphics[width=\textwidth]{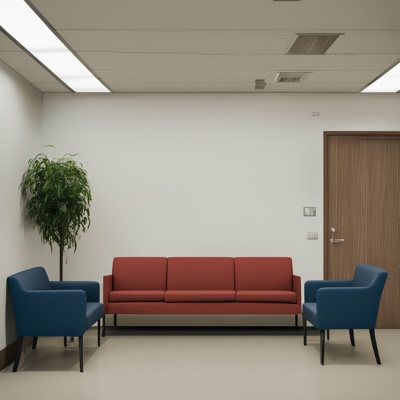}\end{subfigure}
        \caption{Random images for ``A waiting room with a sofa and a table.''}
        \label{fig:sdxl_color_bias_sofa}
    \end{subfigure}
    \begin{subfigure}[b]{1.0\columnwidth}
        \centering
        \includegraphics[width=\textwidth]{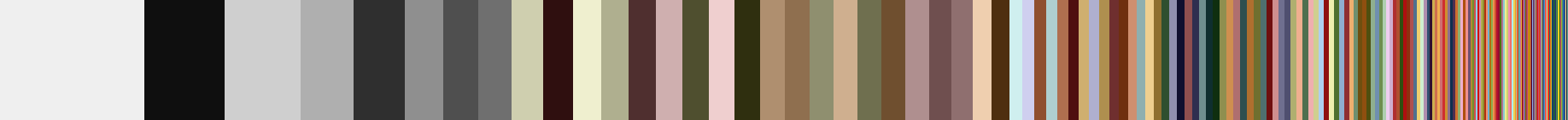}
        \caption{Color distribution in the LAION-Art dataset (sorted by probability).}
        \label{fig:lainon_sdxl_rolor_bias_all_colors}
    \end{subfigure}
    \begin{subfigure}[b]{1.0\columnwidth}
        \centering
        \includegraphics[width=\textwidth]{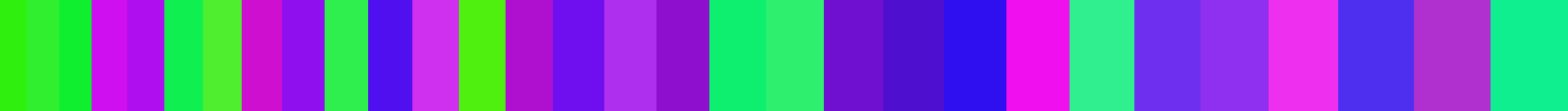}
        \caption{The 30 rarest colors in LAION-Art.}
        \label{fig:lainon_sdxl_rolor_bias_rare_colors}
    \end{subfigure}
    \caption{The colors in random images generated by SDXL 1.0 using minimal color-neutral prompts (a, b) reflect well the color distribution in the LAION-Art dataset \cite{laion_art} (c).
    The 100 most common colors in LAION-Art account for 87\% of the total population, while the 30 rarest colors (d) account for only 0.008253\%.}
    \label{fig:lainon_sdxl_rolor_bias}
    \Description{will be added}
\end{figure}

\begin{figure}
    \centering
    \includegraphics[width=1\linewidth]{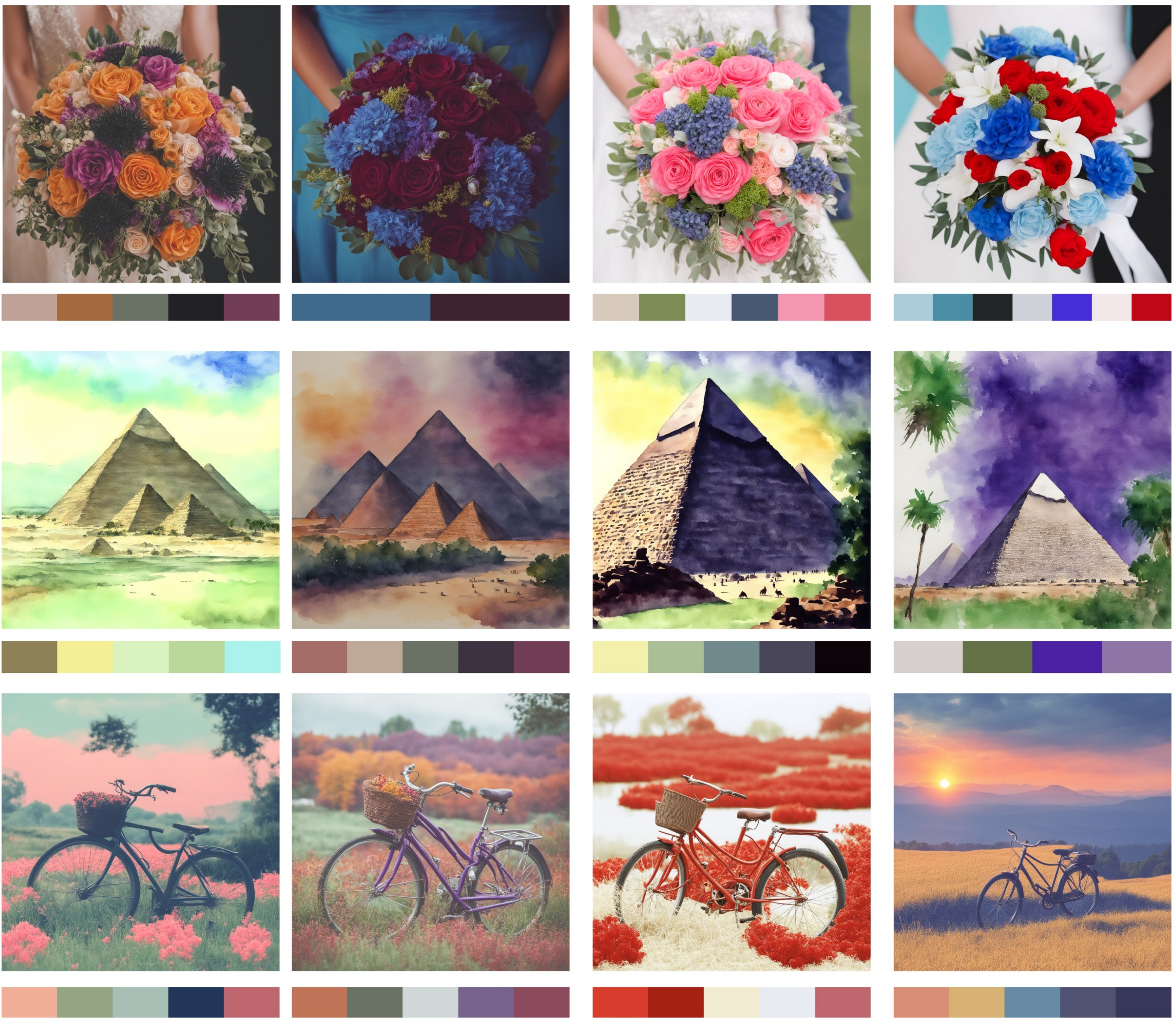 }
    \caption{A collection of images generated using similar prompts per row and different palettes}
    \label{fig:portraits}
    \Description{will be added}
\end{figure}

\begin{figure}
    \centering
    \includegraphics[width=1\linewidth]{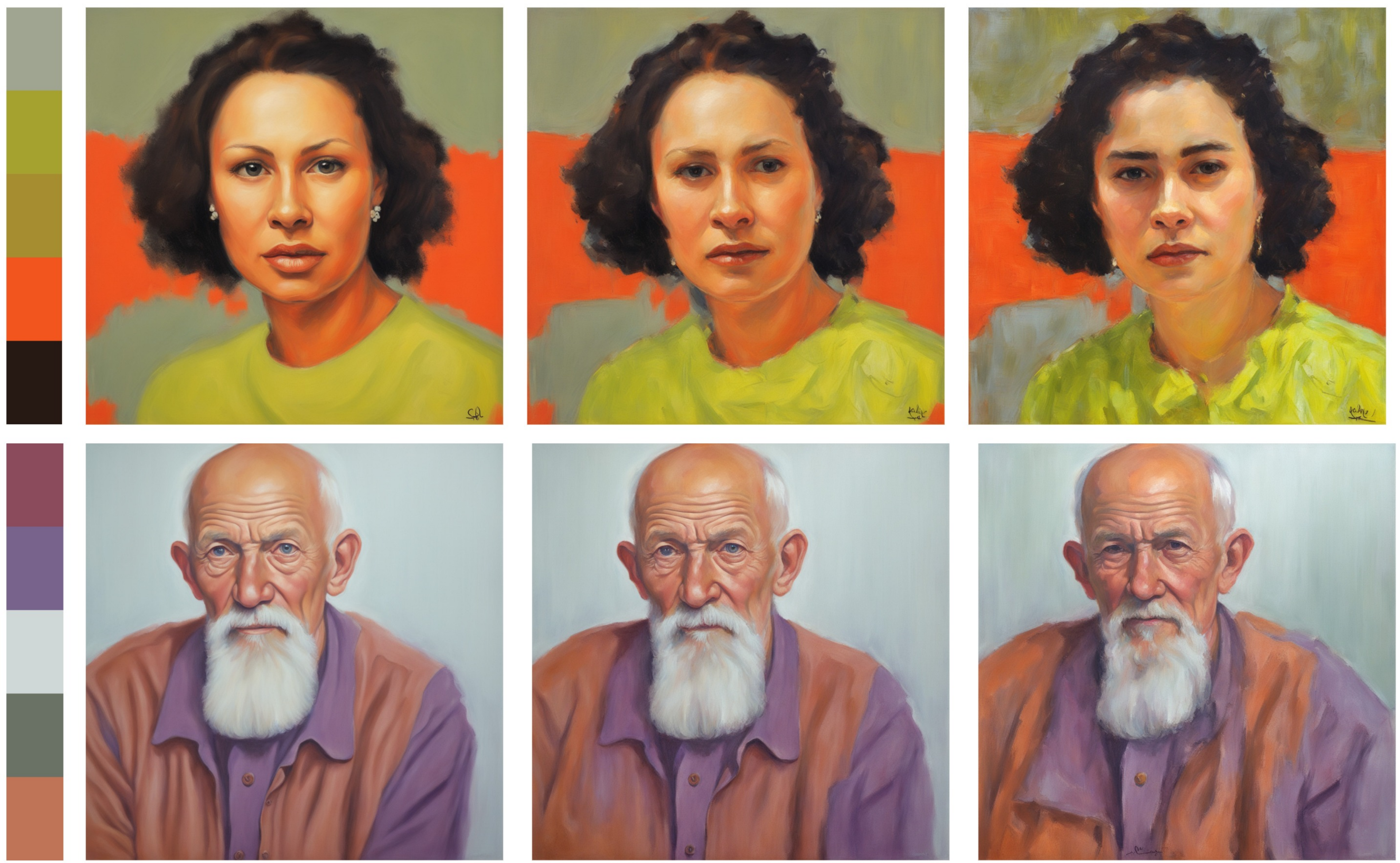}
    \caption{Artwork images generated based on a color condition and different levels of relative entropy guidance. Higher entropy introduces more details and texture.}
    \label{fig:palette_and_entropy_styles}
    \Description{will be added}
\end{figure}

\begin{figure}
    \centering
    \includegraphics[width=1\linewidth]{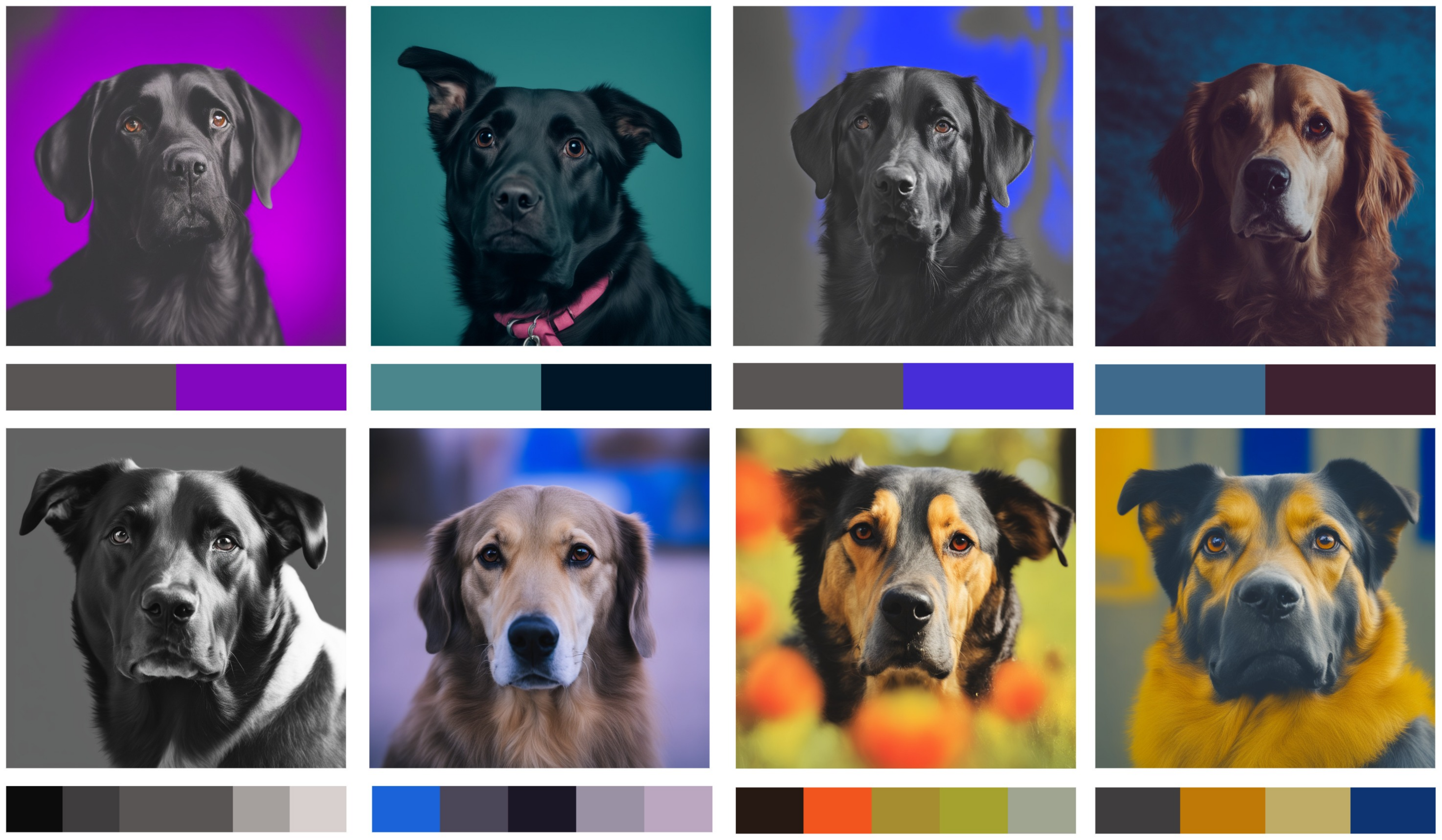}
    \caption{\textbf{Limited and typical palettes}: images generated using different palettes with the same seed and prompt ``A high quality photo portrait of a single centered dog. 8K. DSLR.''}
    \label{fig:limited_and_typical}
    \Description{will be added}
\end{figure}

\begin{figure}
    \centering
    \includegraphics[width=1\linewidth]{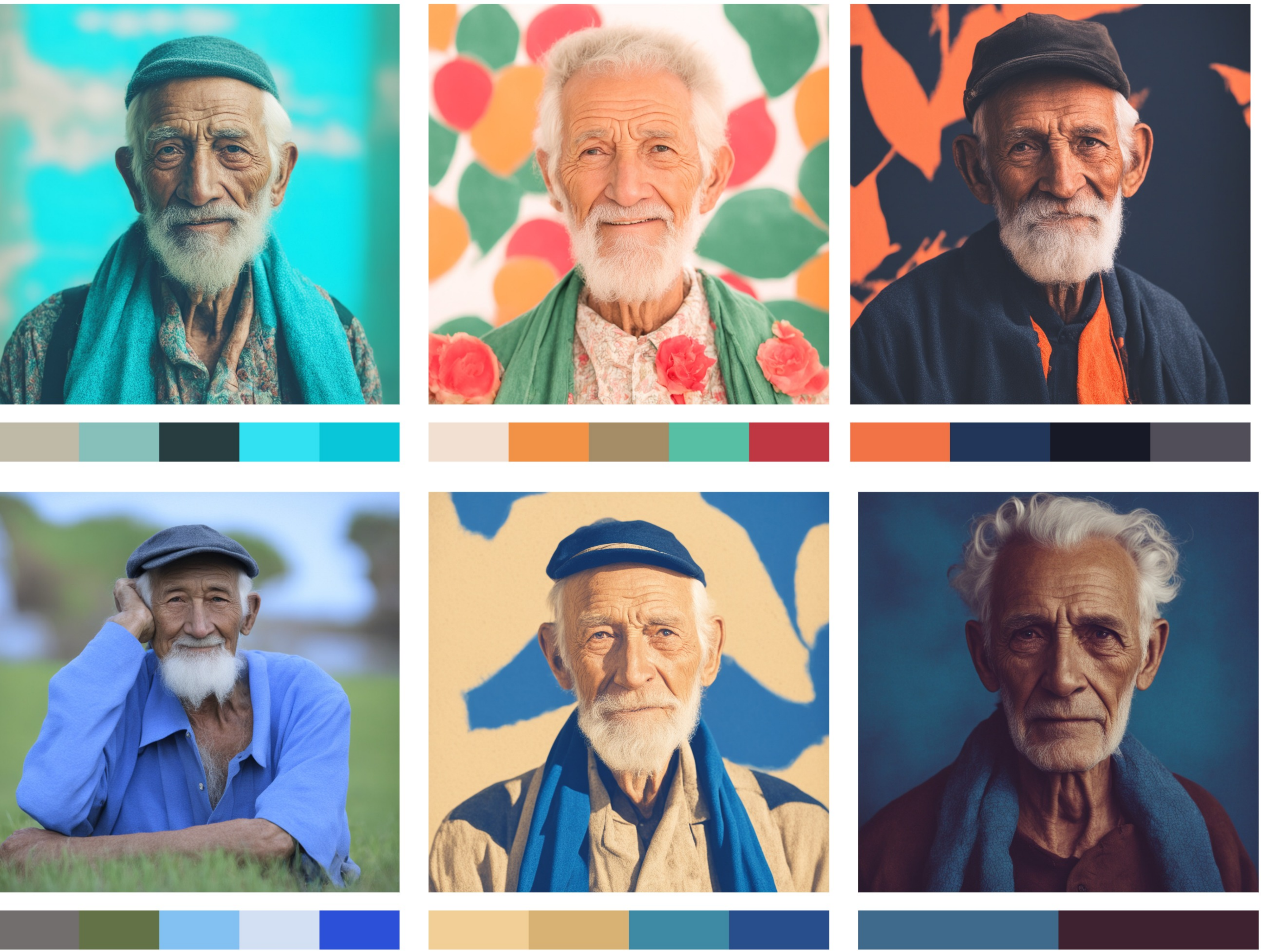 }
    \caption{\textbf{Portraits}: Portraits generated using our methods with the same seed and prompt: ``A photo portrait of an old man.''}
    \label{fig:portraits}
    \Description{will be added}
\end{figure}

\begin{figure}
    \centering
    \includegraphics[width=1\linewidth]{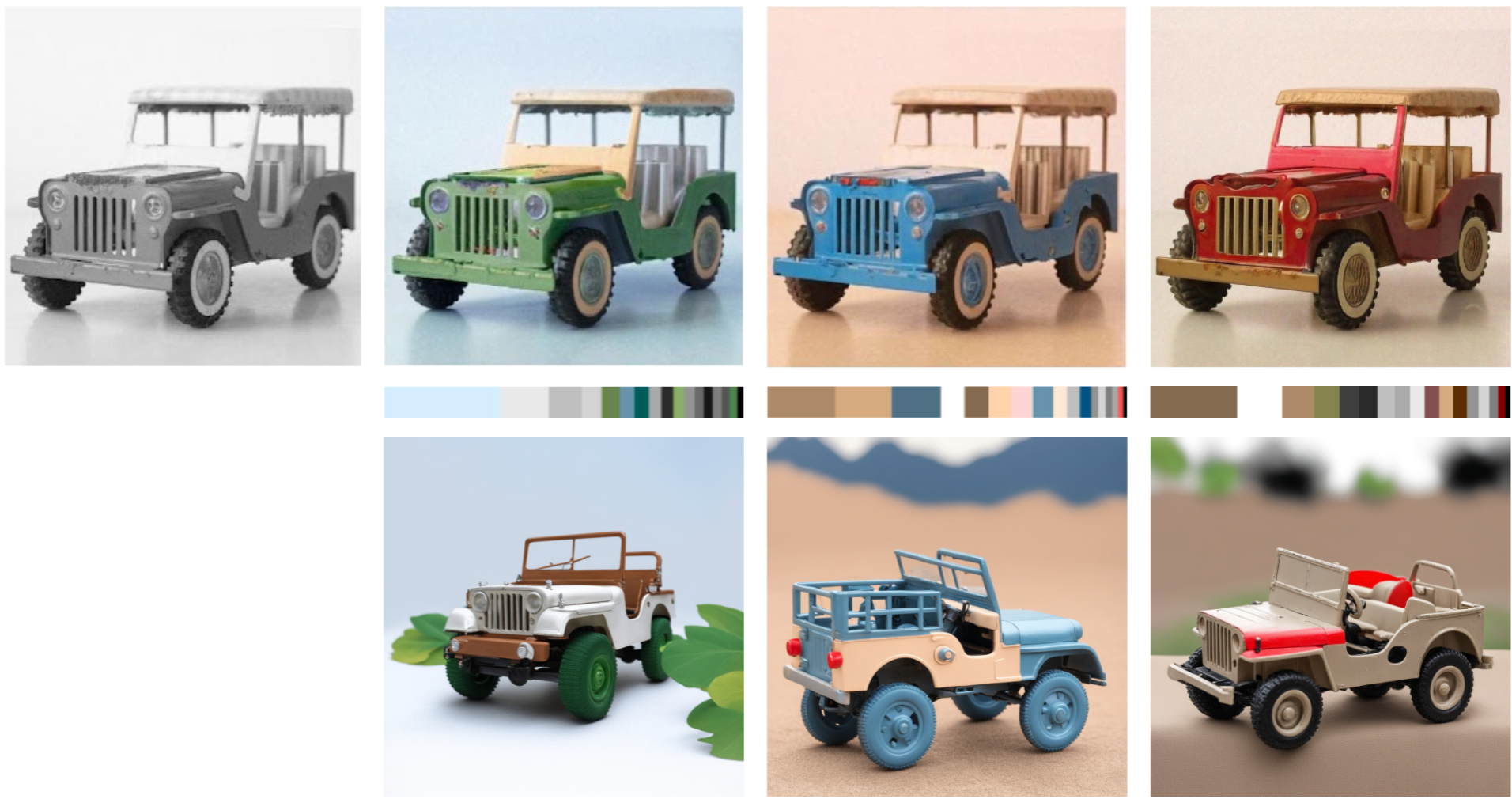}
    \caption{\textbf{Post-hoc recoloring of a grayscale image using Composer vs. histogram guided generation}: we compare Composer's palette-based grayscale recoloring (top row) and our model's palette-aligned generation (bottom row) where the input is provided as a histogram as our model supports both forms. Our model yields superior color alignment. Images generated using the prompt    
    prompt: ``A vintage toy jeep with a soft top roof and open sides, placed on a neutral surface.''}
    \label{fig:recoloring_composer}
    \Description{will be added}
\end{figure}

\begin{figure}
    \centering
    \includegraphics[width=1\linewidth]{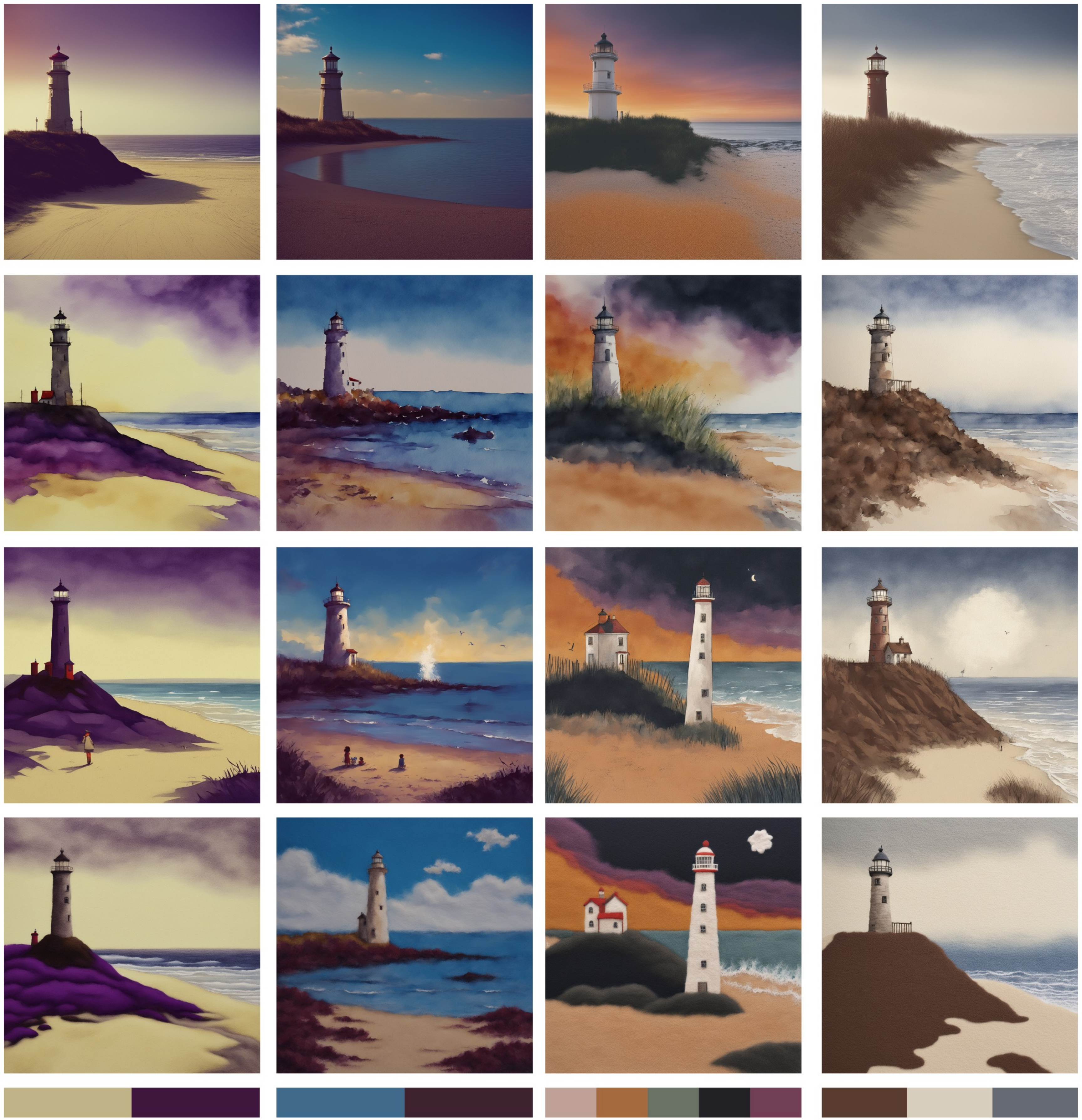 }
    \caption{\textbf{Palette alignment with different styles}: Generated using our method for different styles with the same seed (and similar prompts excluding the style) for photo, watercolor, children book's illustration, and woolfelt.}
    \label{fig:palette_and_style_v2}
    \Description{will be added}
\end{figure}

\begin{figure}
    \centering
    \includegraphics[width=1\linewidth]{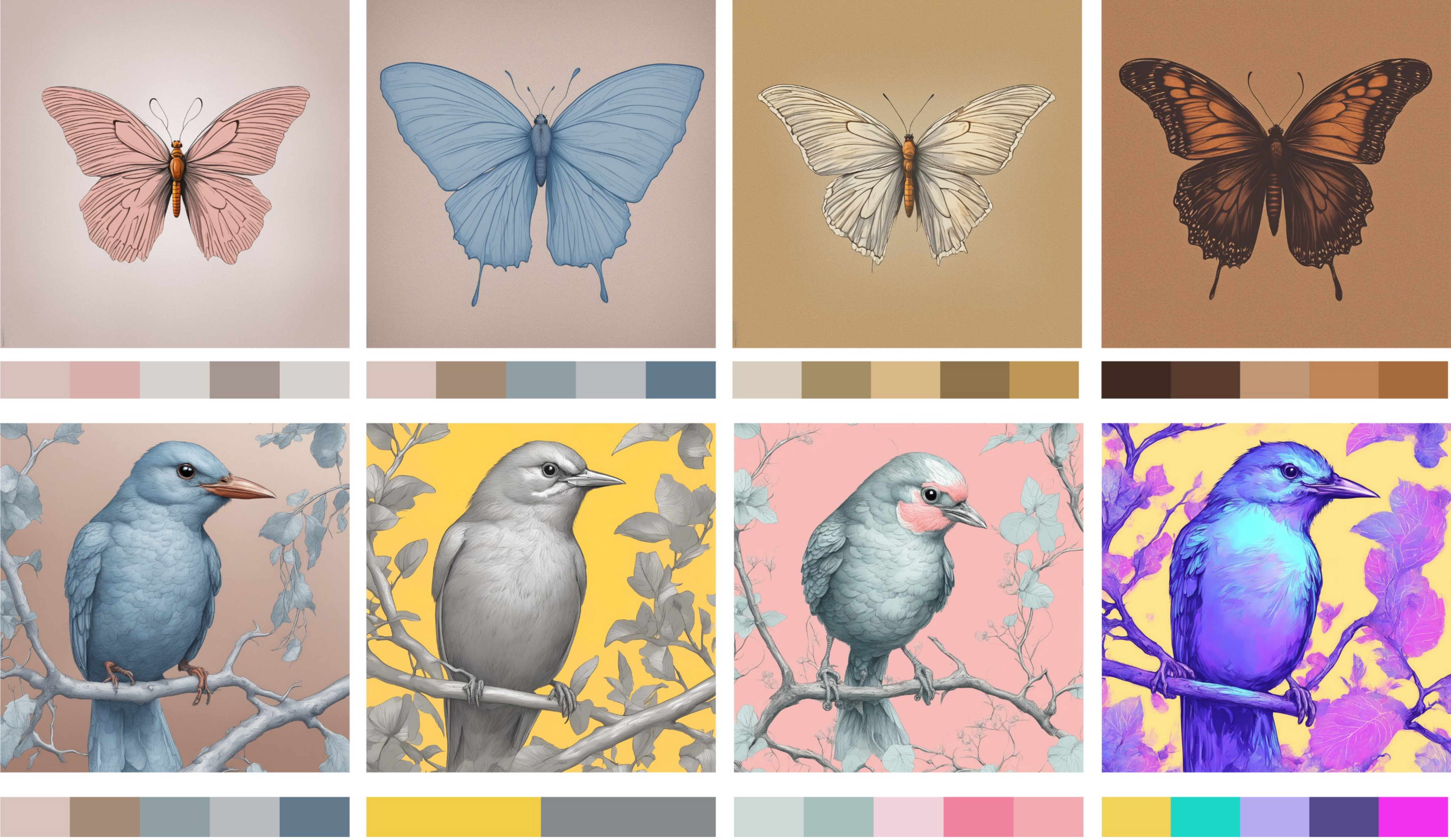 }
    \caption{\textbf{Technical book illustrations generated with our model}: Using the same seed and prompt per row with different palettes. Demonstrating robustness to saturated and mild palettes.}
    \label{fig:recoloring_composer}
    \Description{will be added}
\end{figure}

\clearpage

\makeatletter % Allows access to internal LaTeX macros like \@title
\setcounter{page}{1}
\setcounter{section}{0}
\title{Palette Aligned Image Diffusion Supplementary Materials}

\vspace{1em}
{\centering\bfseries\LARGE Palette Aligned Image Diffusion Supplementary Materials\par}
\vspace{1em}
%{\centering\large SUBMISSION ID: PAPERS\_2010\par}

\section{Training and Guidance}

\subsection{Dataset Curation}
\label{sec:dataset_supp}

Diffusion models are inherently data-driven, as their inference relies on sampling from a learned distribution. Prior work, \eg,~\cite{perera2023analyzingbiasdiffusionbasedface}, has demonstrated that dataset biases concerning gender, race, and age are reflected in the generated imagery. For colors, most datasets are skewed towards natural images, leading to similar color biases in models trained on them. 

For instance, inspecting 3D color histograms ($8^3$ bins) of LAION-Art~\cite{laion_art} images reveals that the top 100 bins account for 87\% of histogram values, while the 100 rarest bins represent a mere 0.0756\%. This highlights a wide variety of significantly under-represented, or ``rare'', colors (see \Cref{fig:lainon_sdxl_rolor_bias_all_colors,fig:lainon_sdxl_rolor_bias_rare_colors}). \Cref{fig:sdxl_color_bias_dress,fig:sdxl_color_bias_sofa} demonstrate correlation between the color distribution of images generated with minimal and color-neutral prompts (\eg, ``A dress''), using SDXL-1.0 \cite{podell2023sdxlimprovinglatentdiffusion}, compared with the color distribution of the LAION-Art dataset at Figure \ref{fig:lainon_sdxl_rolor_bias_all_colors}.

To ensure adequate representation of rare colors, we curate a dataset that contains a more balanced representation of different colors. Specifically, we use a 2M images-subset of the LAION-Art dataset, and augment this set with 400K additional images from the LAION-2B-en dataset~\cite{laion2b_en}. These additional image are sampled among those that contain colors from the low valued histogram bins of the LAION-Art dataset. This results in a total of 2.4M images.

\subsection{Identifying Color Control Blocks}
\label{sec:colorBlocks}

Our training process is multi-staged: initially, all layers are trained for 1 million steps with a batch size of 16. Subsequently, we identify the critical block for color control, as described below, and then fine-tune only this specific block for an additional 1 million steps, employing a batch size of 4.

To identify the layers most critical for color control, we adopt a method inspired by \citep{frenkel2024implicitstylecontentseparationusing, cohen2024conditionalbalanceimprovingmulticonditioning}. First, we train the full adapter, which consists of all 11 attention blocks (totaling 70 layers). Subsequently, each attention block is evaluated in isolation: with only one block active and all others off, we perform inference using the same 20 prompts, each paired with one of 20 color palettes randomly sampled from the COCO dataset. We then compute the average Earth Mover's Distance (EMD) between the generated image's color distribution and the target palette.

The results of this analysis are summarized in the supplementary material Table~\ref{tab:emd_layer_results}. This analysis reveals that the up\_blocks.0.attentions.1 block, corresponding to the second attention module in the first U-Net upsampling block, exhibits the most significant influence on color fidelity, as indicated by achieving the lowest average EMD. 

\begin{table}[htbp]
\centering
\footnotesize
\caption{Combined EMD and Layer Count Results per Attention Block}
\label{tab:emd_layer_results}
\begin{tabular}{lllc}%{l S[table-format=1.4] S[table-format=1.4] S[table-format=2.0]}
\toprule
\textbf{ } & {\textbf{Mean}} & {\textbf{Std}} & {\textbf{Num}} \\
\textbf{Block Name} & {\textbf{EMD}} & {\textbf{EMD}} & {\textbf{Layers}} \\
\midrule
down\_blocks.1.attentions.0              & 0.2218   & 0.0590  & 2  \\
down\_blocks.1.attentions.1              & 0.2290   & 0.0549  & 2  \\
down\_blocks.2.attentions.0              & 0.2033   & 0.0471  & 10 \\
down\_blocks.2.attentions.1              & 0.2503   & 0.0946  & 10 \\
mid\_block.attentions.0.transformer\_blocks & 0.2349   & 0.0624  & 10 \\
up\_blocks.0.attentions.0                & 0.2274   & 0.0598  & 10 \\
up\_blocks.0.attentions.1                & 0.1298   & 0.0234  & 10 \\
up\_blocks.0.attentions.2                & 0.2222   & 0.0574  & 10 \\
up\_blocks.1.attentions.0                & 0.2213   & 0.0569  & 2  \\
up\_blocks.1.attentions.1                & 0.2205   & 0.0554  & 2  \\
up\_blocks.1.attentions.2                & 0.1996   & 0.0516  & 2  \\
\bottomrule
\end{tabular}
\end{table}

\subsection{Conditioning Dropout}
During training we mix between full histograms and palettes and described in the paper. Dropout during training serves multiples purposes including the mixture of histograms and palettes, and to enable Classifier-Free-Guidance (CFG) including inference with a null condition. In ~\Cref{table:cfg_dropout} we show dropout probabilities for text and color conditions. Additionally the distance condition is provided as non zero only when using a palette condition, and entropy condition is dropped with a probability of 10\%.

\begin{table}[h!]
\small
\begin{tabular}{c|c|c|c}
 Embedding Type & Histogram & Palette & None \\ 
 \hline
 Color Embeddings Prob. & 45\% & 45\% & 10\% \\  
 Text Embeddings Prob. & 80\% & 80\% & 5\% \\
\end{tabular}
\caption{Training conditioning discard ratios to enable Classifier Free Guidance. Embedding Type "None" indicates all zeros.}
\label{table:cfg_dropout}
\end{table}

\subsection{Relative Entropy Guidance}
Our image generation process incorporates two key conditioning scalars: \emph{palette to histogram distance} and \emph{entropy}. The entropy measure quantifies global color distribution across histogram bins, independent of the palette. In contrast, the distance measure aims to quantify the deviation from the palette using a predefined distance metric, and should be trained to align with user's expectation. In the paper we use the terminology relative entropy, and relative distance to indicate the CFG direction between the "negative" and "positive" prompt.

For ablation study of entropy guidance, \emph{with no color conditioning}, and specifically the relative entropy, we have trained an IP-Adapter which is only conditioned on entropy (\ie, this IP-Adapter trains on a similar architecture as our full adapter where all the non-entropy inputs, including the color condition, are zero). We train this adapter for 100K steps. In generation we use two entropy values for CFG, marked $E^+$ and $E^-$, where the $E^+$ value is used with the positive condition and $E^-$ is used for the null condition, resulting in a ``relative entropy'' (RE) condition. \Cref{tab:relative_etnropy} shows images generated using different values of $E^+$ and $E^-$, resulting with different relative entropy values. We observe that (1) while entropy is a positive number, it is possible to use a ``negative'' relative guidance to lower the entropy from the unconditioned reference in CFG by providing a high number for the null condition, and a low number for the positive condition; and (2) The visual similarity along the diagonal shows that the relative measure is sufficiently consistent for our purposes. 

\section{Comparison with other methods}
\label{sec:appendix_comparison}

\subsection{User Study}
\label{sec:UserStudy}

\begin{figure}
    \centering
    \includegraphics[width=1\linewidth]{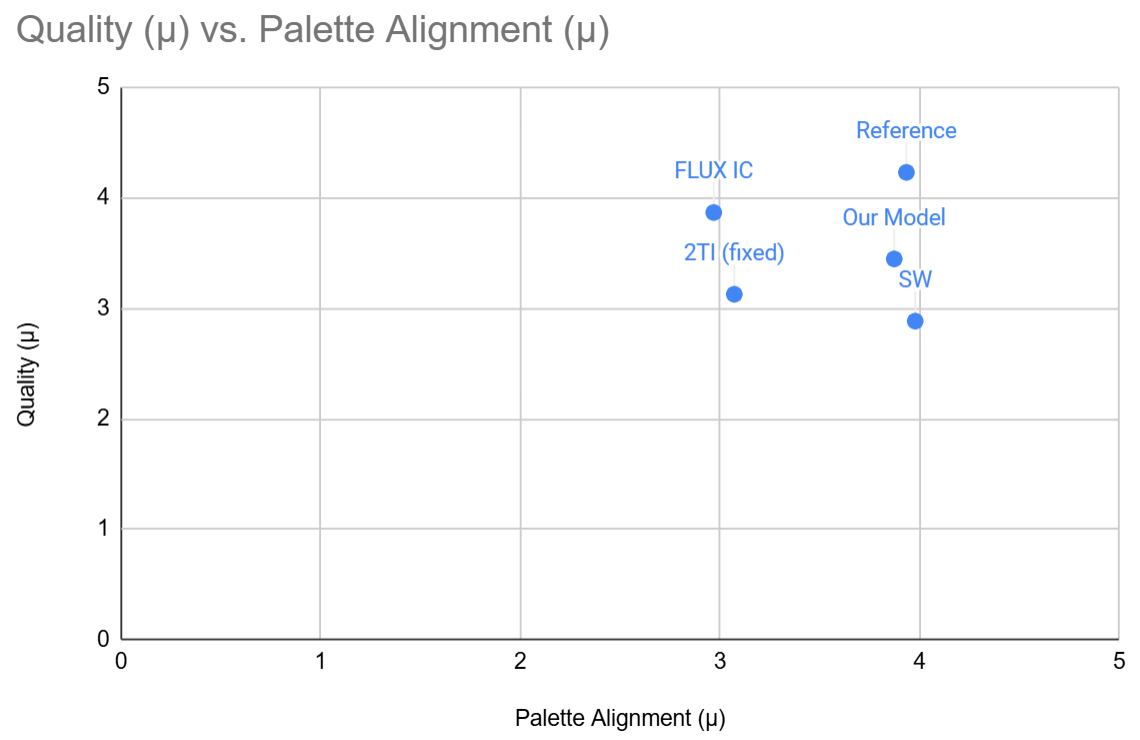}
    \caption{\textbf{User study results:} Our method strikes an optimal balance between palette adherence and image quality, as indicated by its position on the Pareto front.}
    \label{fig:user_study_graph}
    \Description{will be added}
\end{figure}

\begin{figure}
    \centering
    \includegraphics[width=1\linewidth]{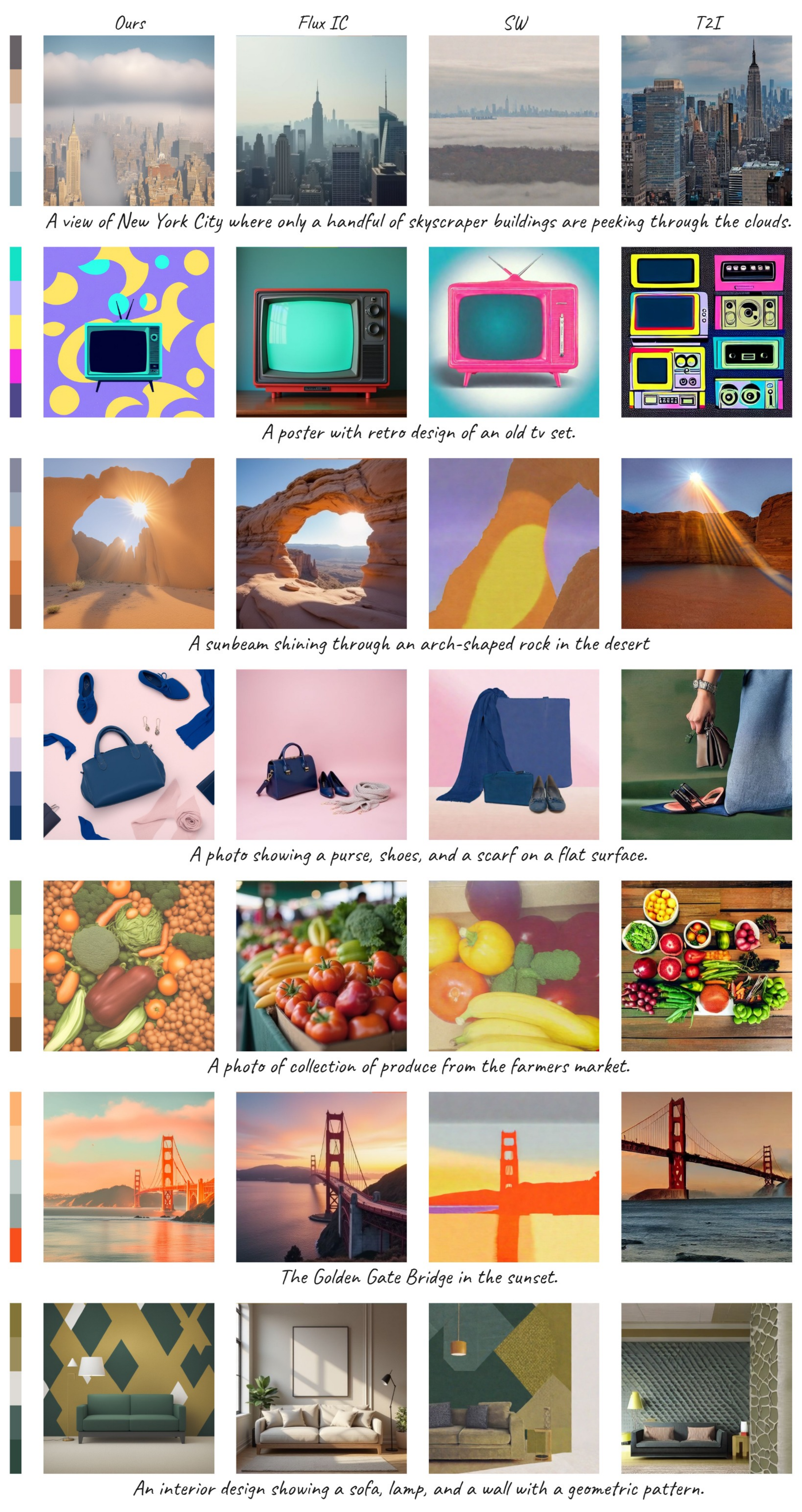}
    \caption{\textbf{User study:} A collection of images taken from the user study.}
    \label{fig:user_study_images}
    \Description{will be added}
\end{figure}

The images generated for the user study, along with their prompts, appear in \Cref{fig:user_study_images}. As previously stated, we compare our method against several other methods (SW, Flux IC, T2I, and the set of reference images). The reference images that we compared against (from which the palette was extracted by designers) can be found at https://www.picmonkey.com/blog/color-combinations-graphic-design. The results are described in the paper. Our method strikes an optimal balance between palette adherence and image quality, as indicated by its position on the Pareto front (See \Cref{fig:user_study_graph}).

\section{Qualitative Results}
\label{sec:QualitativeResults}

\begin{figure*}
    \centering
    \begin{subfigure}[b]{0.47\textwidth}
        \centering
        \includegraphics[width=0.85\textwidth]
        {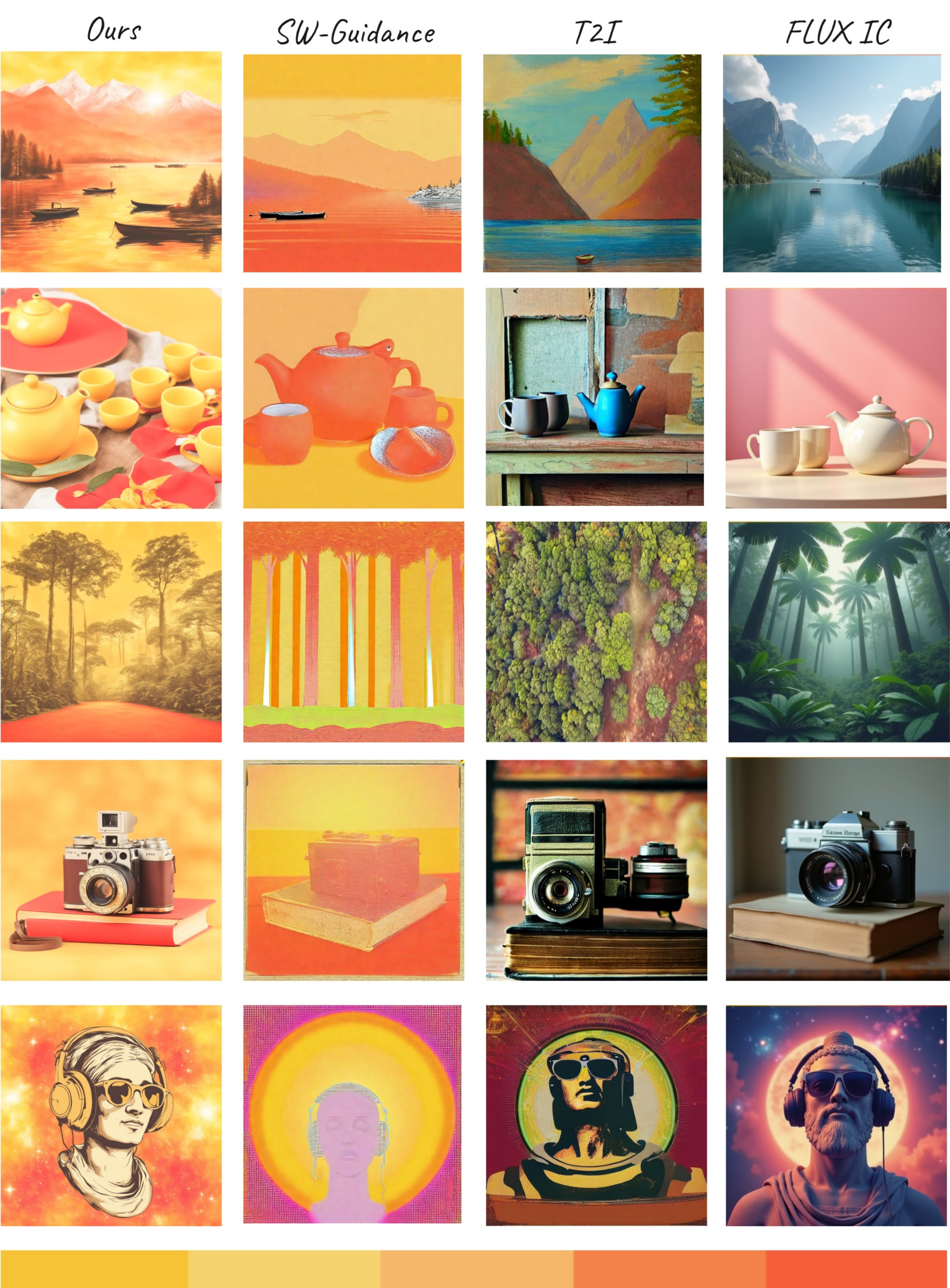}
         %\caption{prompt=``A beach chair under an umbrella on the beach.''.}
    \end{subfigure}
    \begin{subfigure}[b]{0.47\textwidth}
        \centering
        \includegraphics[width=0.85\textwidth]
        {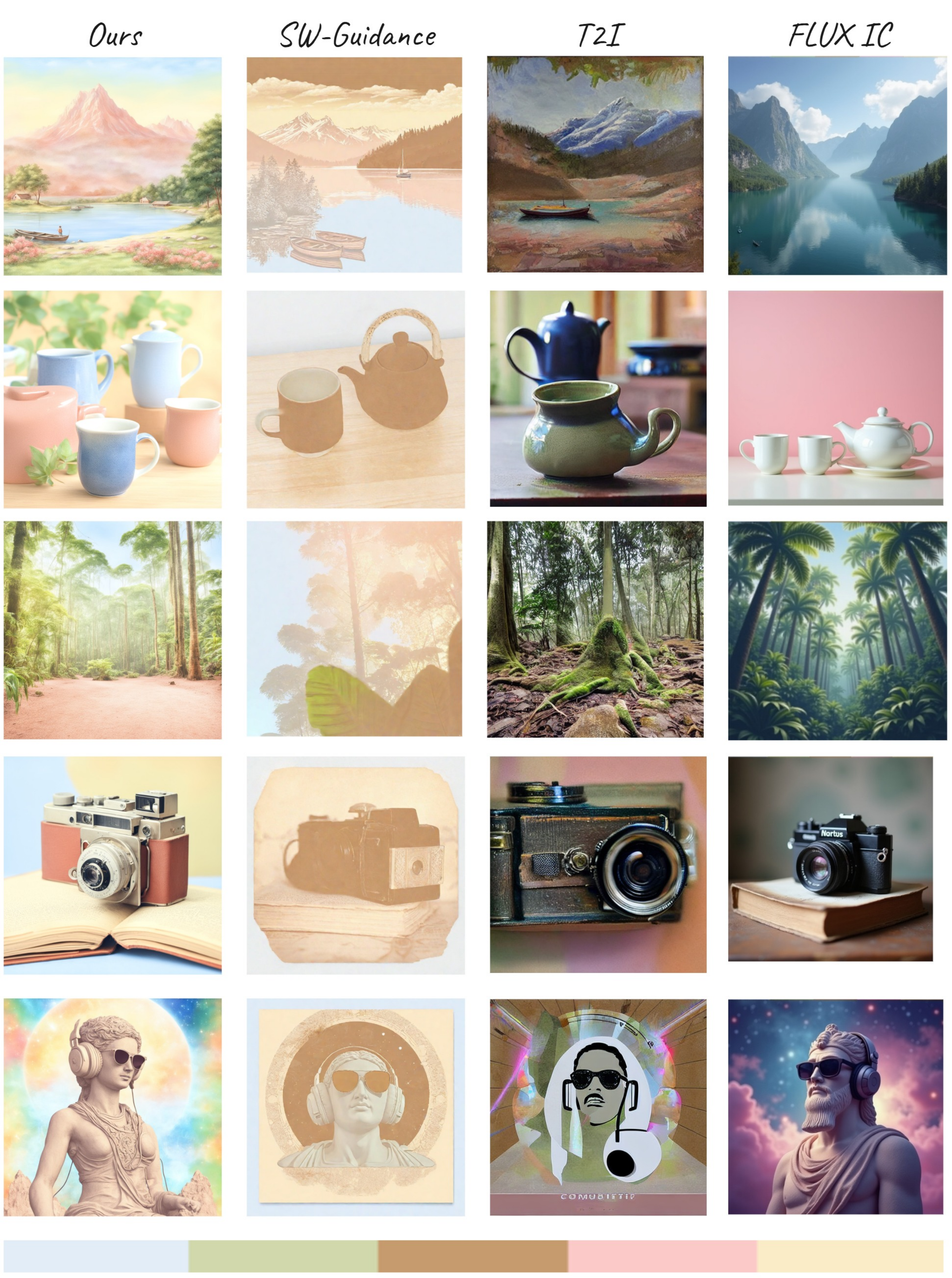}
         %\caption{prompt=``An expensive car driving the highway.''.}
    \end{subfigure}
    \begin{subfigure}[b]{0.47\textwidth}
        \centering
        \includegraphics[width=0.85\textwidth]
        {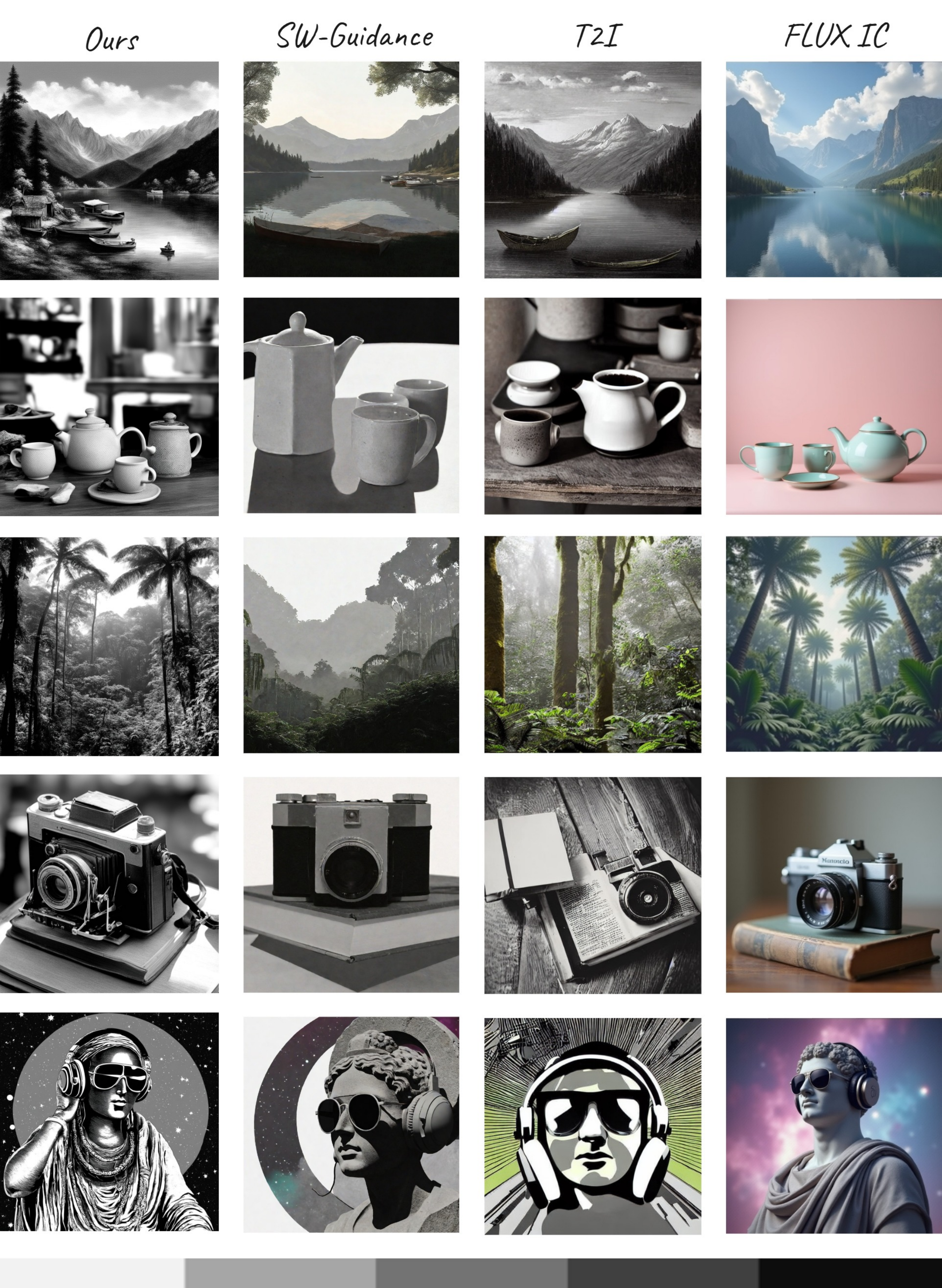}
         %\caption{prompt=``A beach chair under an umbrella on the beach.''.}
    \end{subfigure}
    \begin{subfigure}[b]{0.47\textwidth}
        \centering
        \includegraphics[width=0.85\textwidth]
        {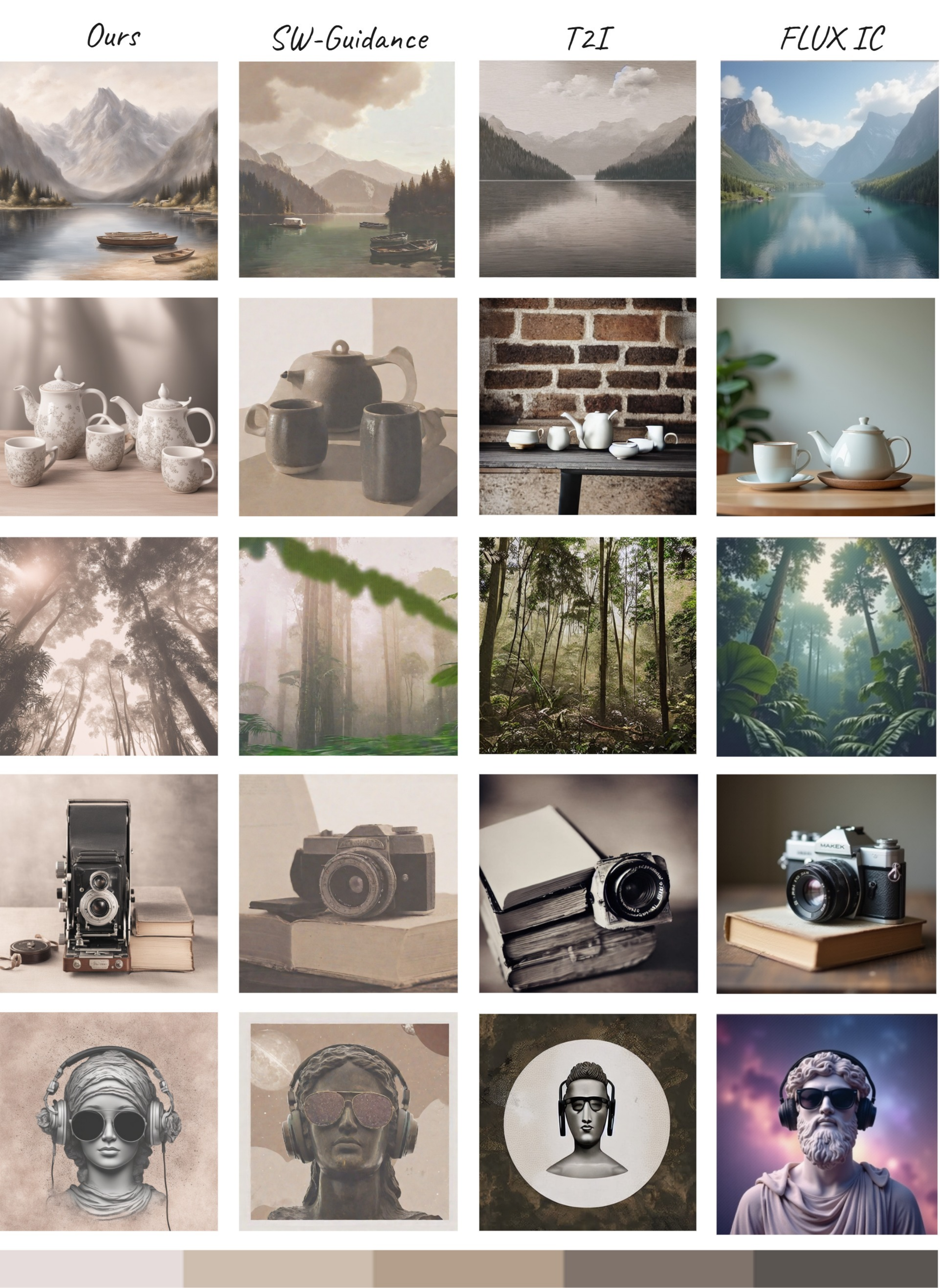}
         %\caption{prompt=``An expensive car driving the highway.''.}
    \end{subfigure}
    \caption{\textbf{Comparison with other methods:} Images generated using different palettes and prompts but same seed. Images are aggregated by palette and prompts by order: "A realistic mountain landscape with a lake and boats", "A ceramic teapot and mugs on a table", "A photo of a rainforest with high trees and big leaves", "A vintage camera resting on an old book.", "A poster design of an ancient statue in headphones and sunglasses on a cosmic background."}
    \label{fig:comparison_cont}
    \Description{will be added}
\end{figure*}

Qualitative results provided in figures in the paper and the appendix. We also include images from the user study as well as comparison with other methods such as in ~\Cref{fig:comparison_cont}

\begin{table*}
\addtolength{\tabcolsep}{-0.4em}
\centering
\begin{tabular}{ccccccc}
& $E^+$ = 0 & $E^+$ = 0.5 & $E^+$ = 1 & $E^+$ = 3 & $E^+$ = 5 & $E^+$ = 8 \\
$E^-$ = 0 &
\includegraphics[width=0.14\textwidth]{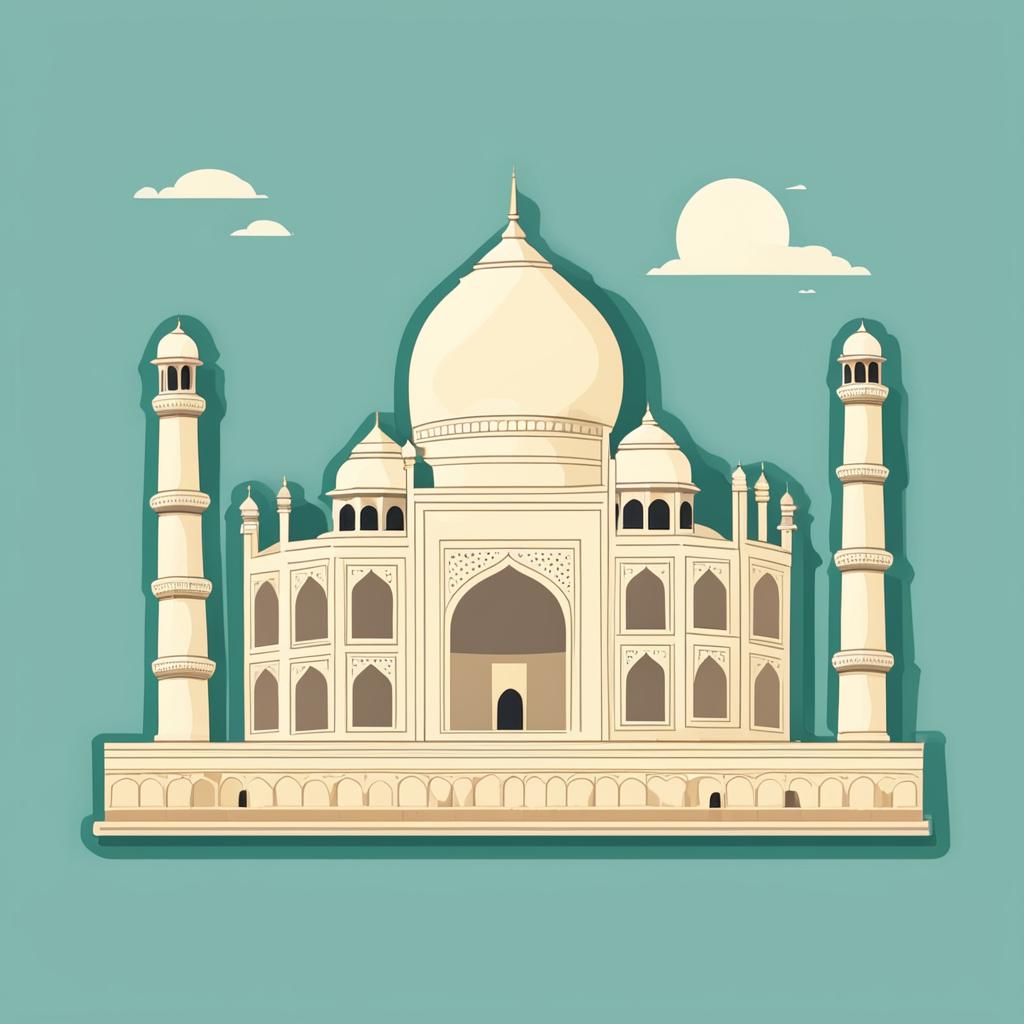}& \includegraphics[width=0.14\textwidth]{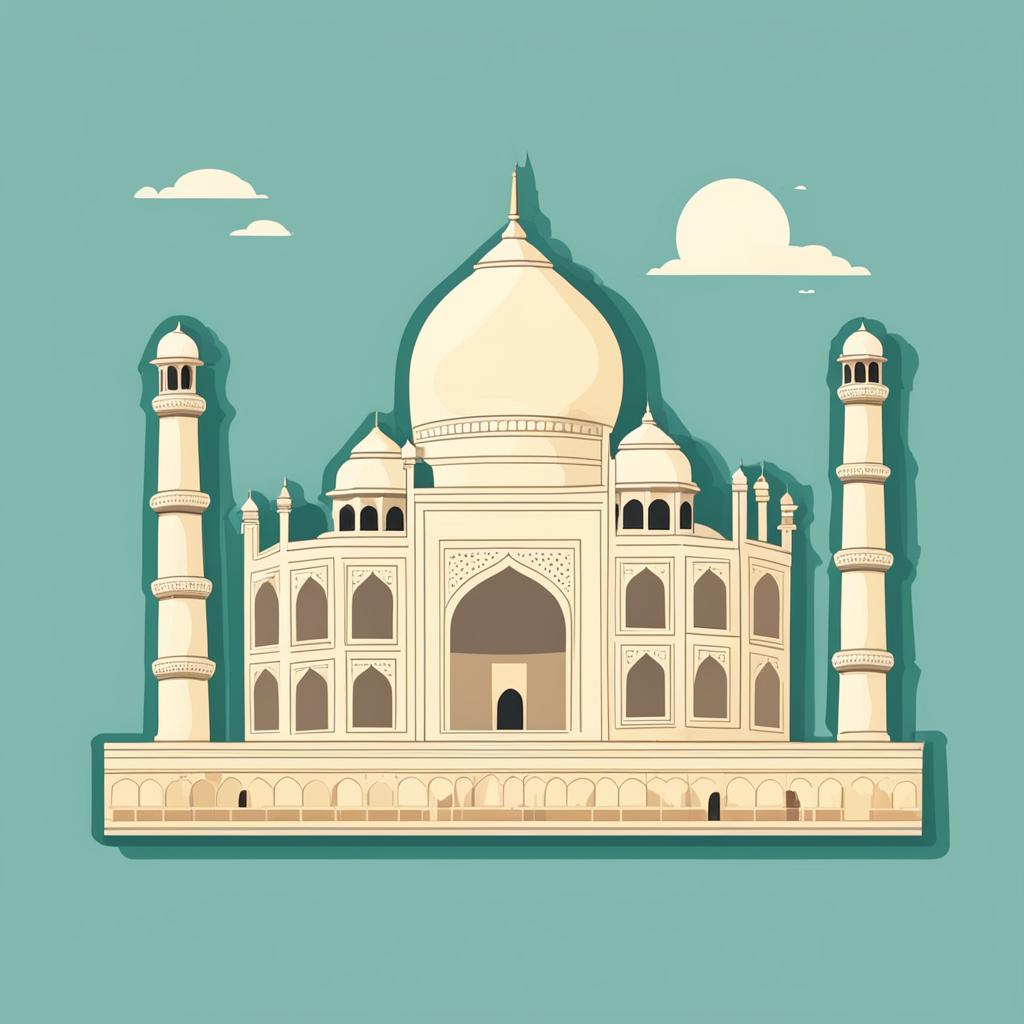}&
\includegraphics[width=0.14\textwidth]{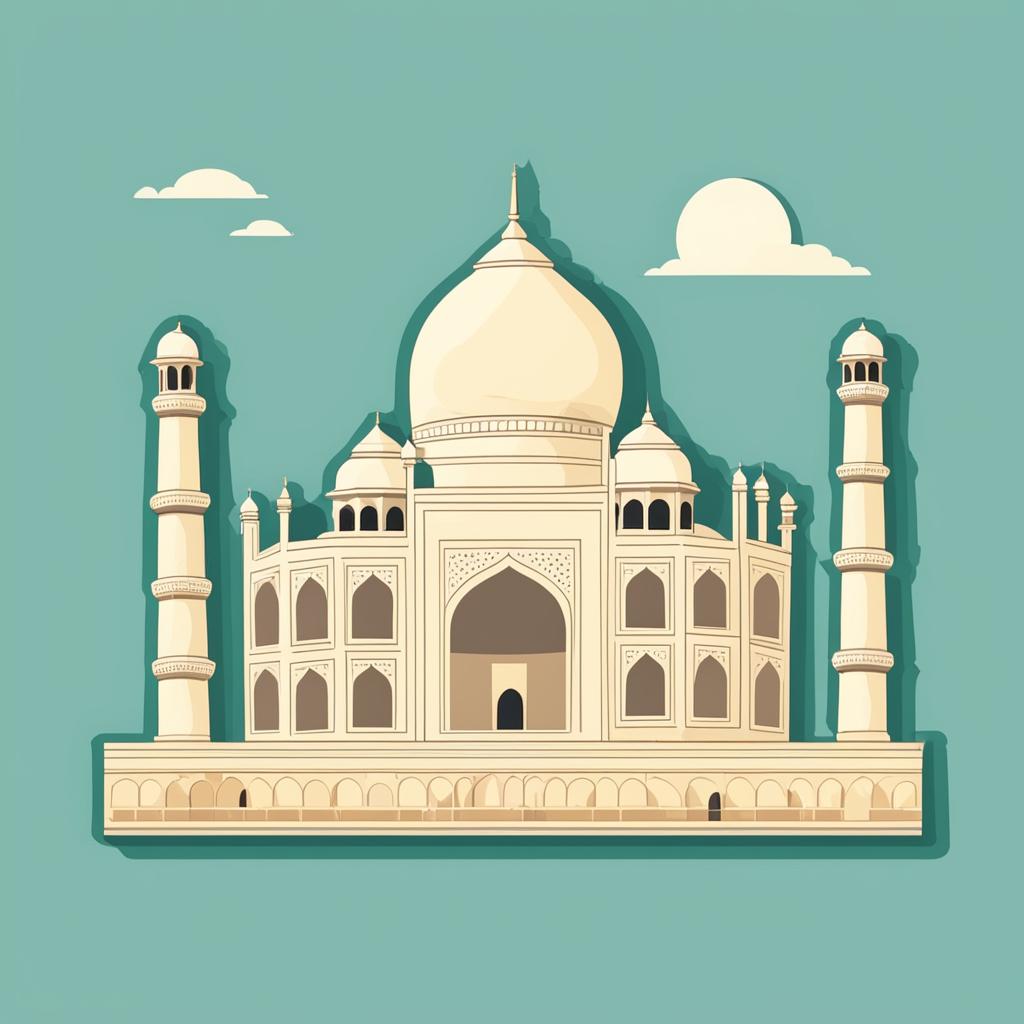}&
\includegraphics[width=0.14\textwidth]
{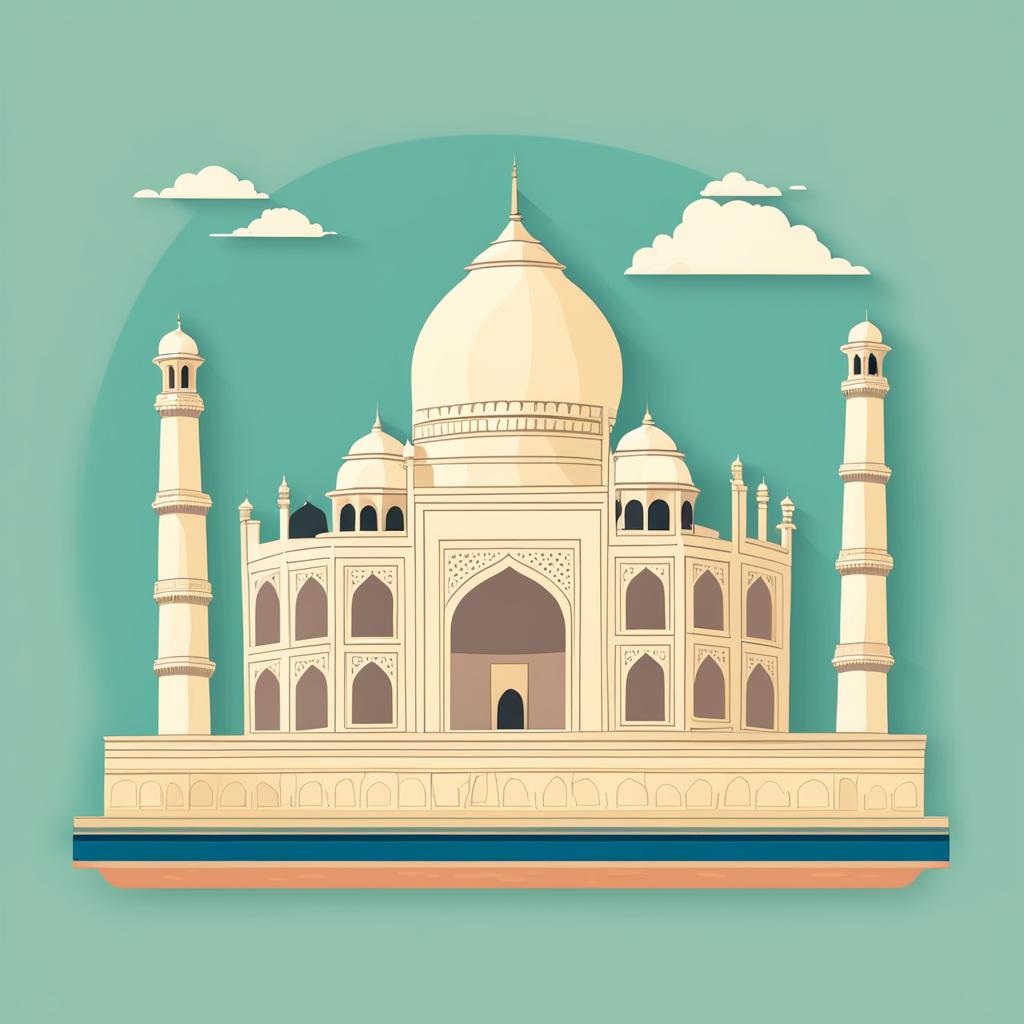}&
\includegraphics[width=0.14\textwidth]{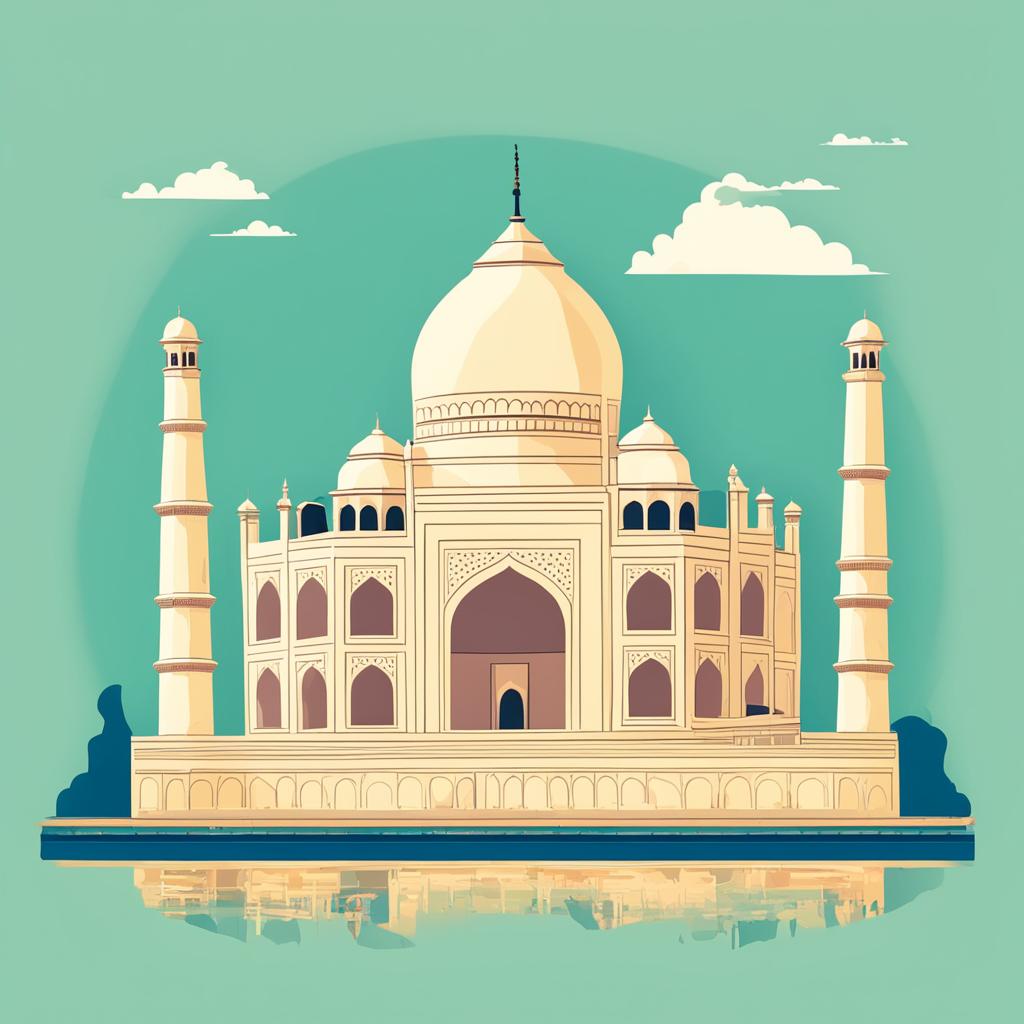}&
\includegraphics[width=0.14\textwidth]{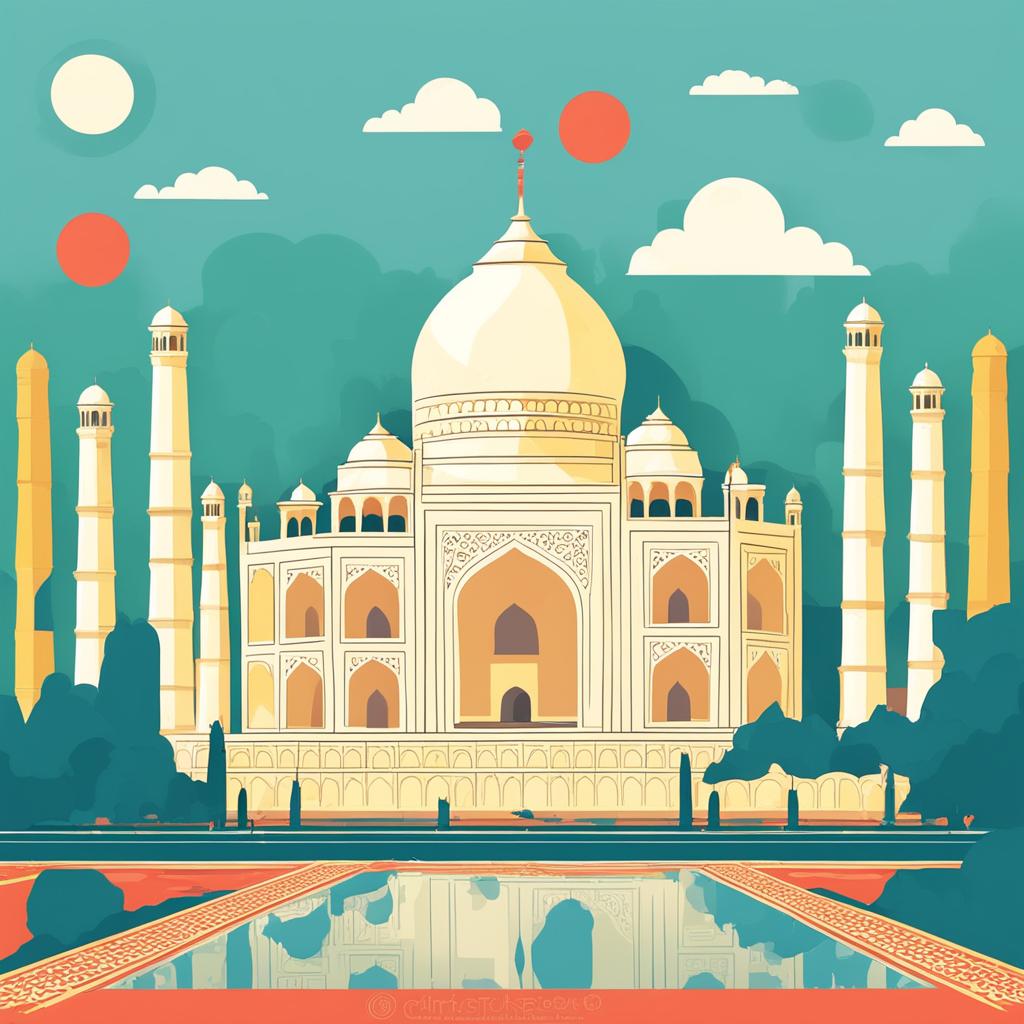}\\
$E^-$ = 0.5 &
\includegraphics[width=0.14\textwidth]{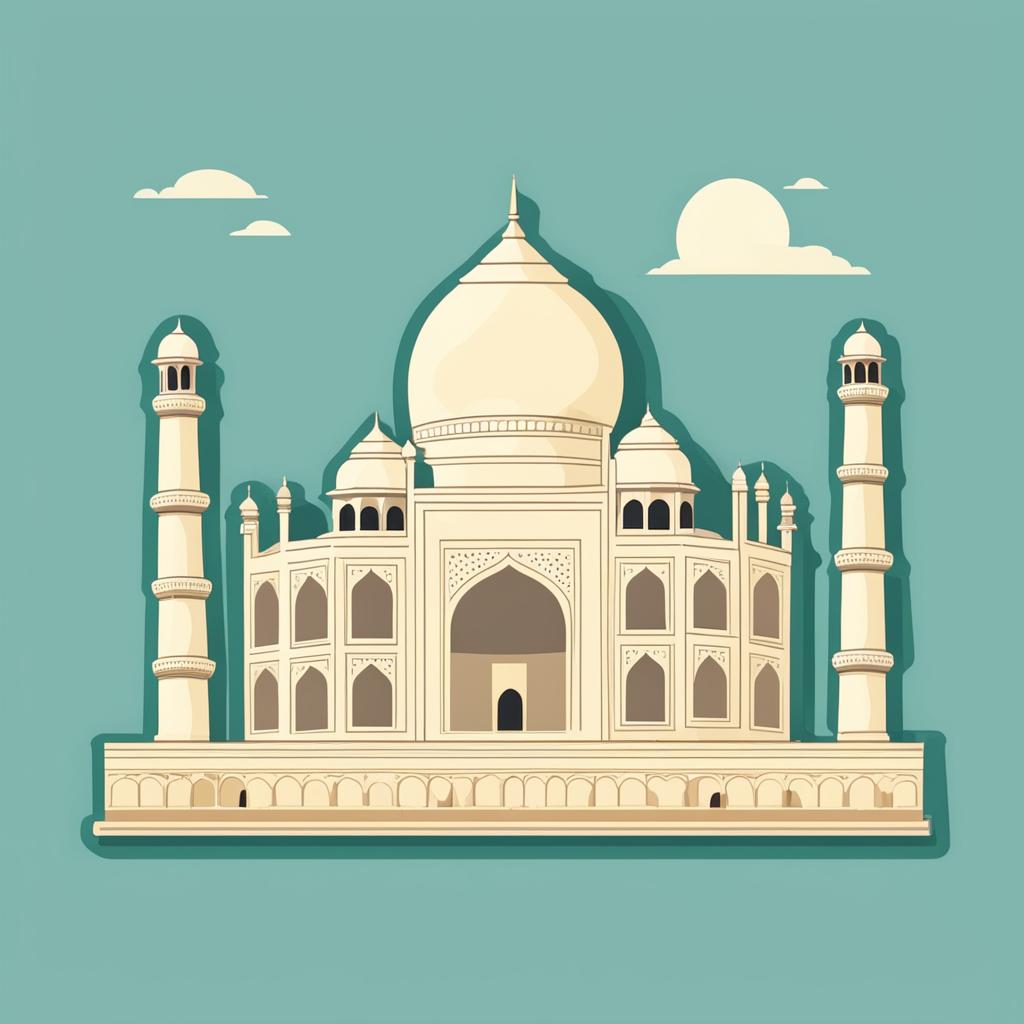}& \includegraphics[width=0.14\textwidth]{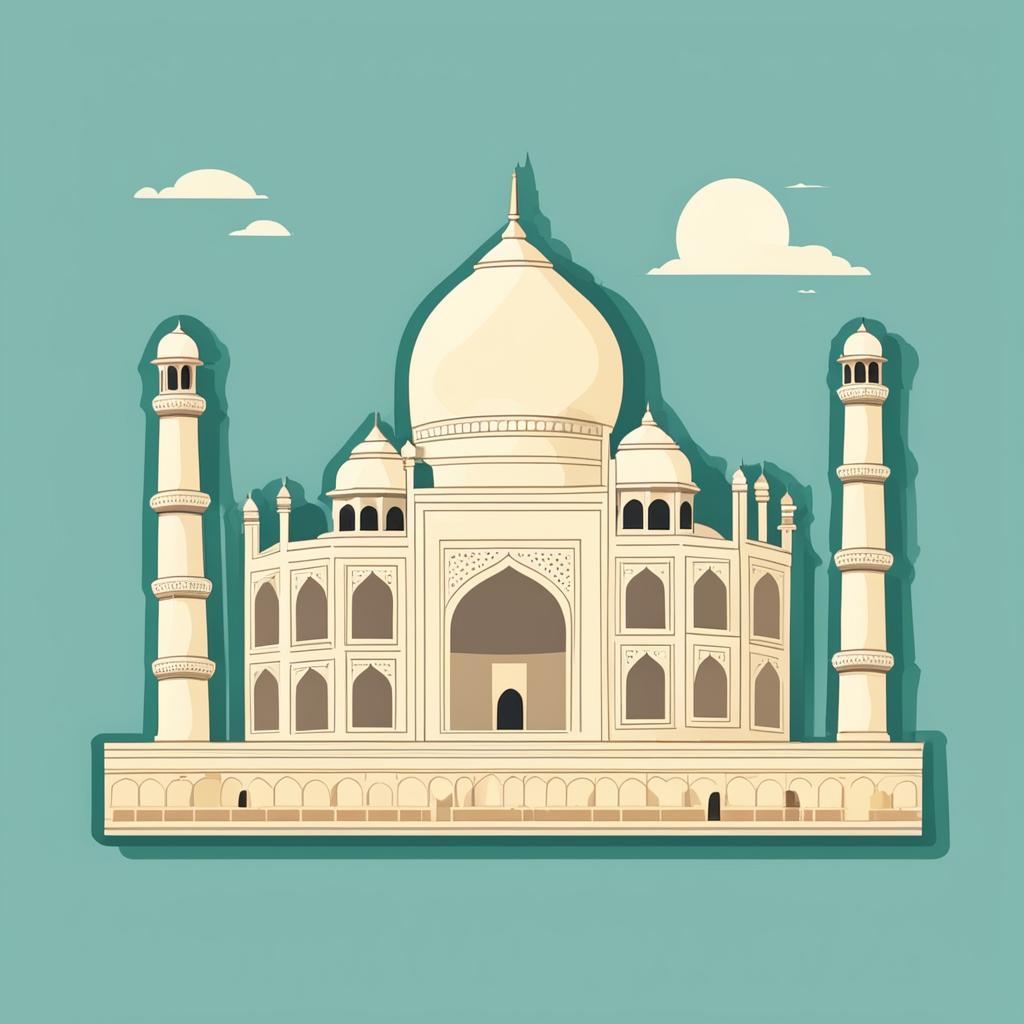}& 
\includegraphics[width=0.14\textwidth]{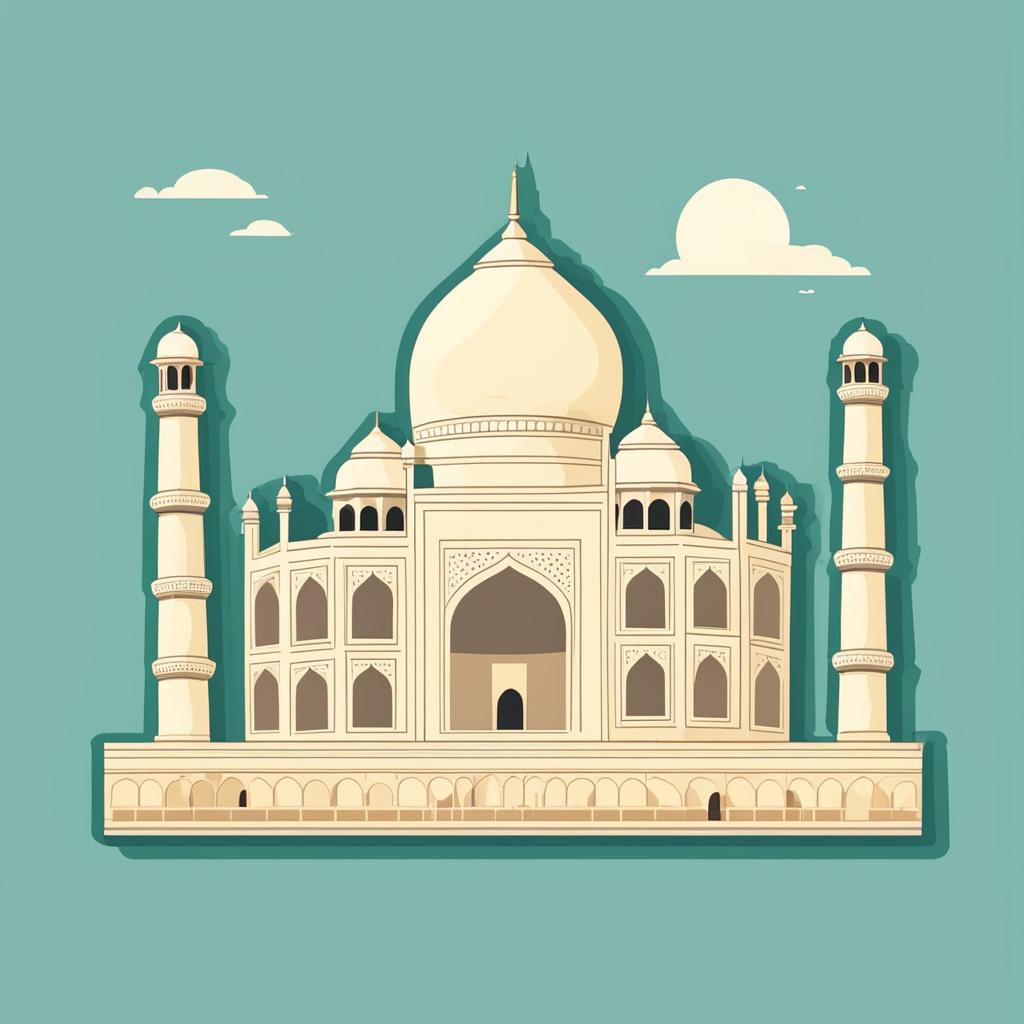}& 
\includegraphics[width=0.14\textwidth]
{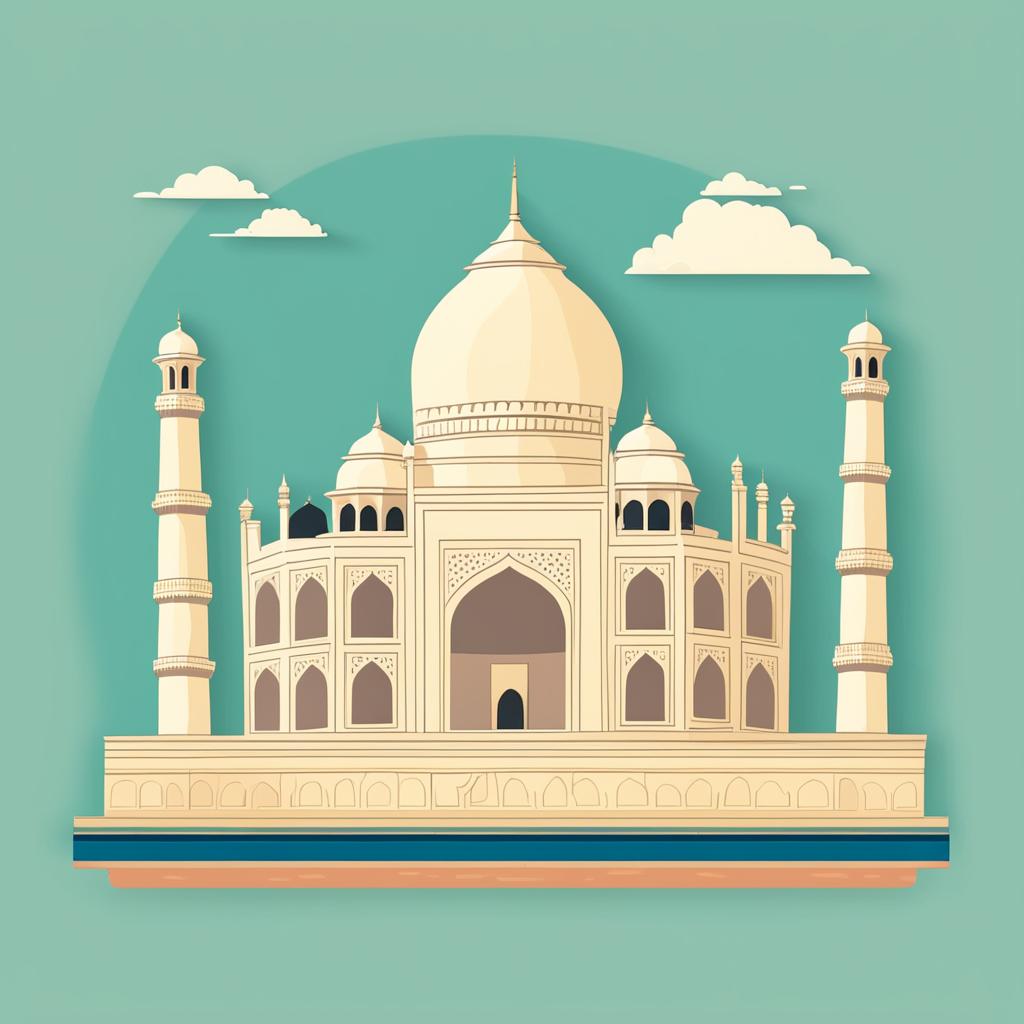}& 
\includegraphics[width=0.14\textwidth]{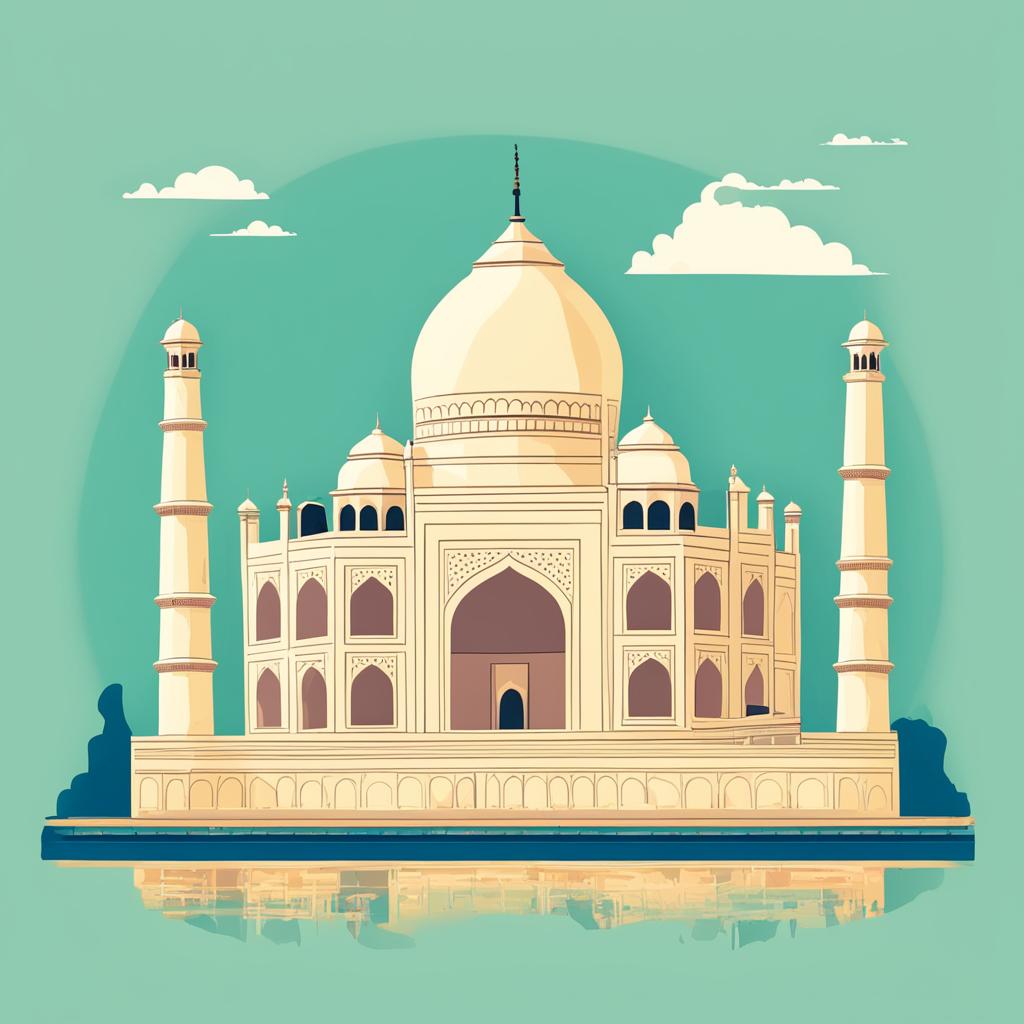}& 
\includegraphics[width=0.14\textwidth]{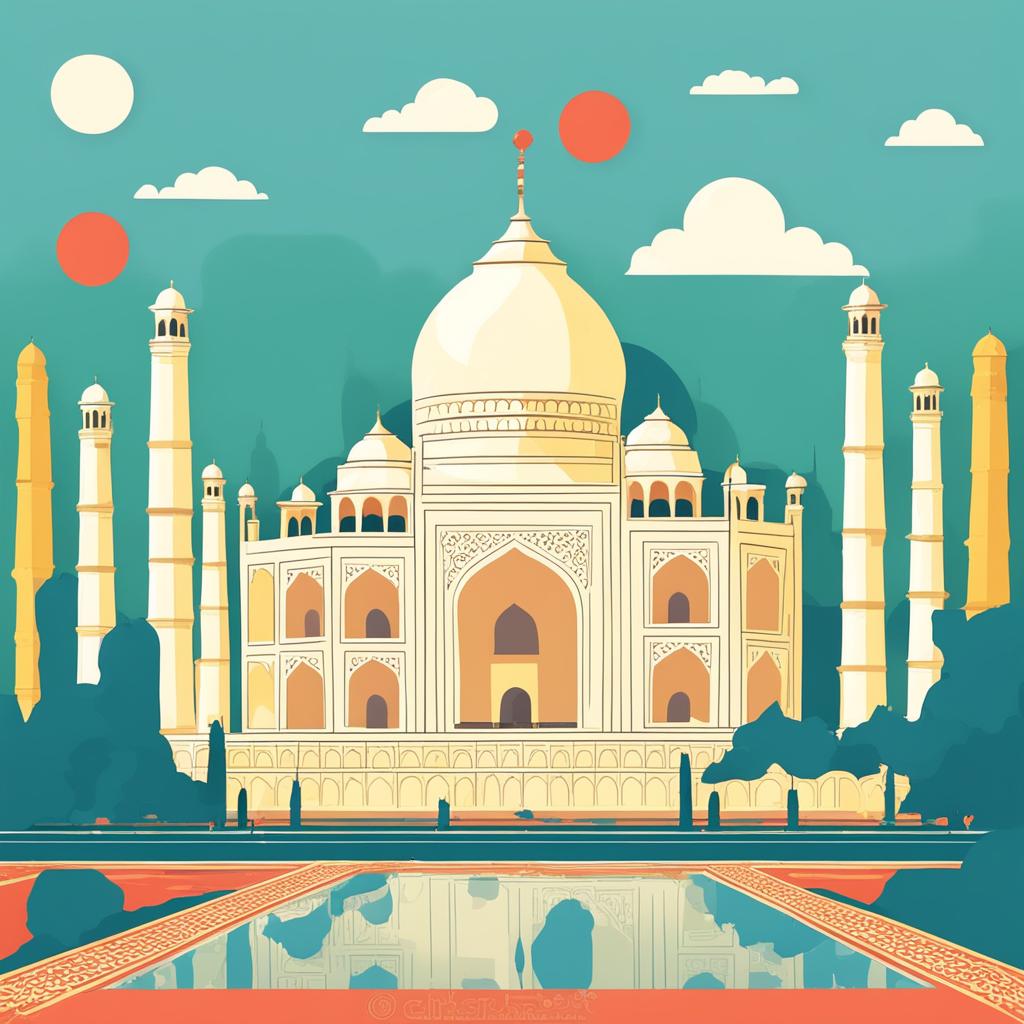}\\
$E^-$ = 1 &
\includegraphics[width=0.14\textwidth]{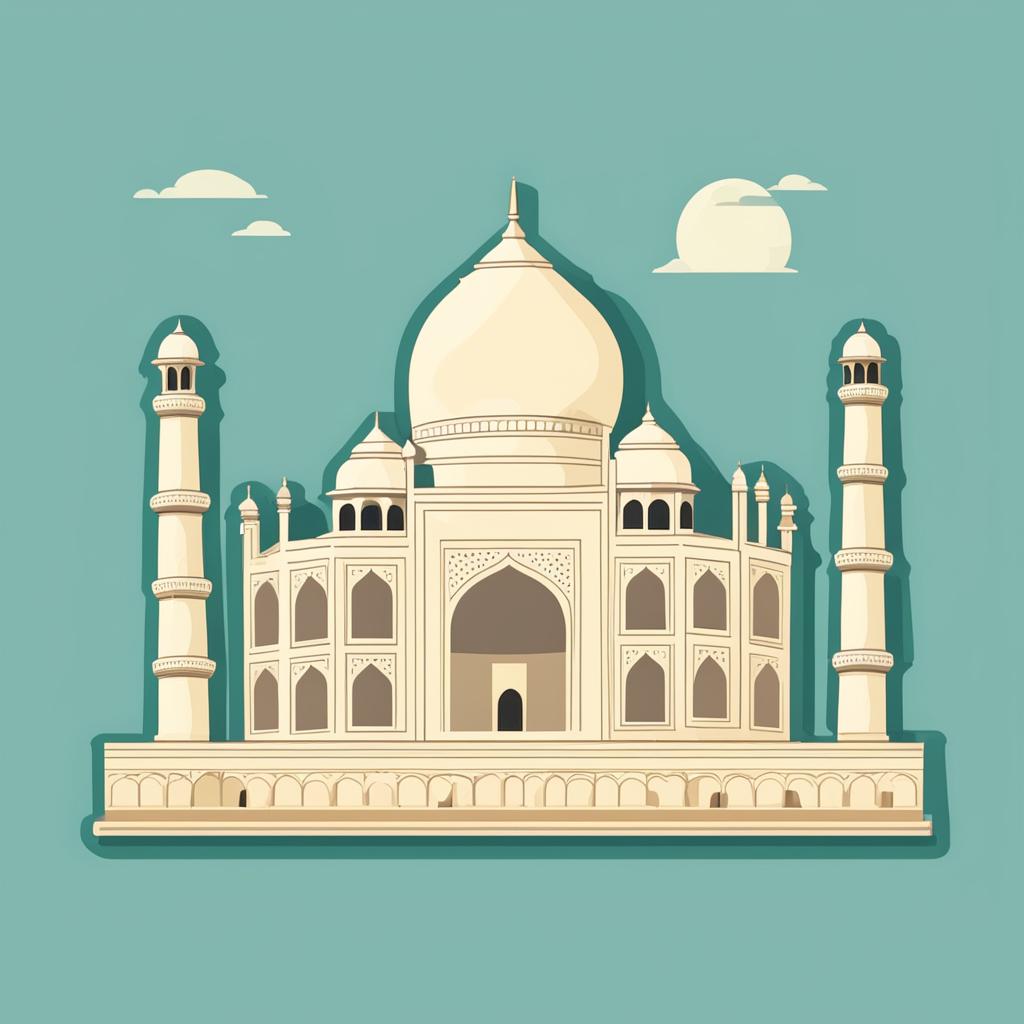}& \includegraphics[width=0.14\textwidth]{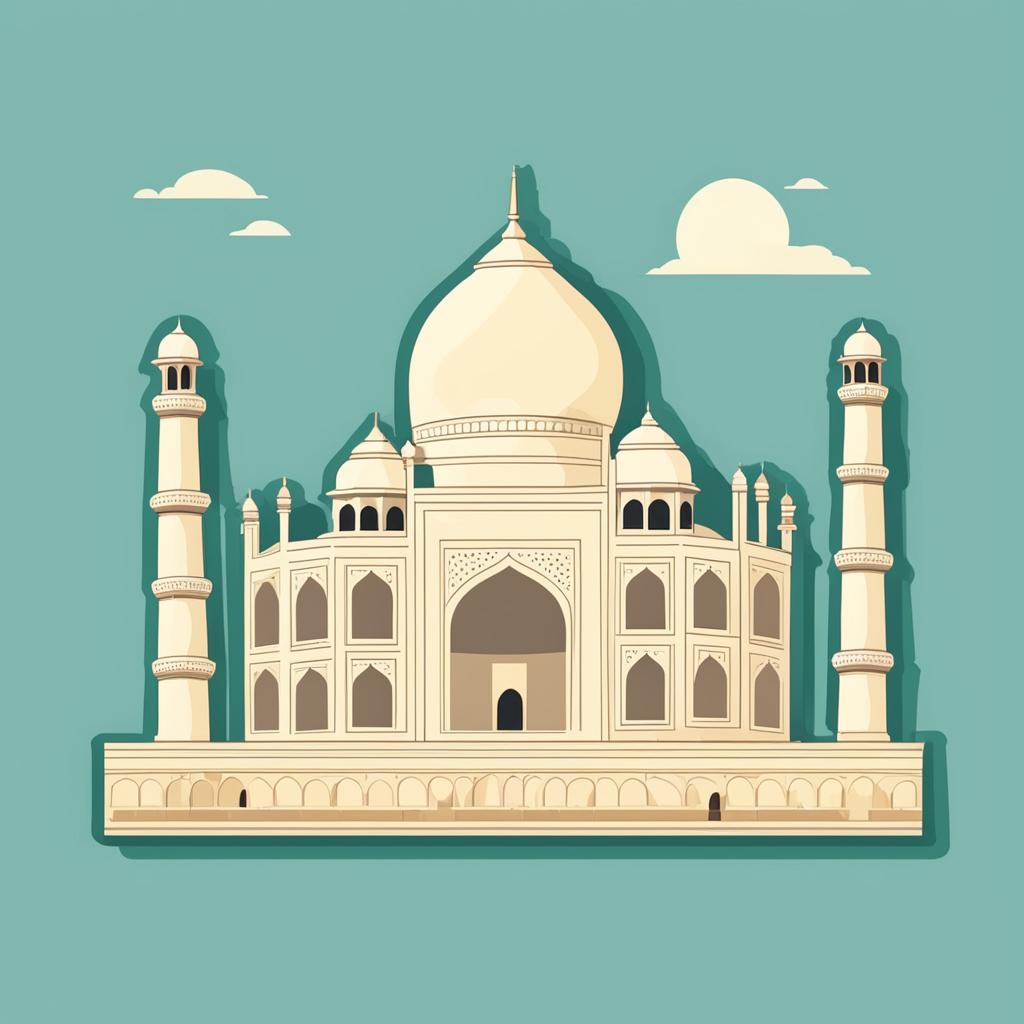}&
\includegraphics[width=0.14\textwidth]{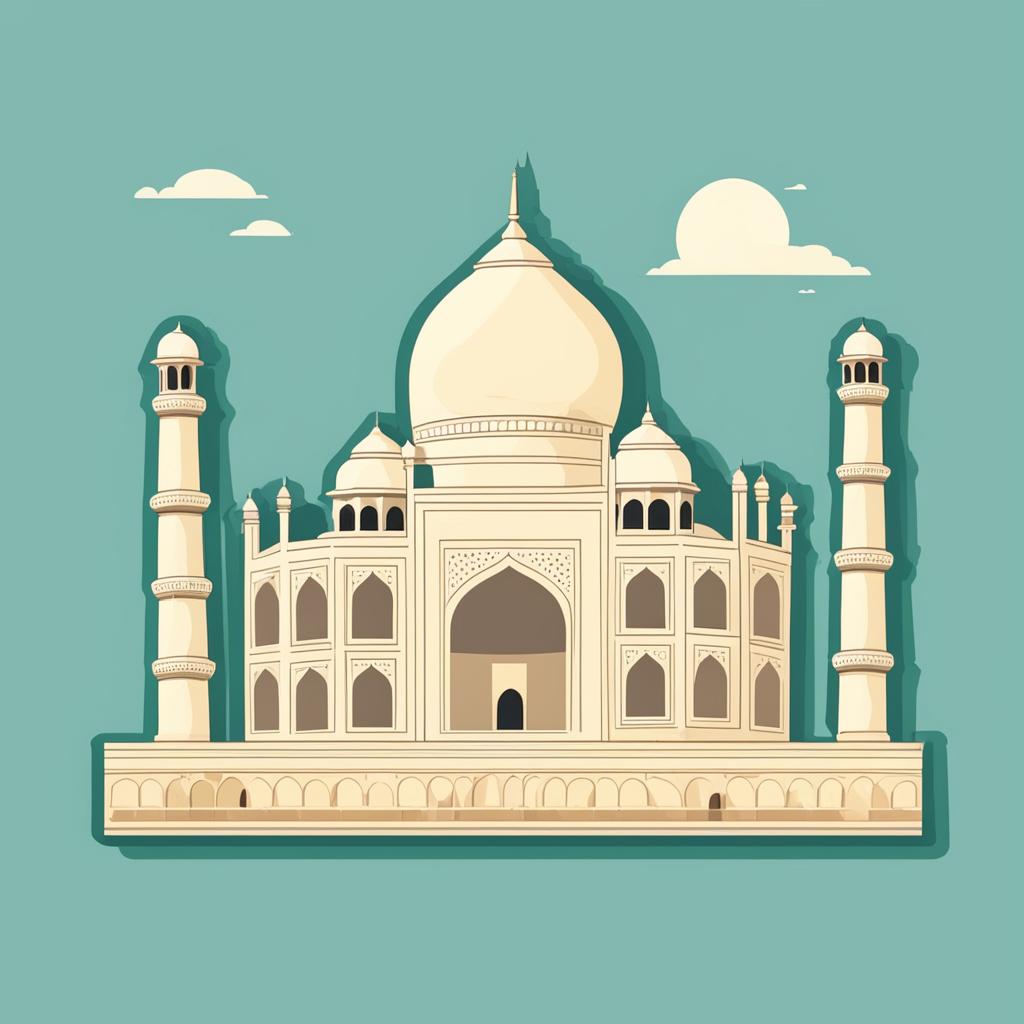}&
\includegraphics[width=0.14\textwidth]
{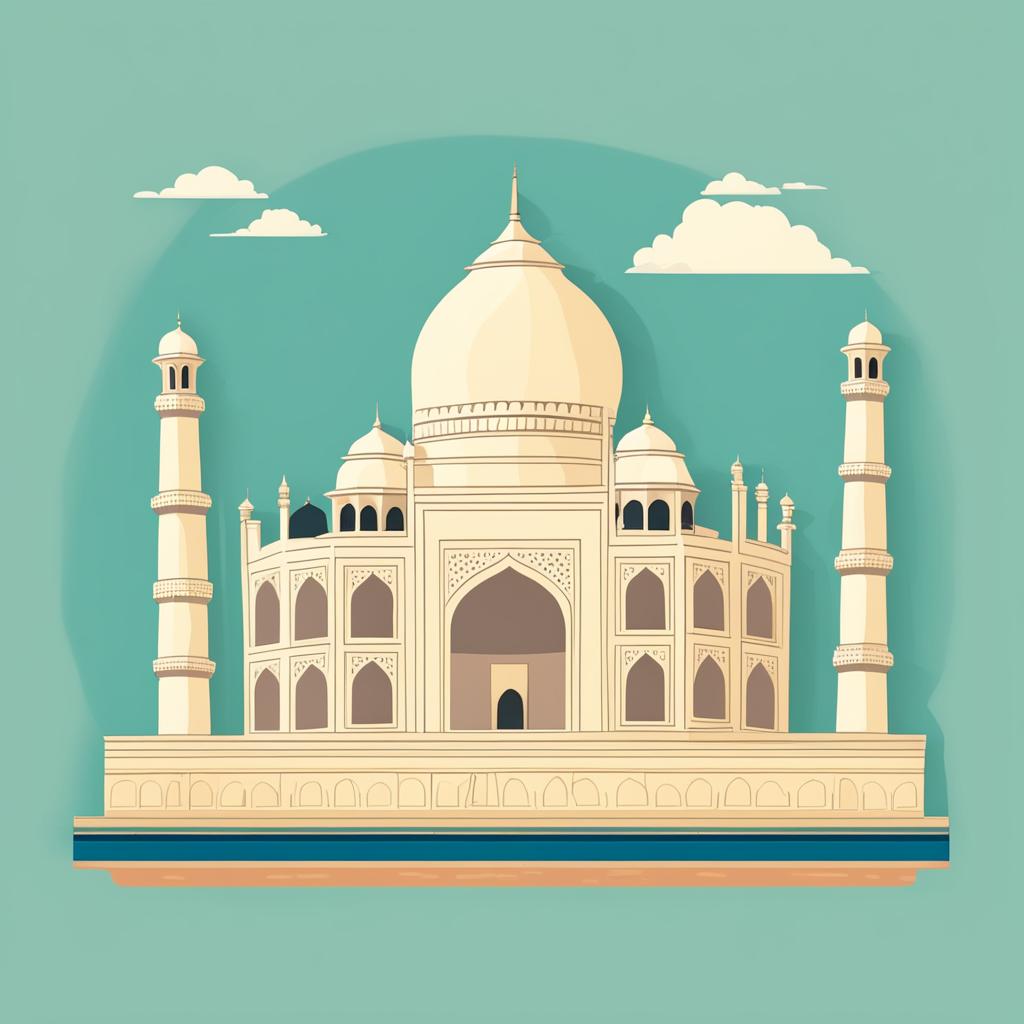}&
\includegraphics[width=0.14\textwidth]{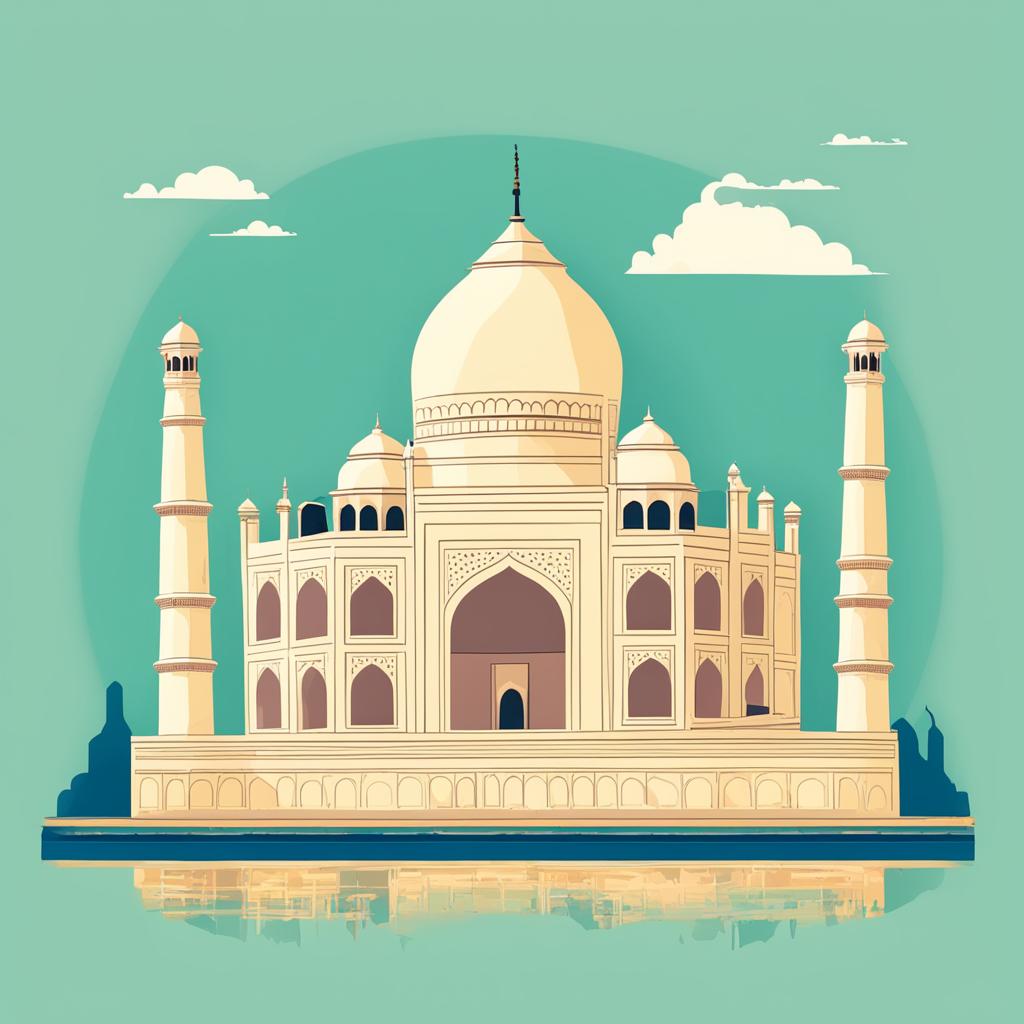}&
\includegraphics[width=0.14\textwidth]{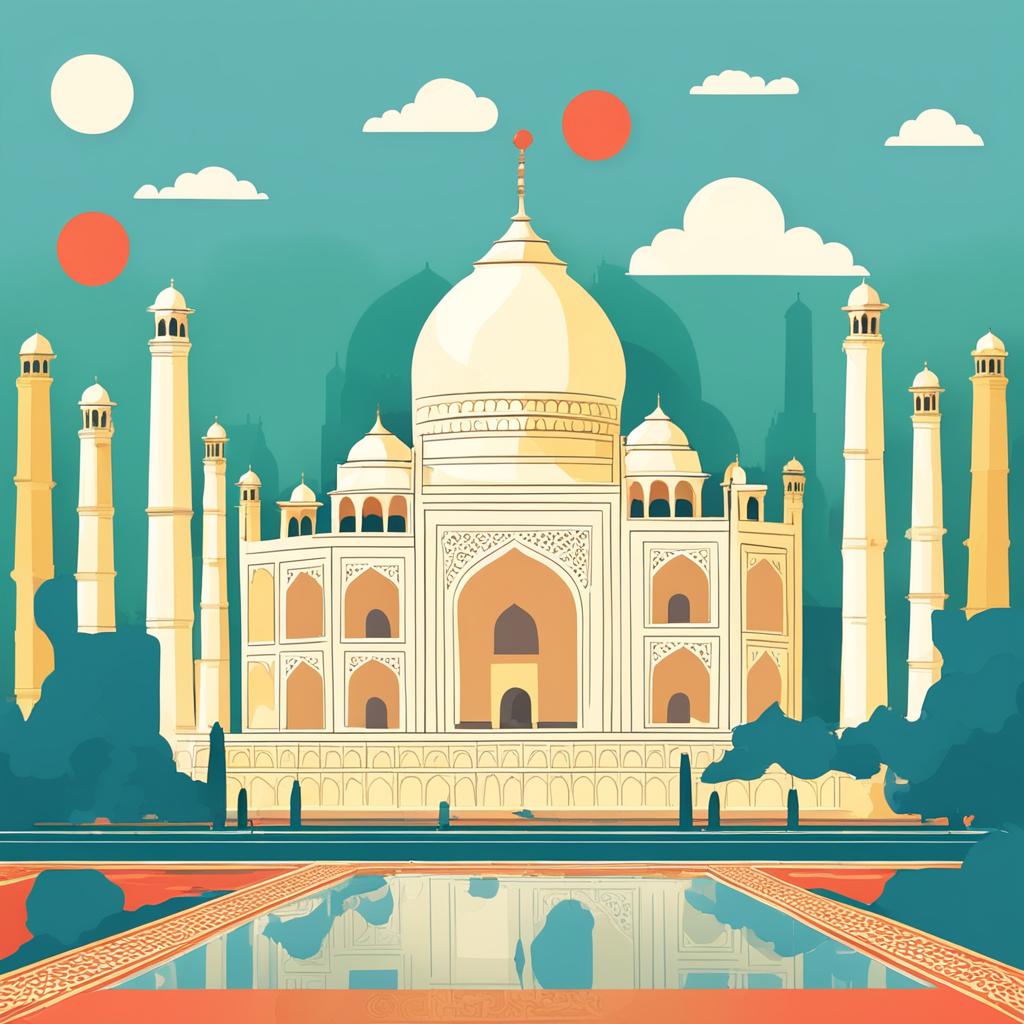}\\
$E^-$ = 3 &
\includegraphics[width=0.14\textwidth]{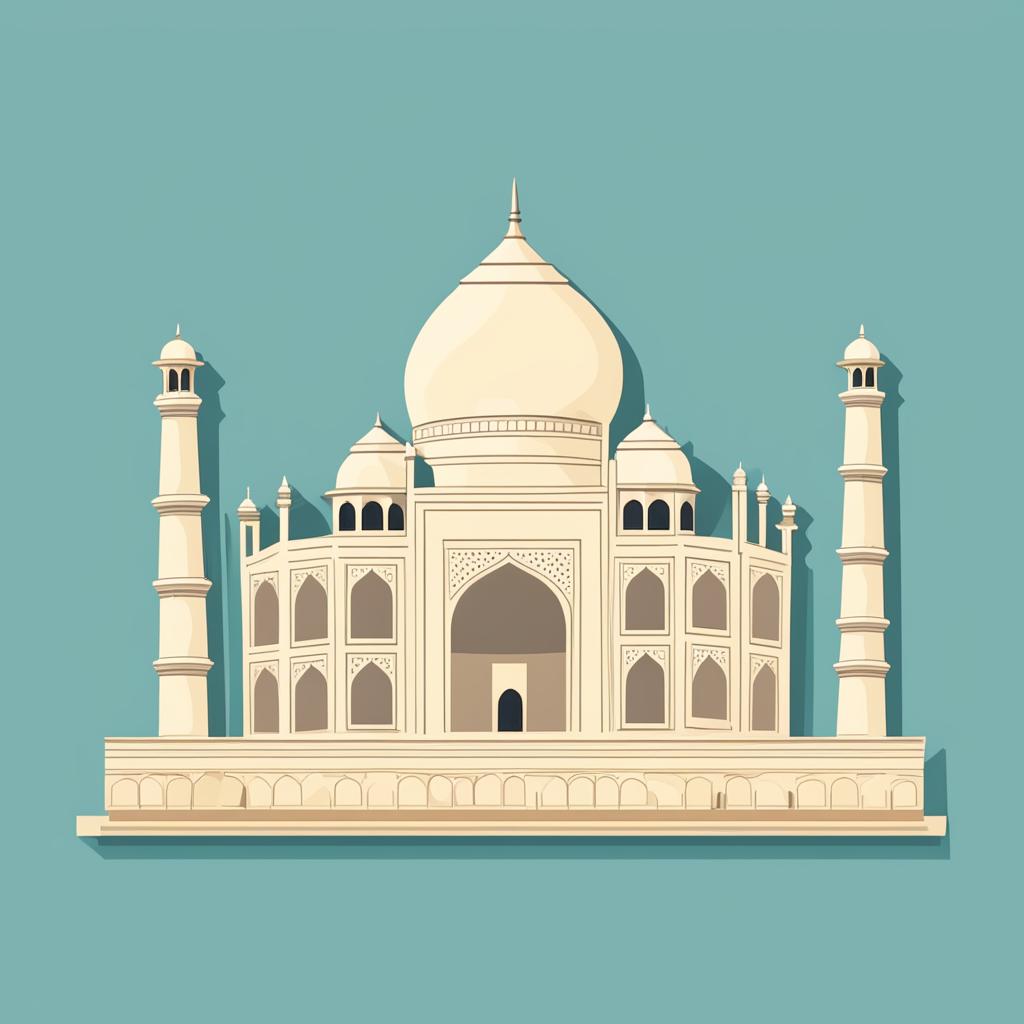}& \includegraphics[width=0.14\textwidth]{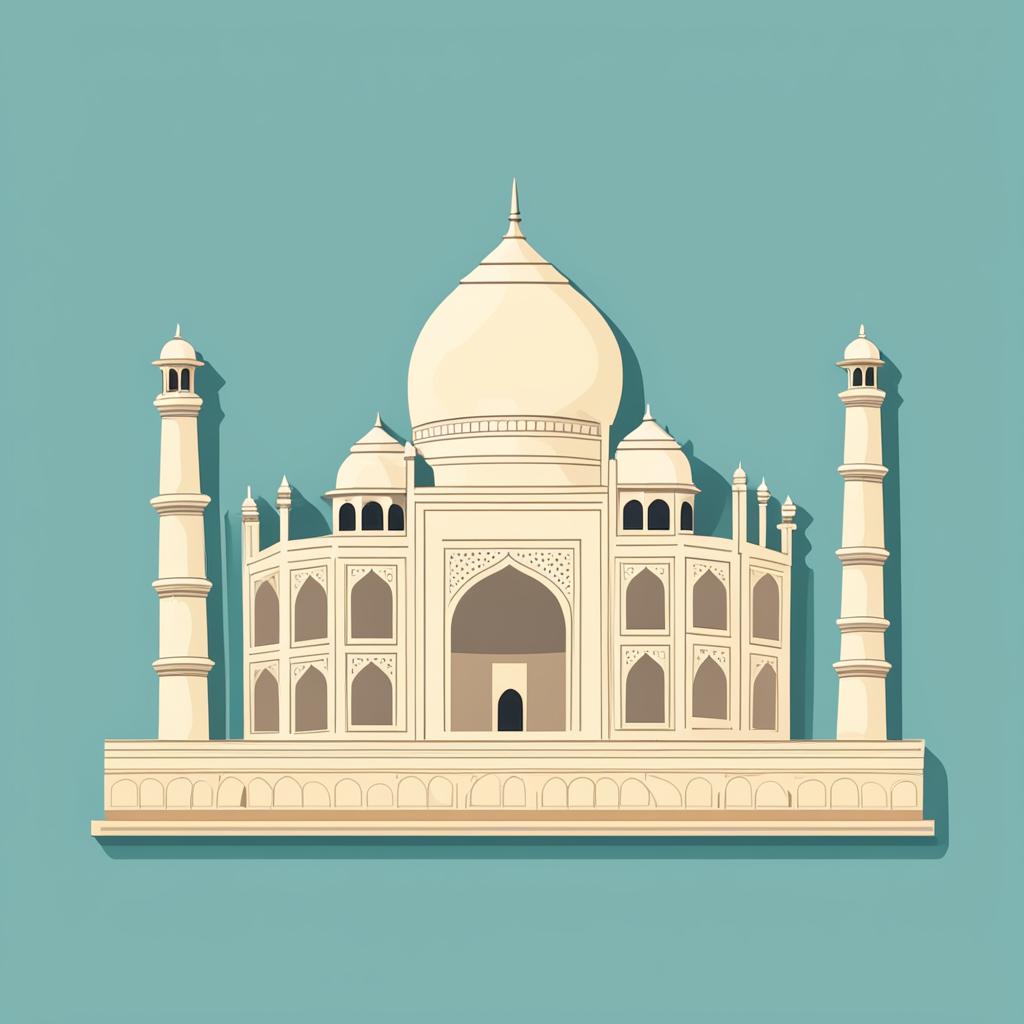}&
\includegraphics[width=0.14\textwidth]{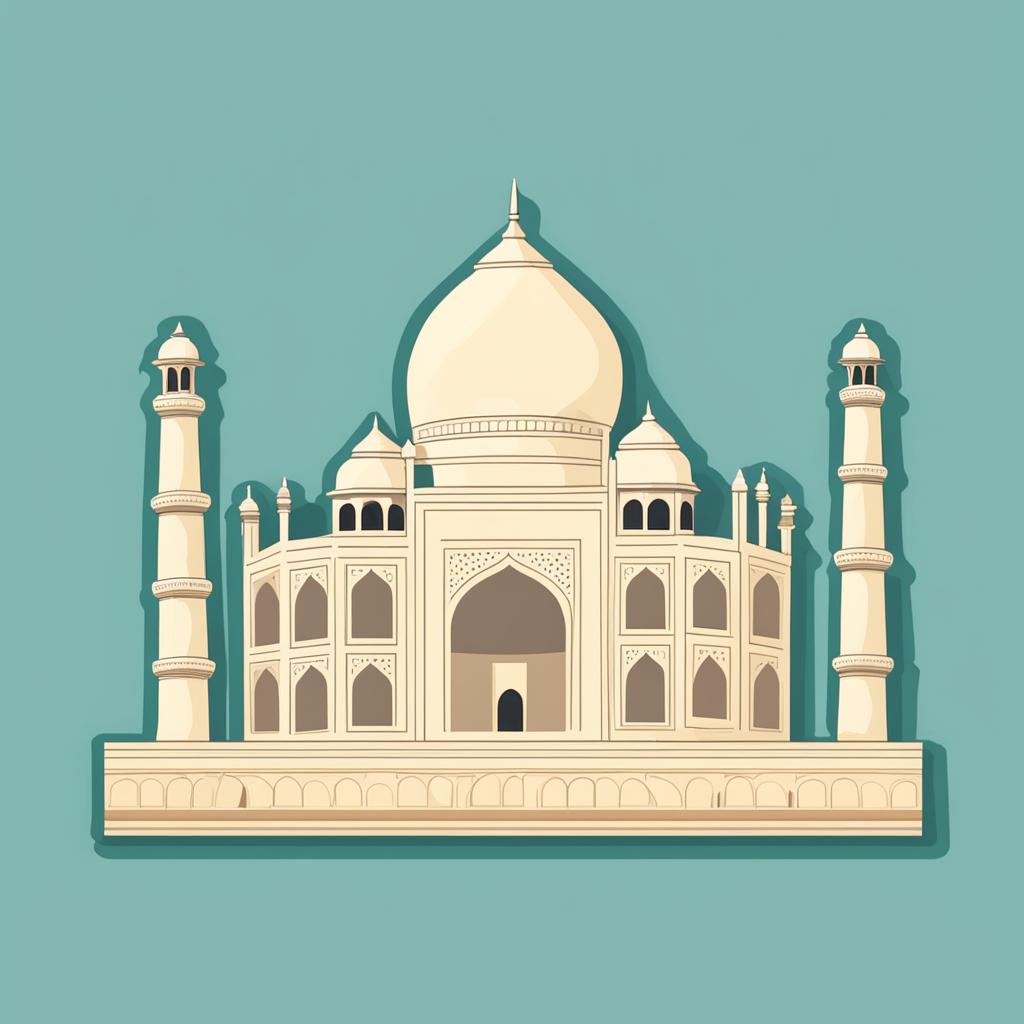}&
\includegraphics[width=0.14\textwidth]
{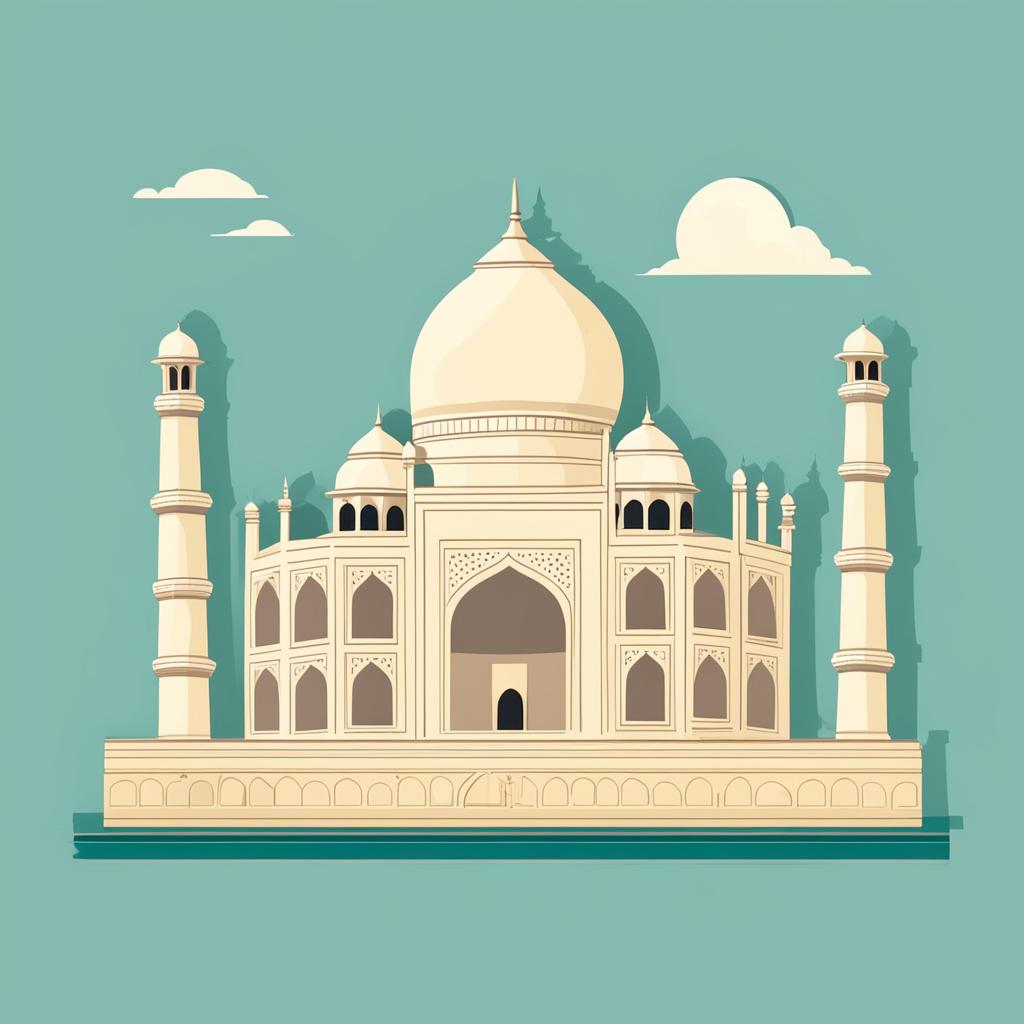}&
\includegraphics[width=0.14\textwidth]{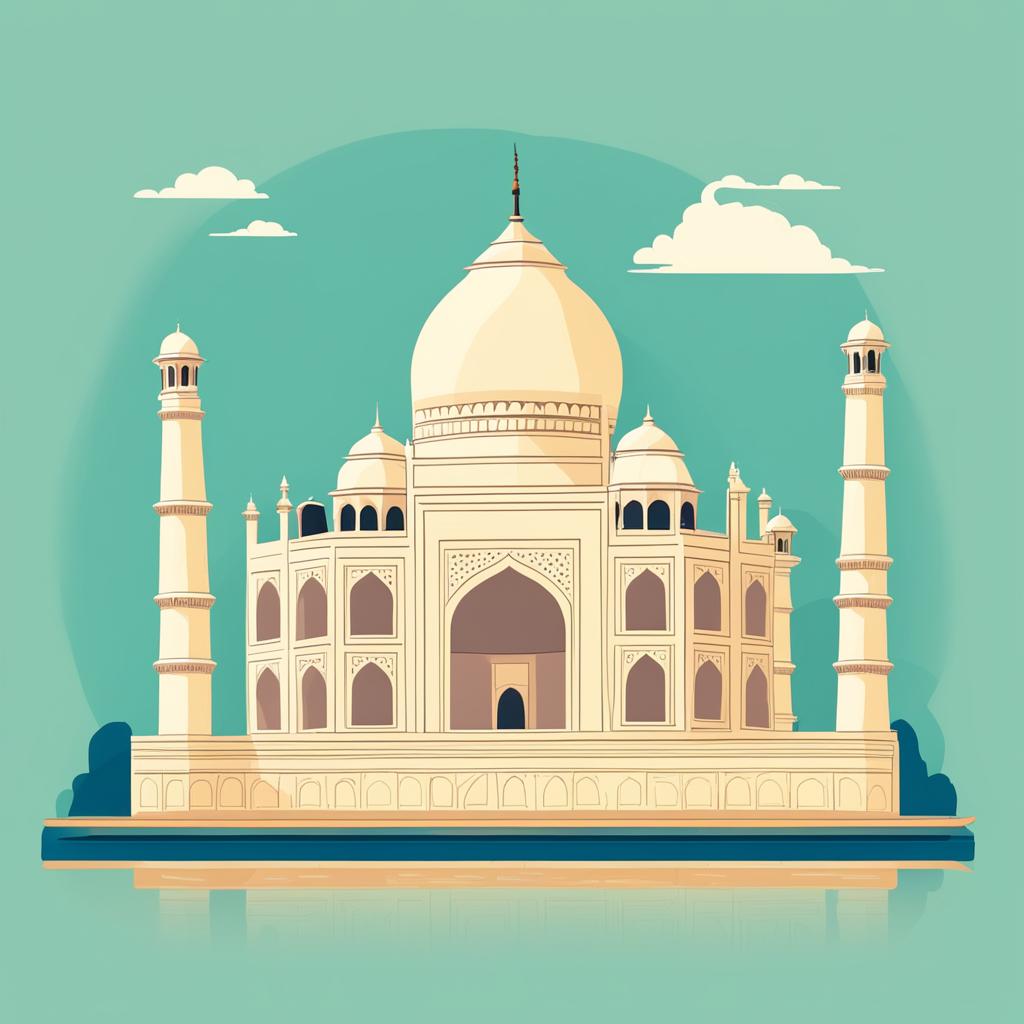}&
\includegraphics[width=0.14\textwidth]{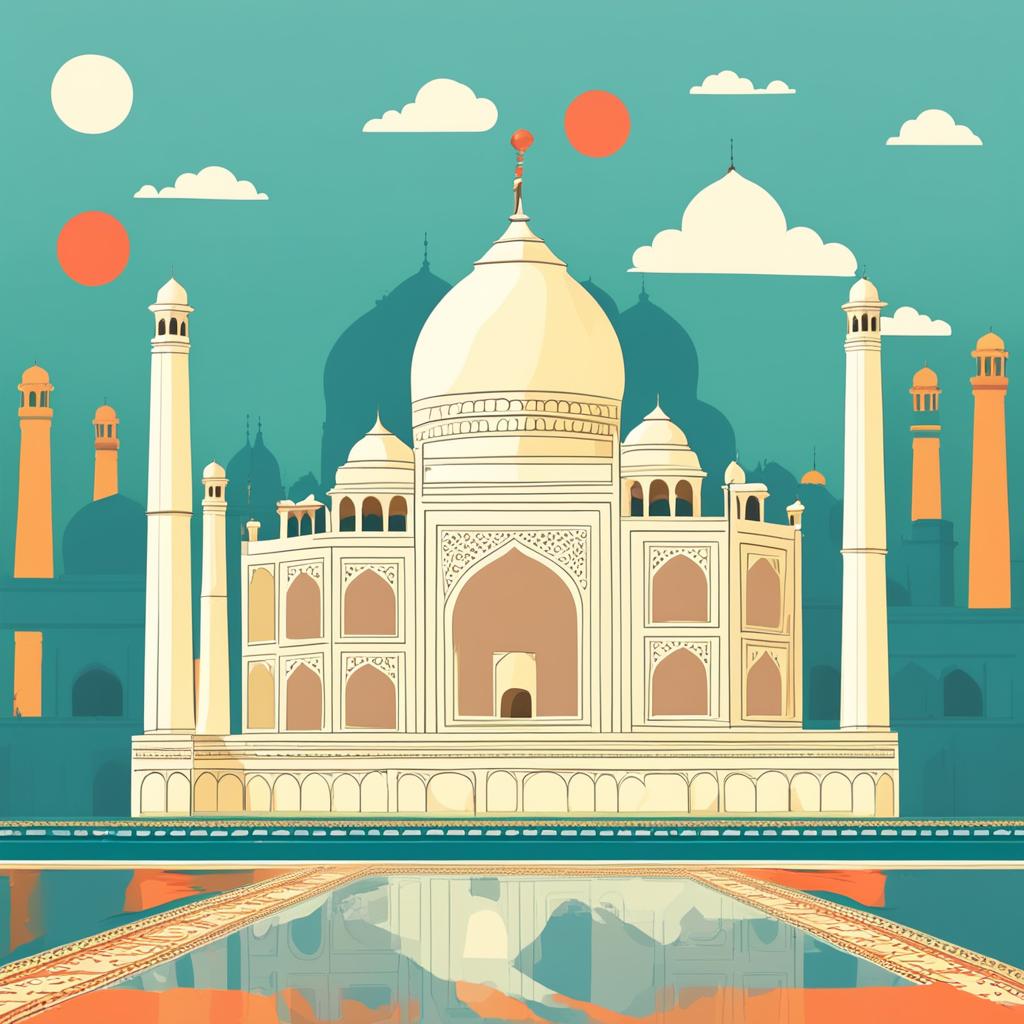}\\
$E^-$ = 5 &
\includegraphics[width=0.14\textwidth]{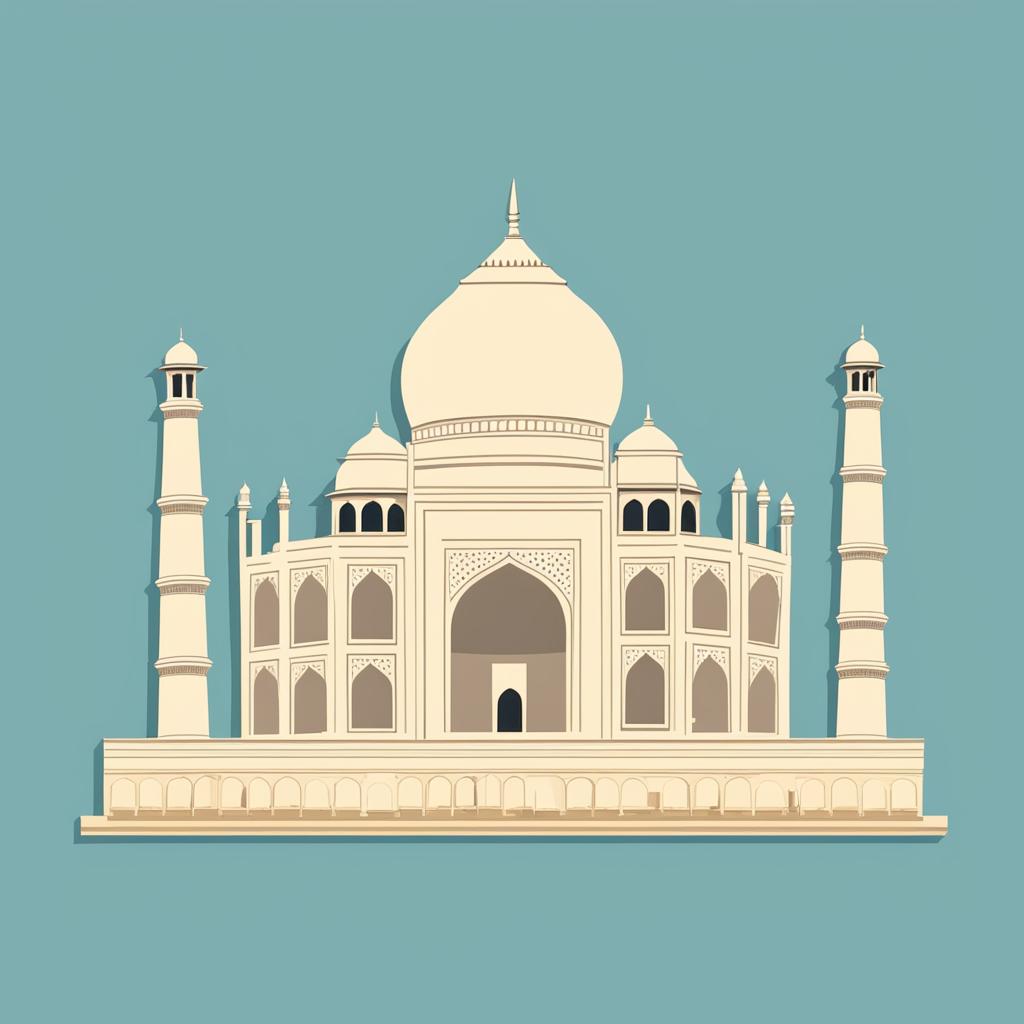}& \includegraphics[width=0.14\textwidth]{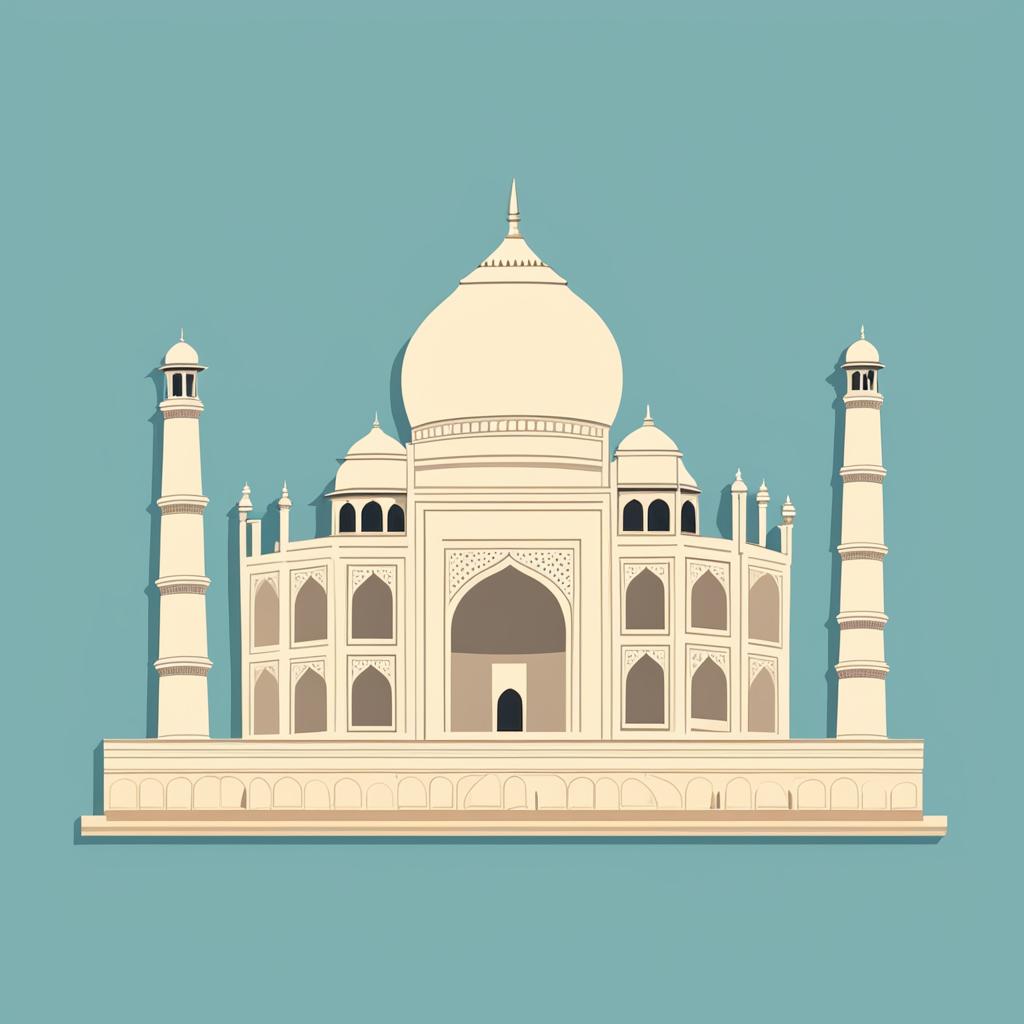}&
\includegraphics[width=0.14\textwidth]{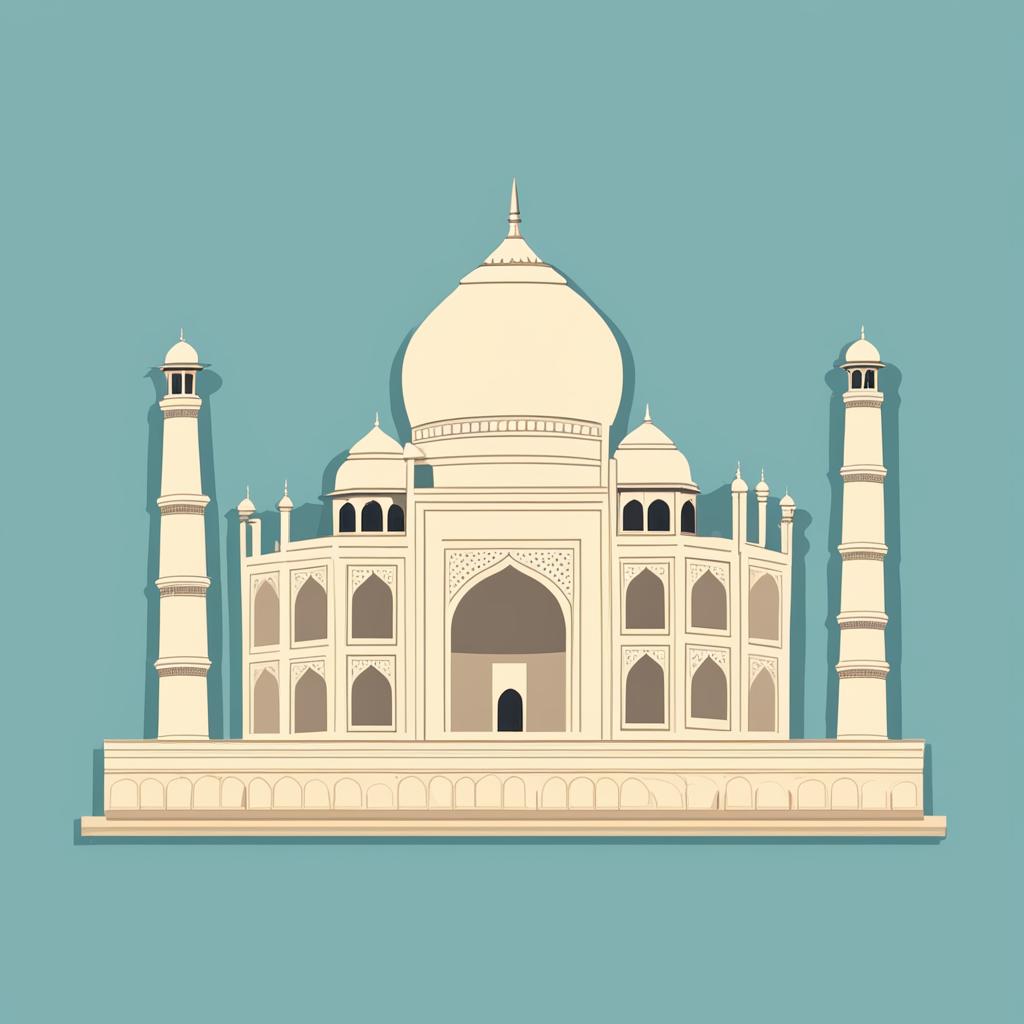}&
\includegraphics[width=0.14\textwidth]
{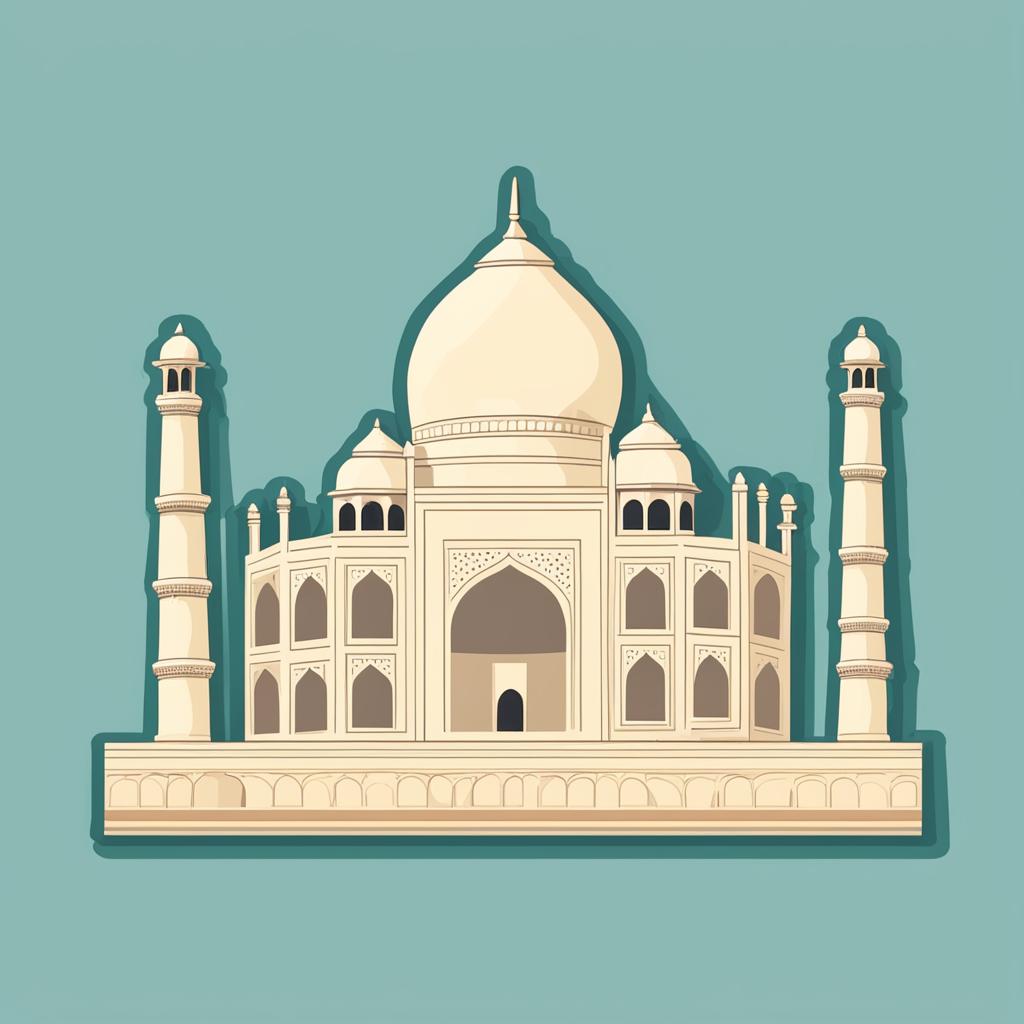}&
\includegraphics[width=0.14\textwidth]{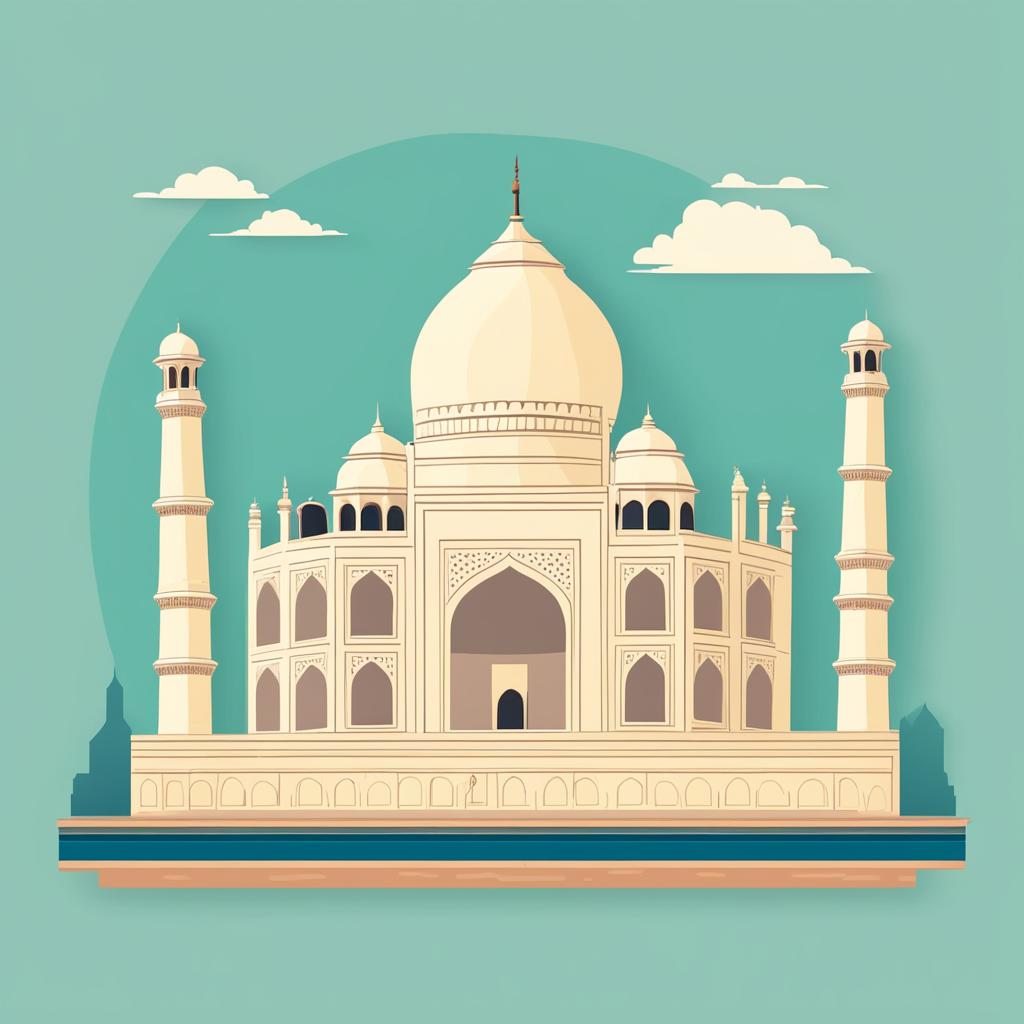}&
\includegraphics[width=0.14\textwidth]{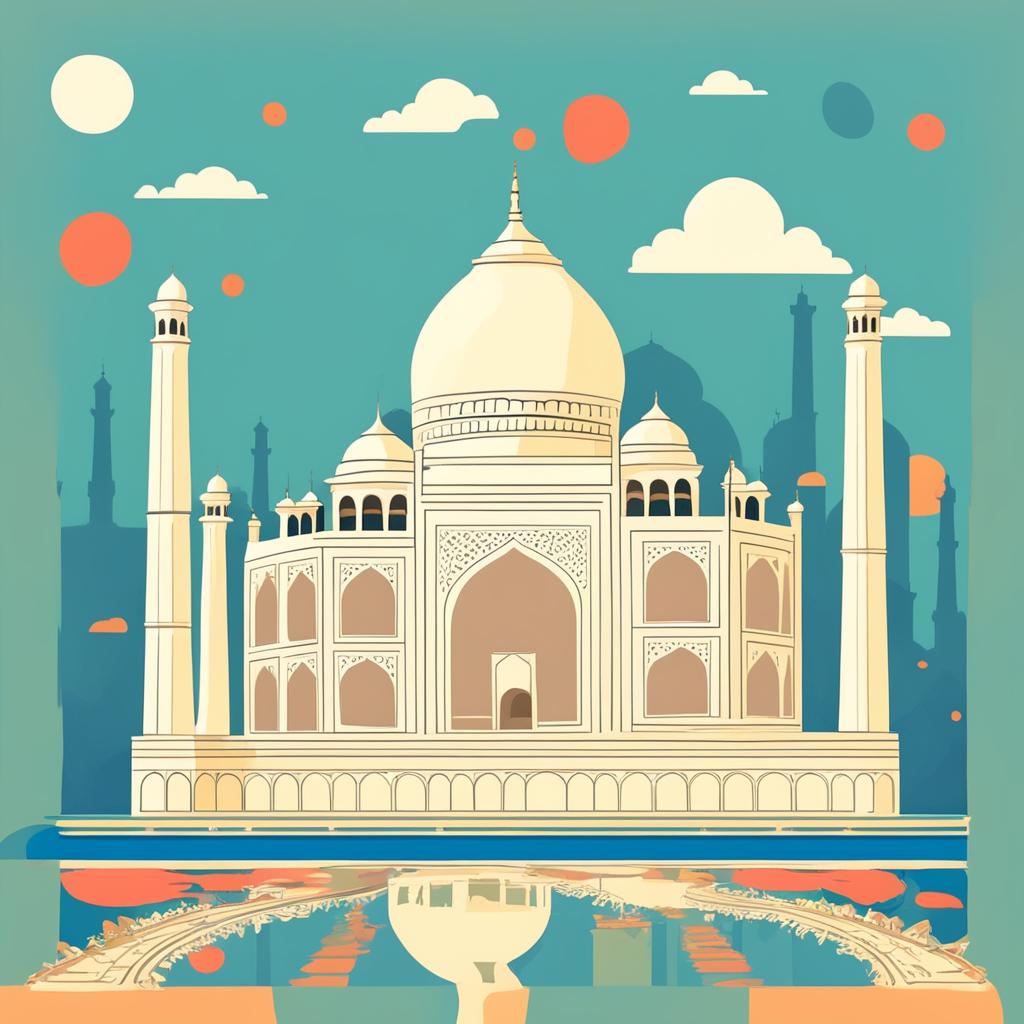}\\
$E^-$ = 8 &
\includegraphics[width=0.14\textwidth]{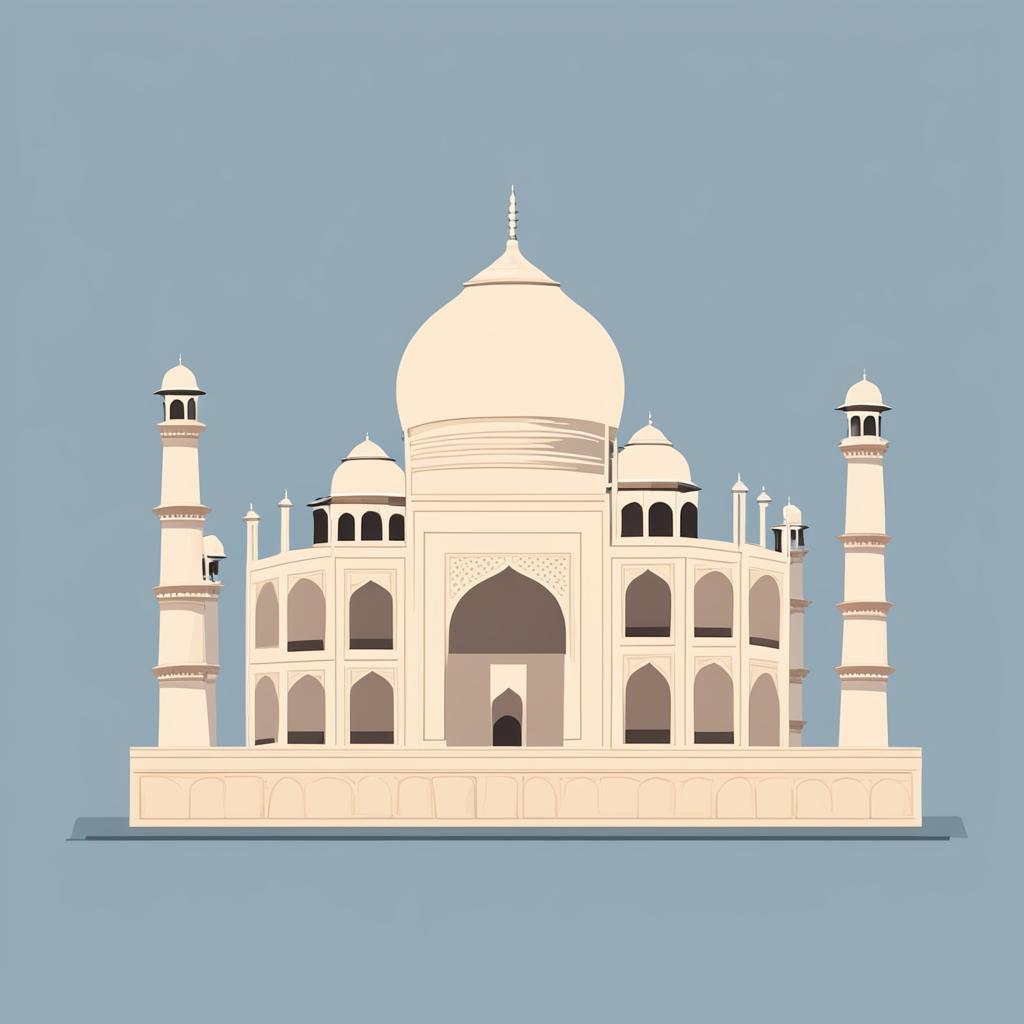}& \includegraphics[width=0.14\textwidth]{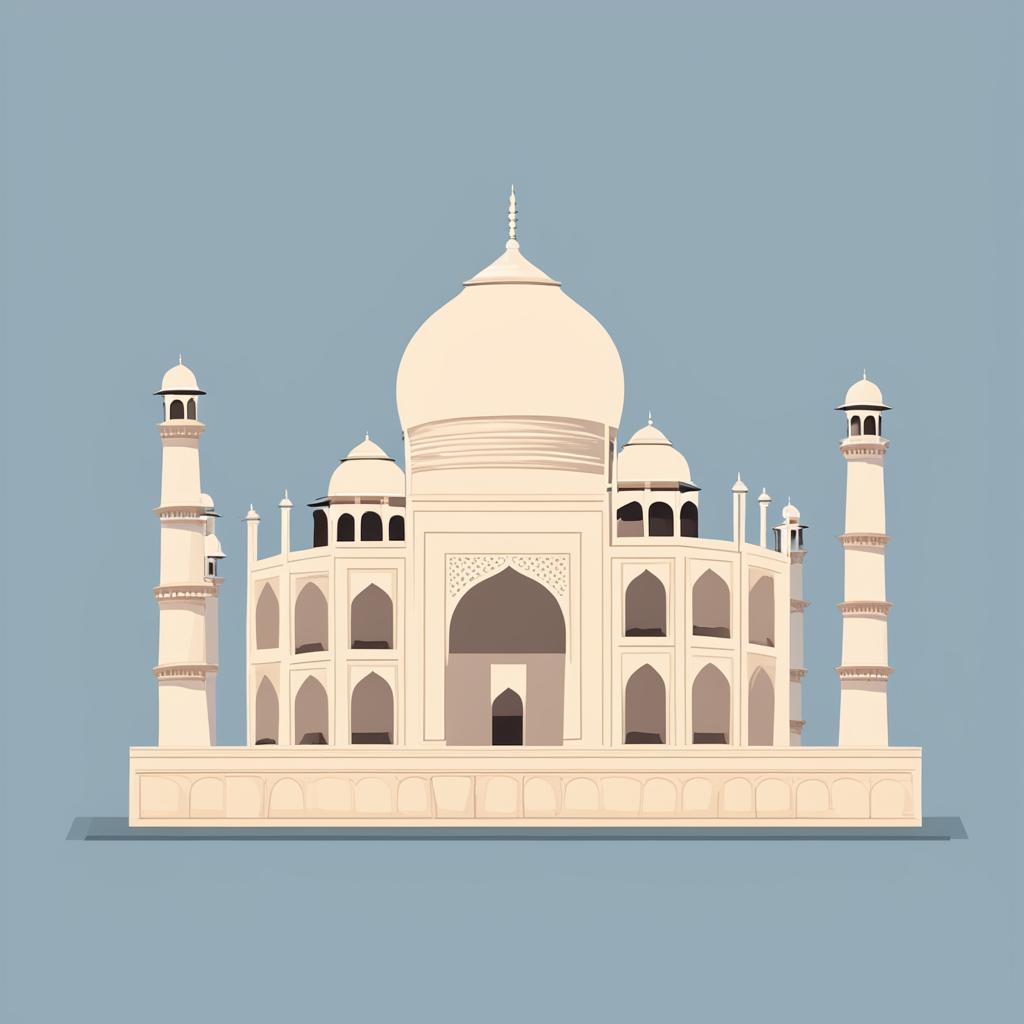}&
\includegraphics[width=0.14\textwidth]{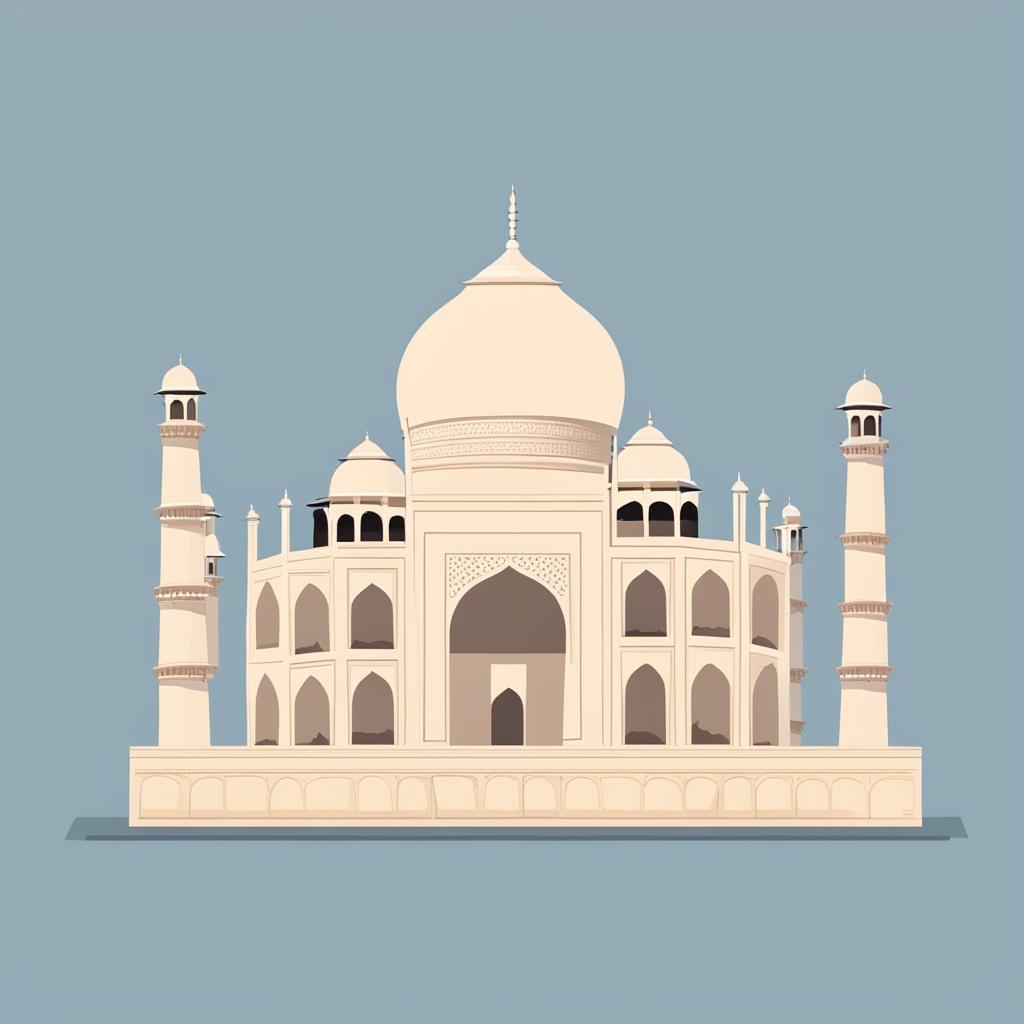}&
\includegraphics[width=0.14\textwidth]
{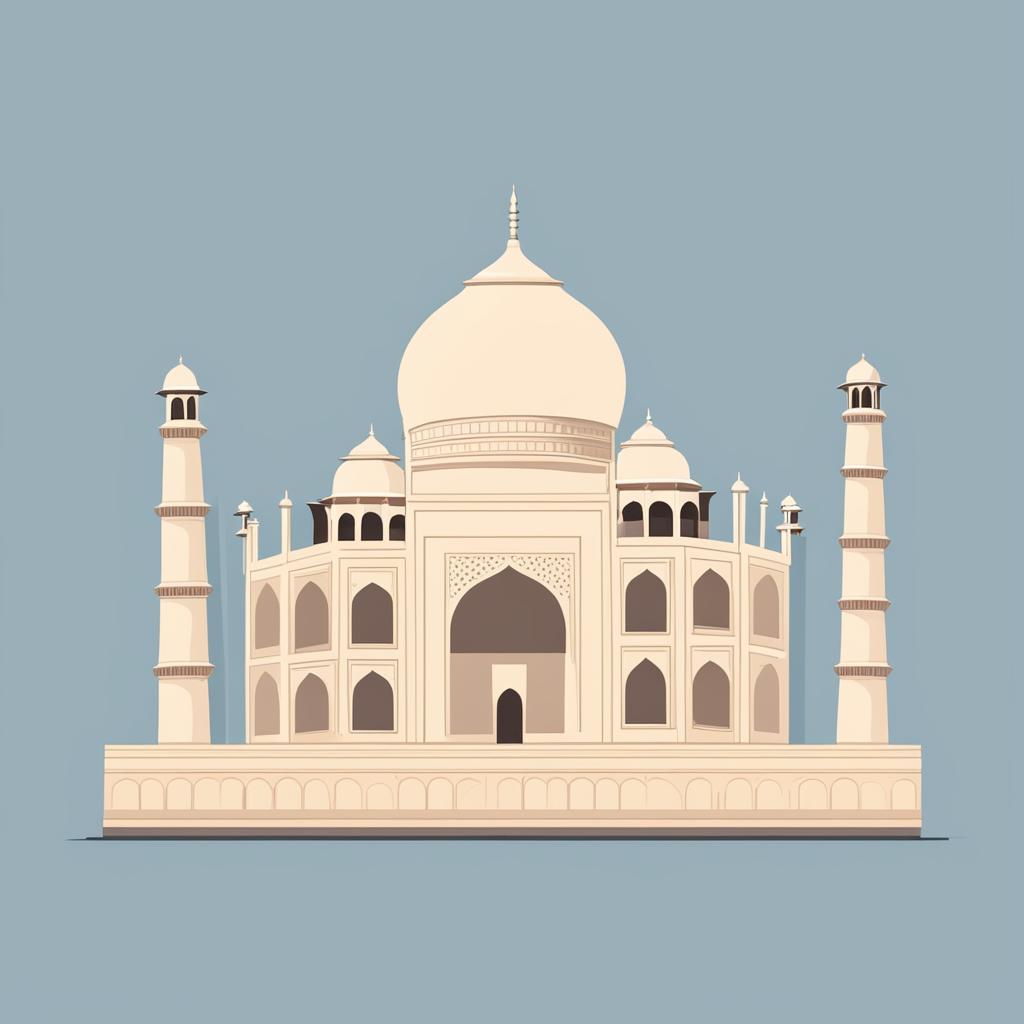}&
\includegraphics[width=0.14\textwidth]{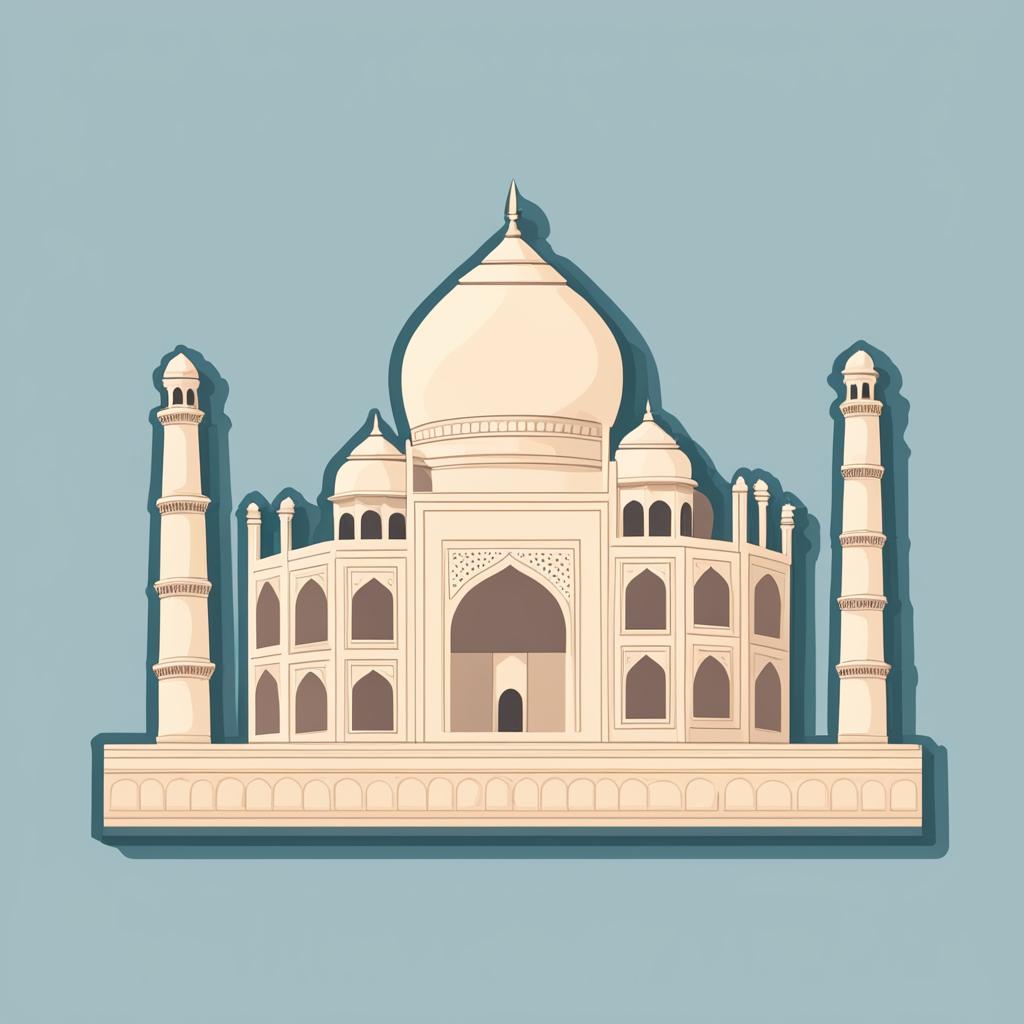}&
\includegraphics[width=0.14\textwidth]{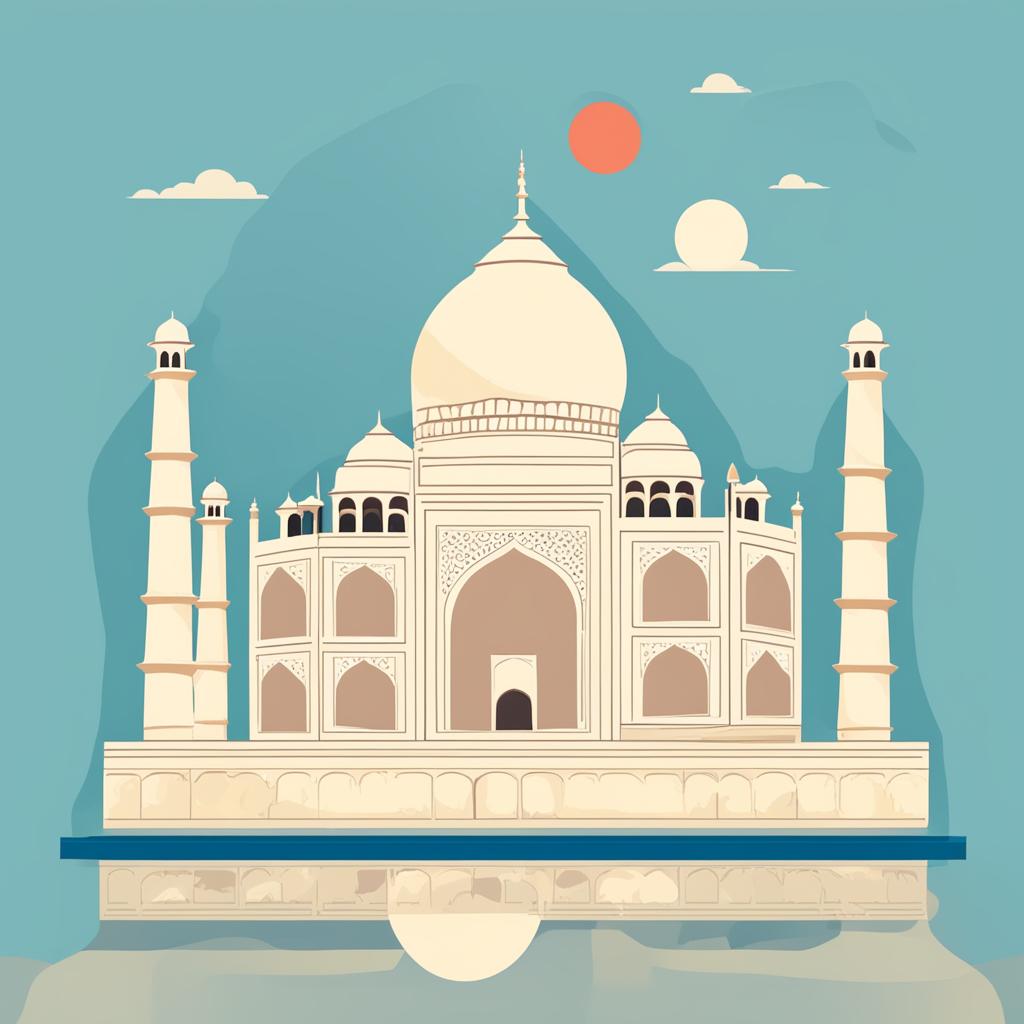}\\
\end{tabular}
\caption{Images generated using entropy only conditioned model, with different values of relative entropy. $E^+$ and $E^-$ represent positive and null entropy condition values respectively, resulting with a relative entropy guidance of $RE=E^+ - E^-$. prompt=``A poster of the Taj Mahal. Flat colors. Flat vector sticker style.''}
\label{tab:relative_etnropy}
\end{table*}

%\clearpage

%\input{sec/6_supplementary}

\end{document}